\documentclass[10pt,twocolumn,letterpaper]{article}

\usepackage{cvpr}     
\usepackage{adjustbox}
\usepackage{multirow}
\usepackage{caption}

\usepackage[dvipsnames]{xcolor}
\definecolor{cvprblue}{rgb}{0.21,0.49,0.74}

\usepackage{customization}
\usepackage[pagebackref,breaklinks,colorlinks,citecolor=cvprblue]{hyperref}
\def\paperID{XXX} 
\def\confName{CVPR}
\def\confYear{2024}

\title{Long-Tailed Anomaly Detection with Learnable Class Names}

\author{ Chih-Hui Ho\textsuperscript{1} \quad Kuan-Chuan Peng\textsuperscript{2} \quad Nuno Vasconcelos\textsuperscript{1}\\
\textsuperscript{1}University of California, San Diego \quad \textsuperscript{2}Mitsubishi Electric Research Laboratories (MERL)\\
{\tt\small {chh279@ucsd.edu} \quad  \tt\small {kpeng@merl.com} \quad \tt\small {nvasconcelos@ucsd.edu}}
}

\begin{document}
\hypersetup{citecolor=blue}
\maketitle
\begin{abstract}
Anomaly detection (AD) aims to identify defective images and localize their defects (if any). Ideally, AD models should be able to detect defects over many image classes; without relying on hard-coded class names that can be uninformative or inconsistent across datasets; learn without anomaly supervision; and be robust to the long-tailed distributions of real-world applications. To address these challenges, we formulate the problem of long-tailed AD by introducing several datasets with different levels of class imbalance and metrics for performance evaluation. We then propose a novel method, LTAD, to detect defects from multiple and long-tailed classes, without relying on dataset class names. LTAD combines AD by reconstruction and semantic AD modules. AD by reconstruction is implemented with a transformer-based reconstruction module. Semantic AD is implemented with a binary classifier, which relies on learned pseudo class names and a pretrained foundation model. These modules are learned over two phases. Phase 1 learns the pseudo-class names and a variational autoencoder (VAE) for feature synthesis that augments the training data to combat long-tails. Phase 2 then learns the parameters of the reconstruction and classification modules of LTAD. 
Extensive experiments using the proposed long-tailed datasets show that LTAD substantially outperforms the state-of-the-art methods for most forms of dataset imbalance. The long-tailed dataset split is available at \href{https://zenodo.org/records/10854201}{https://zenodo.org/records/10854201}.

\end{abstract}

\vspace{-5pt}
\section{Introduction} 
\begin{figure}[t]
    \centering
    \includegraphics[width=\linewidth]{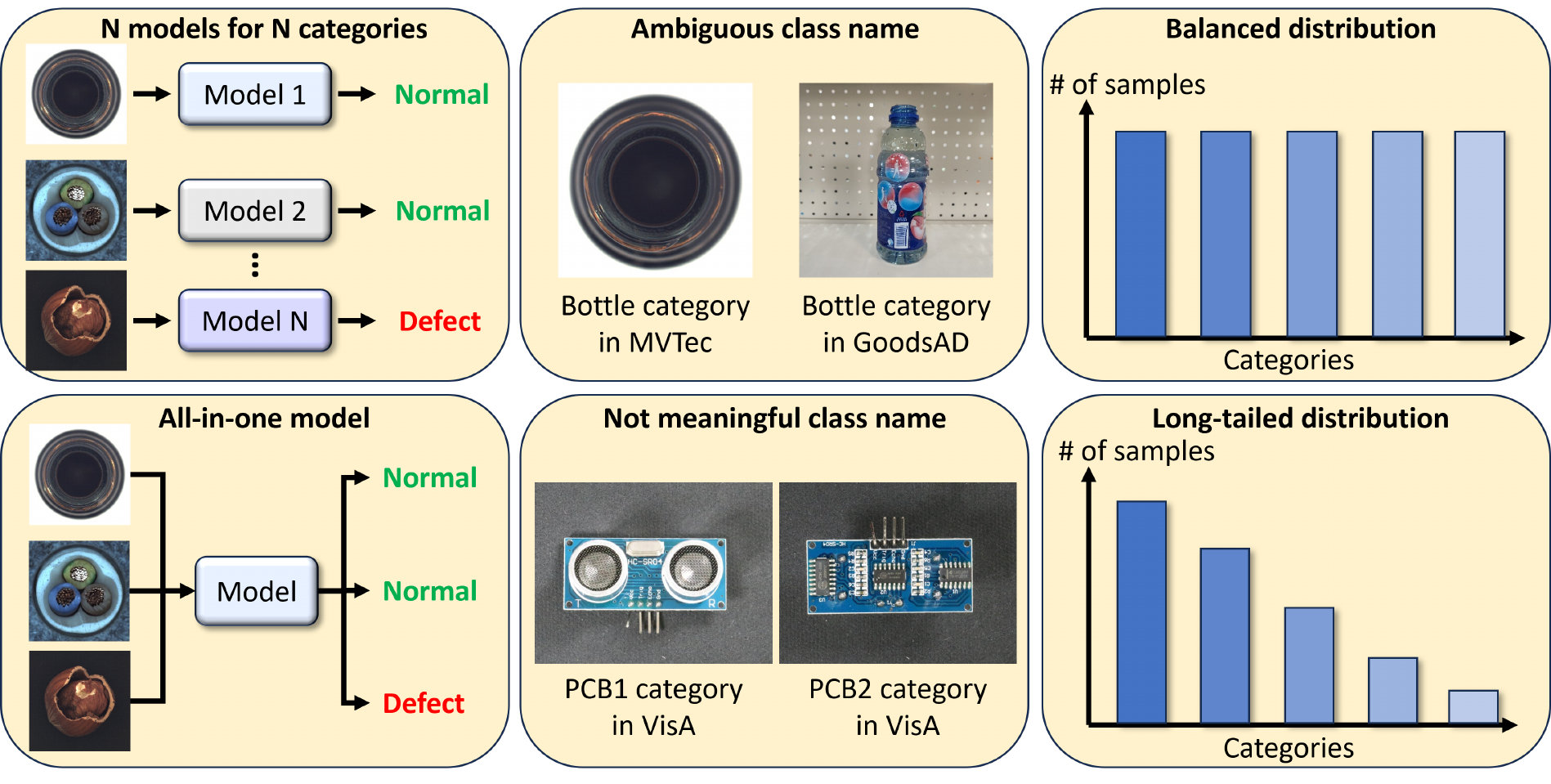}
    \caption{Challenges of long-tailed AD include (Left) designing a single model to detect anomalies over multiple image classes, (Middle) uninformative class names, and (Right) long-tailed data distributions. }
    \label{fig:teaser}
    \vspace{-10pt}
\end{figure}

\vspace{-3pt}
Anomaly detection (AD) is an important problem for 
many manufacturing settings~\cite{tire_defect,pcb_defect,patchcore,wafer_defect}. To reflect practical manufacturing constraints, most datasets~\cite{mvtec,visa,goodad} are curated under the unsupervised AD setting, where no defect images are available for training. Various methods~\cite{padim,patchcore,cutandpaste,RD,RD_plus} have shown that this problem can be solved with high accuracy; \eg,~\cite{ReConPatch,Li2023TargetBS,Bae2022PNII,RD,RD_plus,FastFlow,Lee2022CFACF,Kim2022FAPMFA,DeSTSeg, draem} have success rates $>$95\% for anomaly detection and localization on the MVTec dataset~\cite{mvtec}. However, as illustrated in Fig.~\ref{fig:teaser}, these methods require a different model per image category.
which compromises scalability to many classes.
Recently, there has been interest in more efficient methods that use a single model to detect anomalies in all object classes~\cite{uniad, omnial, regad, winclip, anomalygpt, sam_ad, april_gan}. These methods can be grouped according to the level of image semantics where they operate. On one hand, {\it AD by reconstruction\/} methods~\cite{uniad, omnial} use a reconstruction module to project the input image into the manifold of normal images. The difference between the image and its projection is then used to detect possible defects. On the other hand, {\it semantic AD\/} methods \cite{winclip, anomalygpt, sam_ad} build explicit models of normal/abnormal images. Given the absence of abnormal training data, this is done by leveraging  the knowledge of visual-language foundation models~\cite{clip, sam,imageblind,Su2023PandaGPTOM}. Abnormal regions are detected using the predefined text prompts for normal and abnormal plus an image class name, \eg, ``a normal photo of a [CLASS]" and ``an abnormal photo of a [CLASS]," where [CLASS] is a class name in the dataset, \eg, ``bottle" in the MVTec dataset. 

The two types of methods have limitations. Reconstruction methods require modeling the complex manifold, especially for problems requiring many classes. Even when trained on large datasets, the distance between the input image and this manifold can be smaller for certain anomalies.
The foundation model used by semantic AD methods can provide additional clarity, because it enables framing AD as a binary classification problem. However, this is difficult when the dataset class names are ambiguous or unknown to the foundation model. Fig.~\ref{fig:teaser} shows an example of ambiguity due to the fact that the class name ``bottle" refers to visually different concepts in MVTec~\cite{mvtec} (where it means ``bottle bottom") and the GoodsAD dataset~\cite{goodad} (where it means ``bottle side"). 
Hence, in MVTec, the ``bottle" label may not be the most informative for the foundation  model, which may associate the images with alternative labels, \eg 
%``hockey puck" or 
``black sphere." Sometimes, class names can be simply unknown to the foundation model, \eg the classes ``PCB1" and ``PCB2" also shown in Fig.~\ref{fig:teaser}. This suggests that the foundation model should 
learn what are the class names that best align with these images.
When this is difficult, semantic AD might benefit from the flexibility of the AD by reconstruction methods, which are not constrained by class names. 
Hence, this work investigates the design of AD methods that combine both AD by reconstruction and semantic AD.

Beyond this, it remains unclear whether the resulting models generalize to the {\it long-tailed\/} setting~\cite{He2008ADASYNAS,Zou2018DomainAF, Arora2017GeneralizedZL,Cui2019ClassBalancedLB,Kang2019DecouplingRA} where, as illustrated in Fig.~\ref{fig:teaser}, the sample distribution is skewed. 
This is particularly important because long-tailed distributions are natural in manufacturing, where different objects 
can have very different popularity.
We formulate the problem of {\it long-tailed AD} by introducing long-tailed datasets, which are obtained by resampling current AD benchmarks with different imbalance factors and types of imbalance. We also propose a set of performance metrics for the  long-tailed setting.

To address the challenges above, we propose a new method, \ours, which combines AD by reconstruction and semantic AD. AD by reconstruction is implemented by combining the ALIGN~\cite{align} image encoder and a transformer-based reconstruction module (RM), trained to project image patches into the manifold of normal images. An anomaly score is then obtained by computing the difference between the input image and the result of this projection. The latent patch representation produced by the ALIGN image encoder is also mapped to the feature space of the ALIGN text encoder, to enable the implementation of semantic AD. For this, a binary normal/abnormal classifier is implemented in ALIGN text space, by using as classifier weight vectors the ALIGN representation of text prompts for ``normal" and ``abnormal." The posterior probability of the abnormal class, under this classifier is then used as a second anomaly score. Anomalies are detected with a combination of the two AD scores.

To address the long-tailed setting, we propose a preliminary training phase for data augmentation. This consists of learning a VAE~\cite{vae,vae_intro}, which is
then used to synthesize features.  To make these class sensitive, the VAE is conditioned by the text encoding of the [CLASS] name, according to the ALIGN model. However, to address the ambiguity of class names, a set of learnable [CLASS] prompts are learned by backpropagation, during VAE training. In a second training phase,
a mix of real and synthetic examples is used to train the \ours model. Since there are no training anomalies, these are simulated by adding noise to the synthesized features during this stage.

Overall, we make the following contributions:
\begin{enumerate}[label=\arabic*., leftmargin=*, topsep=0pt]
\setlength\itemsep{-0.0em}
    \item We show that prior methods do not perform well on long-tailed setting and formulate long-tailed AD based on 3 datasets with 9 imbalance settings and performance metrics.
    \item We propose \ours, which combines AD by reconstruction and semantic AD, performs multi-class AD, and overcomes dataset [CLASS] name ambiguity by learning names consistent with the semantic space of the ALIGN model.
    \item We propose a new training strategy for \ours which uses a novel data-augmentation procedure to address the data scarcity of long-tailed data, and learn [CLASS] names.
    \item We show that \ours outperforms the SOTA methods on the long-tailed AD. Extensive ablations confirm the efficacy of the various \ours modules, showing that \ours generalizes across various datasets and imbalance configurations.
\end{enumerate}

\section{Related works}
\begin{table}[!t]
\setlength{\tabcolsep}{10pt}
\resizebox{\columnwidth}{!}{%
\begin{tabular}{@{}rc@{\hspace{.5em}}c@{\hspace{.5em}}c>{\columncolor{red!10}}c}
\toprule
\multirow{2.5}{*}{\textbf{{Unsupervised AD Method} Conditions}} & \multicolumn{4}{c}{\textbf{Unsupervised AD Categories}} \\ \cmidrule(l){2-5} 
 & $\cC_0$ & $\cC_1$ & $\cC_2$  & \ourso \\
\midrule
Single model for all classes  & \ccross & $\ccheck$ & $\ccheck$  & $\ccheck$ \\
No class name prior & $\ccheck$  & $\ccheck$   & \ccross & $\ccheck$ \\
Designed for the long-tailed setting
  & \ccross & \ccross & \ccross  & $\ccheck$ \\
Learnable class names
  & \ccross & \ccross & \ccross  & $\ccheck$ \\
\bottomrule
\end{tabular}%
}
\caption{\ours addresses some important challenges for real-world AD, previously not considered in (\eg, $\cC_0$: \cite{patchcore,padim,spade,cutandpaste,DeSTSeg,RD,RD_plus,ReConPatch,draem,mkd,FastFlow,differnet,JimenezRezende2015VariationalIW,defectgan,rgi,DiffusionAD}, $\cC_1$: \cite{uniad,omnial,regad,anomalygpt,Lu2023HierarchicalVQ}, $\cC_2$: \cite{Zhang2023ExploringGP,winclip,Cao2023TowardsGA,sam_ad, april_gan}). 
}
\vspace{-1pt}
\label{tab:related_works}
\vspace*{-\baselineskip}
\end{table}

\textbf{Unsupervised anomaly detection (AD):} Unsupervised AD aims to identify defective images and localize the defects without observing any defect images during training. Tab.~\ref{tab:related_works} groups recent AD methods into 3 categories. \textbf{Category $\cC_0$~}contains earlier works~\cite{patchcore,padim,spade,cutandpaste,DeSTSeg,RD,RD_plus,ReConPatch} that use a different model per image category. 
Student-teacher methods~\cite{DeSTSeg,RD,RD_plus} use a pretrained teacher encoder and a student encoder optimized to match its predictions. AD is based on the difference of their predictions.  
Flow based methods~\cite{FastFlow, differnet,JimenezRezende2015VariationalIW} fit a Gaussian distribution to the feature vectors of normal images 
and use out-of-distribution criteria to perform AD. Reconstruction based methods~\cite{defectgan,rgi,DiffusionAD, draem} 
train models to reconstruct normal samples and use reconstruction error for AD.  Some other methods in category $\cC_0$ are discussed in the AD survey papers~\cite{Liu2023DeepII,Chalapathy2019DeepLF,Chandola09,salehi2022a}. More parameter efficient than these methods, \ours only needs a single model for all the classes instead of one model per class.

This is the setting adopted by  the more recent methods of \textbf{Category $\cC_1$}~\cite{regad, uniad,omnial, Lu2023HierarchicalVQ, anomalygpt}. For example, UniAD~\cite{uniad} improves prior reconstruction-based methods by using a neighborhood attention mask to avoid information leak, while
AnomalyGPT~\cite{anomalygpt} uses a large vision-language model to provide explanations for defective regions. While typically leveraging foundation models, these methods do not use the class name to detect anomalies. This akin to \ours but unlike methods in \textbf{category $\cC_2$}~\cite{Zhang2023ExploringGP, winclip, Cao2023TowardsGA, sam_ad, april_gan}, which also use a single model for all classes but require class names.  For example, 
WinCLIP~\cite{winclip} uses the CLIP~\cite{clip} model to compute the anomaly score, by measuring similarity between image and text feature vectors for several predefined normal/abnormal text prompts.
Some works in categories $\cC_1$ and $\cC_2$~\cite{april_gan, anomalygpt} also leverage auxiliary training data (\eg, train on VisA~\cite{visa} and test on MVTec~\cite{mvtec}), a setting that is not considered in  this work. 

More importantly, all the prior AD works assume balanced datasets~\cite{mvtec,visa,goodad,dagm}, where the number of samples is relatively balanced across classes. However, this is an unlikely setting for real-world applications where different objects tend to have different popularity. 
As summarized in Table~\ref{tab:related_works}, this work addresses the combinatorial challenge of imbalanced training set, absent class name, and single model for multiple object classes.

\begin{figure}
    \centering
    \resizebox{\columnwidth}{!}{%
    \begin{tabular}{cc}
    \includegraphics[width=0.49\linewidth]{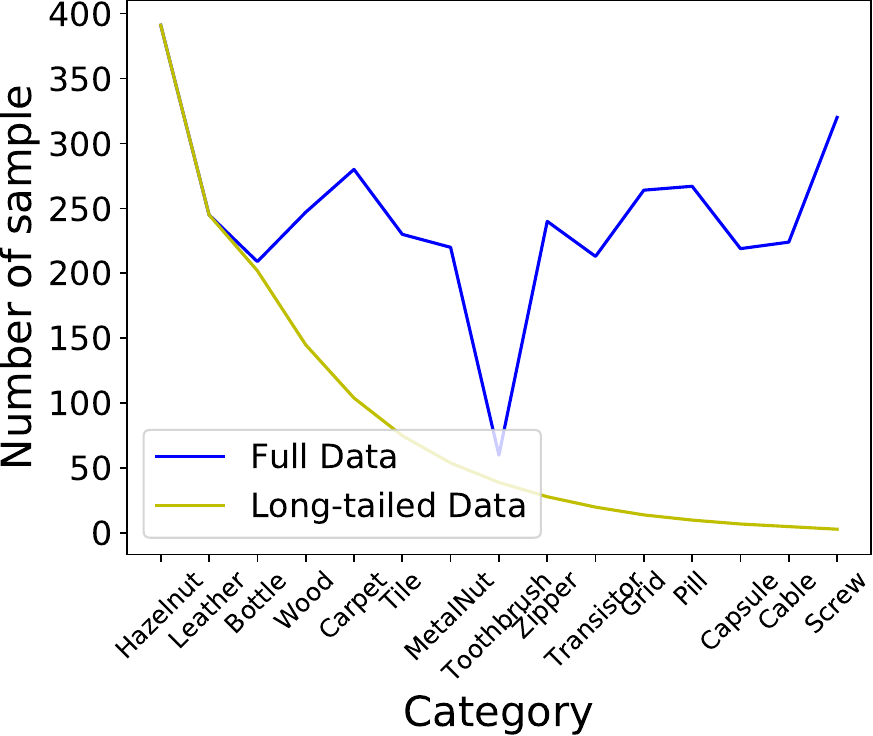} &
    \includegraphics[width=0.49\linewidth]{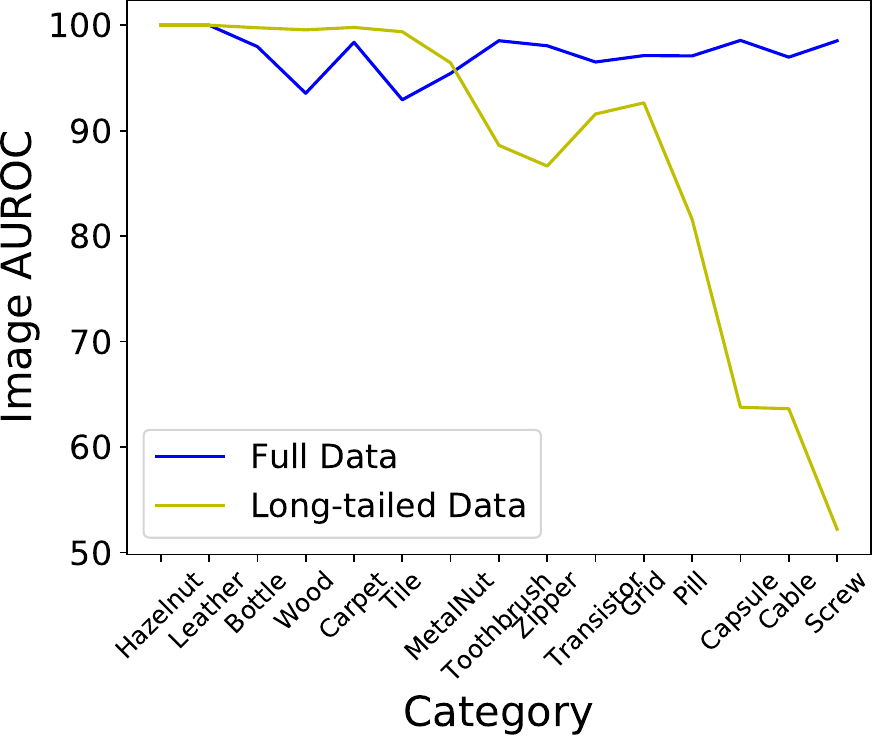}
    \\
    \small{(a) dataset distribution} & \small{(b) anomaly detection performance}
    \end{tabular}
    }
    \caption{Preliminary study with UniAD on MVTec. Image classes (x-axis) are sorted by popularity. (a) Dataset distribution of MVTec vs. long-tailed version. (b) AD performance on the two datasets.}
    \label{fig:prelim_study}
    \vspace{-10pt}
\end{figure}

\textbf{Long-tailed recognition:} Real-world data distributions are often imbalanced across classes. Prior works in classification~\cite{Cui2019ClassBalancedLB,Kang2019DecouplingRA,Krishnan2018AnAA,Li2021MetaSAugMS,Wu2020SolvingLR} have shown that the data imbalance degrades performance for minority, or tail, classes. Long-tailed recognition methods aim to avoid this. Prior long-tailed recognition methods related to this work can be mainly categorized into (1) data re-sampling, (2) loss re-weighting, and (3) representation learning. Data re-sampling methods~\cite{He2009LearningFI,Buda2017ASS,Chawla2002SMOTESM,Yap2013AnAO,Drummond2003C4,Han2005BorderlineSMOTEAN,Chu2020FeatureSA} balance the sample distribution by under-sampling majority classes or augmenting minority classes. For example, \cite{He2008ADASYNAS,Zou2018DomainAF, Arora2017GeneralizedZL,Krishnan2018AnAA,Li2021MetaSAugMS} utilize a generative model to synthesize samples or features from minority classes.
Loss re-weighting methods~\cite{focal_loss, Shu2019PushTS,Cao2019LearningID,Cui2019ClassBalancedLB,Kang2019DecouplingRA,Ren2020BalancedMF, Wu2020DistributionBalancedLF} weigh the loss function by class cardinality, usually assigning higher weights for loss terms dependent on minority classes. Representation learning methods~\cite{kang2021exploring, Yang2020RethinkingTV,dong2023lpt} focus on learning a more powerful feature encoder. For example,
\cite{Ma_2022_BMVC,Ma2021ASL,dong2023lpt} leverage the generalizable knowledge from large foundation models~\cite{clip}. 
For a  more detailed review, please see the recent survey papers~\cite{zhang2021tricks, Yang2022ASO,Zhang2021DeepLL}. While prior long-tailed works focus on image classification, we are the first to investigate the long-tailed setting for AD.

\section{Long-tailed anomaly detection} 

\noindent\textbf{Motivation:} Previous AD works assume that different image classes are equally populated. However, in most industrial applications, different objects have different costs, production schedules, \etc. 
This creates long-tailed distributions where certain classes have much higher example cardinality than others. Extensive research in areas like classification, indicates that systems not trained to account for this class imbalance tend to overfit on popular classes and ignore the less popular ones~\cite{Kang2019DecouplingRA,Krishnan2018AnAA,Li2021MetaSAugMS,Cui2019ClassBalancedLB,Kang2019DecouplingRA,Ren2020BalancedMF, Wu2020DistributionBalancedLF}. To test whether this also holds for AD, we perform some preliminary experiments, using the MVTec dataset and a long-tailed version, obtained by image resampling. The sample distributions of the two datasets are shown in  Fig.~\ref{fig:prelim_study}(a). Fig.~\ref{fig:prelim_study}(b)  compares the performance on the two datasets of UniAD~\cite{uniad}, one of the best open-source AD methods based on a single model that detects multi-class anomalies. The figure confirms that UniAD performs well on MVTec but 
degrades considerably for the long-tailed version, where it overfits to majority classes and severely underperforms on minority ones. This experiment highlights the need for the methods that explicitly address the long-tailed AD problem. This requires datasets and performance metrics, which we discuss next. 

\begin{table}
    \centering
    \resizebox{.77\columnwidth}{!}{%
    \begin{tabular}{ccc}
      \toprule
      Dataset   & Max Class Sample & Imbalanced Factor $\beta$ \\
      \midrule
      MVTec~\cite{mvtec}   &  391  &  \{100, 200\}\\
      VisA~\cite{visa}    &  905  &  \{100, 200, 500\}\\
      DAGM~\cite{dagm}    &  1000  &  \{50, 100, 200, 500\}\\
      \bottomrule 
    \end{tabular}
    }
    \caption{The statistics of the long-tailed splits we use across the 3 datasets. For all datasets, we consider both the \e~and \s~imbalance. 
    }
    \label{tab:dataset_stats}
    \vspace{-10pt}
\end{table}

\noindent\textbf{Dataset collection:} 
Following ~\cite{cao2019learning}, given a balanced dataset, we create a long-tailed version by sampling the training set, while the test set remains unchanged. The distribution of the sampled training set depends on two factors: the imbalance factor $\beta$ and the imbalance type. $\beta$ is the ratio between the cardinalities of the most and least populated classes, \ie, $\beta = \frac{\max_c\{N_c\}}{\min_c \{N_c\}}$ where $N_c$ is the number of samples of class $c$. 
To prevent overfitting on a specific imbalance distribution, we consider two types of imbalance: exponential (\e) and step (\s). While the former indicates that $N_c$ decays exponentially across classes $c$, the latter indicates a binary split into majority classes of size $\max\{N_c\}$ and minority classes of size $\min\{N_c\}$. 
Individual long-tailed datasets are denoted by type and imbalance factor. For example, \e{100} indicates a training set exponentially imbalanced with $\beta=100$. We define the half most/least populated classes as majority/minority classes and construct the long-tailed dataset by randomly sampling from the original dataset, without repetition. When the number of samples of a class is less than the desired number of samples, all samples are kept. 

\noindent\textbf{Tasks and metrics:} Both anomaly detection (AD) and anomaly segmentation (AS) are considered. Following ~\cite{uniad,regad,anomalygpt,draem}, we use the Area Under the Receiver Operating Curve (AUROC) at image and pixel level for AD and AS, respectively. We report two types of results. The first is average performance (Avg), which is the performance averaged across all classes.  The second is the pair of average performance for majority classes (High) and average performance for minority classes (Low). 

\noindent\textbf{Datasets:} We consider 3 datasets: MVTec~\cite{mvtec}, VisA~\cite{visa} and DAGM~\cite{dagm}. Various long-tailed datasets are built from each. Tab.~\ref{tab:dataset_stats} shows the maximum number of samples across classes and the imbalance factors considered per dataset. For all datasets, we consider both the \e~and \s~imbalance types. Our proposed dataset splits will be released upon publication.

\vspace{-3pt}
\section{The \ours anomaly score} \vspace{-3pt}

\begin{figure}
    \centering
    \includegraphics[width=\linewidth]{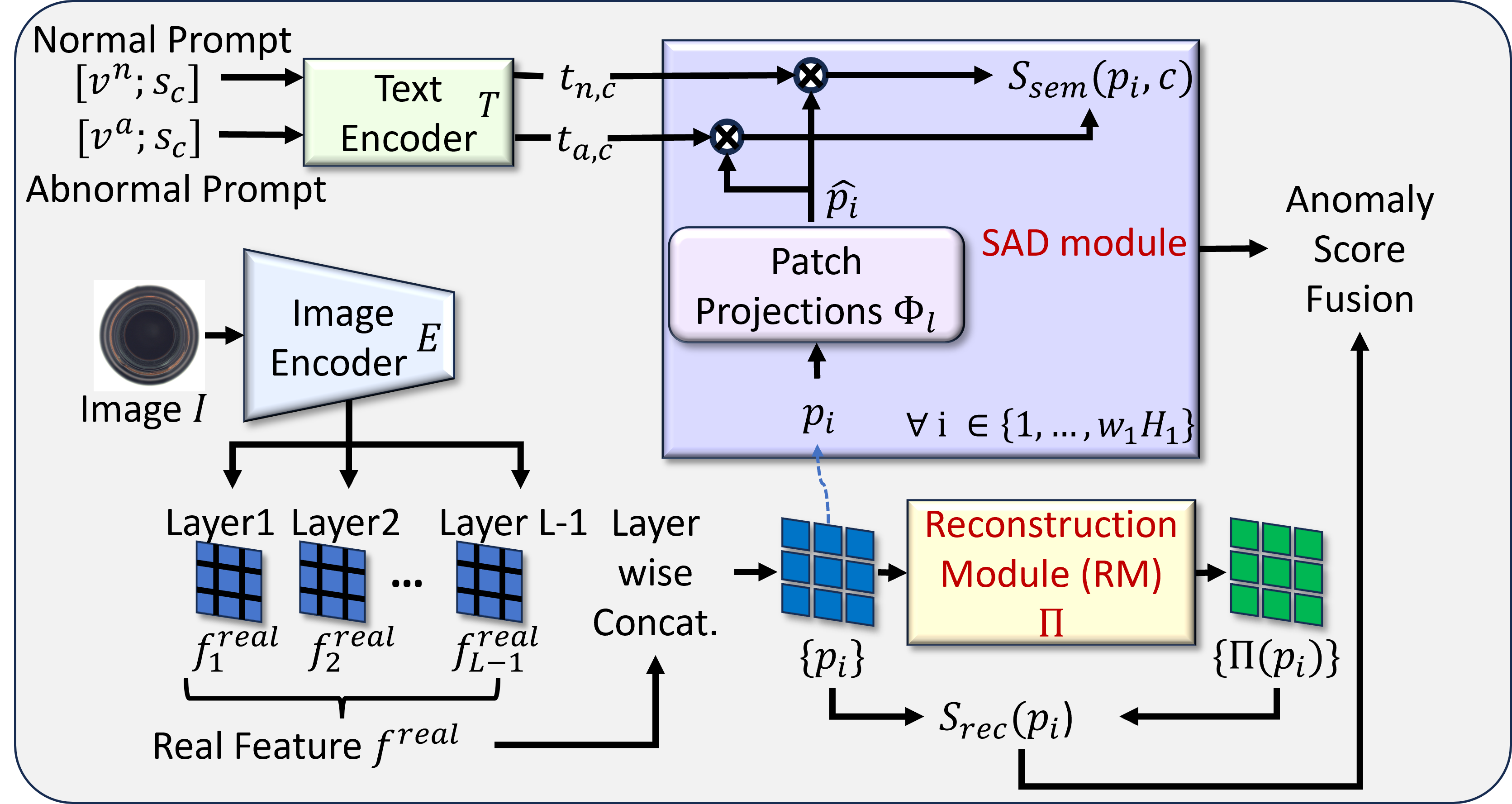}
    \caption{The \ours architecture combines AD by reconstruction and semantic AD scores (${\cal S}_{rec}$ and ${\cal S}_{sem}$, respectively), implemented by the RM and SAD modules. We use an image $E$ and text encoder $T$ from a pretrained foundation model to extract images features $f^{real}$ and text features $t_{n,c}, t_{a,c}$ derived from text-prompts that include a static component to discriminate between normal ($v^{n}$) and abnormal ($v^{n}$) and a learned component $s_c$ to make this discrimination class-sensitive.}
    \vspace{-10pt}
    \label{fig:main_inference}
\end{figure}

In this section, we propose a novel method, \ours, for the unsupervised long-tailed anomaly detection task. 
As shown in Fig.~\ref{fig:main_inference}, 
\ours uses a combination of (1) AD by reconstruction and (2) semantic AD. AD by reconstruction is a common AD approach, where a model is trained to reconstruct normal images~\cite{Akay2018GANomalySA,Bergmann2018ImprovingUD,Collin2020ImprovedAD,Liu2019TowardsVE,sabokrou2018adversarially, uniad,Zaheer2020OldIG}. At inference, this model projects abnormal images into the normal image manifold. AD can thus be implemented by thresholding the magnitude of the reconstruction error. Semantic AD explicitly trains a classifier to discriminate between normal and abnormal images. This is less commonly used, since abnormal images are not available in the training set. \ours overcomes this challenge by leveraging the understanding of the ``abnormal" concept by the pretrained ALIGN~\cite{align} foundation model and a learned semantic descriptor for the classes in the training set. The implementation of these modules is as follows.

\noindent{\bf AD by reconstruction.}
The reconstruction module (RM) $\Pi(.)$ is a transformer~\cite{vtadl,uniad,Pirnay2021InpaintingTF,AnoViT} trained to reconstruct the features extracted from the image $I$ 
by a pretrained encoder $E$ of $L$ layers. 
Given an image $I \in \RR^{W \times H \times 3}$ from class $c \in \cC$, $E$ extracts feature tensor $f_l^{real} \in \RR^{W_l \times H_l \times C_l}$ from layer $l \in \{1 \dots L\}$. 
Since the feature tensor from the last layer $f_L^{real}$ represents the global semantics of $I$, it tends to degrade the AD performance ~\cite{uniad, regad, patchcore}, which requires local semantics. Hence, to perform AD, this tensor is dropped and the first $L-1$ tensors $\{f_l^{real}\}_{l=1}^{L-1}$ are remapped to the dimensions of $f_1^{real}$ (\ie, $W_1 \times H_1$) by bilinear interpolation along the spatial dimension. In the following, we use the notation $f_l^{real}$ to represent this interpolated version and define $f^{real}= [f_1^{real} ; \dots ; f_{L-1}^{real}]$ as the feature  tensor extracted across the $L-1$ layers. 
This tensor is then split into ${W_1 \times H_1}$ patch feature  vectors $\{p_i\}_{i=1}^{W_1 \times H_1}$, which are fed as tokens to the RM transformer $\Pi(.)$. Given patch $i$, the anomaly score of the AD by reconstruction  module  is the squared error
\begin{equation}
    \cS_{rec}(p_{i}) = ||\Pi(p_i) -  p_i||^2.
    \label{eq:pta_rm}
\end{equation}

\noindent{\bf Semantic AD (SAD).} The goal of semantic AD is two-fold: 1) to give the anomaly detector sensitivity to normal/abnormal classes, and 2) to leverage the prior 
knowledge about normality/abnormality available in a large foundation  model. This allows the AD to discriminate between the two conditions without requiring abnormal images for training. As shown in Fig.~\ref{fig:main_inference}, the semantic AD module is a binary classifier of a projection $\hat{p}_{i}$ of patch $p_i$ into normal/abnormal classes. The layer-wise components $p_{il}$ of the patch feature vector $p_i$ are first projected into vectors $\Phi_l(p_{il})$ with the dimension $d$ of the text embedding of the ALIGN model. 
These projections are implemented by projection modules $\Phi_l:\RR^{C_l}\to\RR^{d}, \ l=1, \ldots, L-1$, where each $\Phi_l$ is implemented with a linear layer. The layer-wise features are then aggregated into a single patch feature vector 
\begin{equation}
    \hat{p}_{i} = \max_l (\{\Phi_l(p_{il})\}_l) \label{eq:hatp}
\end{equation}
by max pooling over layers.
The resulting vector $\hat{p}_{i}$ is then fed to a binary classifier of parameters $t_{n,c}$ (normal) and $t_{a,c}$ (abnormal), where $c$ is the class of image $I$, which computes the posterior probability of an anomaly using a softmax layer with temperature scaling $\tau$ 
\begin{equation}
    \cS_{sem}(p_{i}, c) = \frac{\exp(t_{a,c} \ \bigcdot \  \hat{p}_{i}/\tau)}{\exp(t_{n,c} \ \bigcdot \ \hat{p}_{i}/\tau) + \exp(t_{a,c} \ \bigcdot \  \hat{p}_{i}/\tau)},
    \label{eq:pta_as}
\end{equation}
where ``$\bigcdot$" denotes the dot product.
This is used as the semantic AD score for the images of class $c$. 

The main difficulty of this process is to learn the classifier parameters $t_{n,c}, \ t_{a,c}$ without explicit supervision, since there are no training images of anomalies. To overcome this problem we leverage the prior for normal/abnormal classification provided by the ALIGN model. 
This is implemented by feeding to ALIGN a normal text prompt $v_{n}$ and an abnormal text prompt $v_{a}$ that apply to all classes. While we have considered several possibilities (see Tab.~\ref{tab:prompt_ablate}), the best AD performance was achieved by setting $v_{n}=$``a" and $v_{a}=$``a broken." Unless otherwise noted, we use these prompts in what follows.
To further make the anomaly score sensitive to the image semantics, this is complemented by an image class prompt $s_c$. Unlike prior works~\cite{winclip,sam_ad,april_gan} that assume the access to the class names (\eg, ``bottle," or ``hazelnut" in MVTec~\cite{mvtec}), we assume that the class name is unknown. This is important to support the classes that are unknown to the ALIGN model or even to most humans, such as ``PCB1" vs. ``PCB2" in Fig.~\ref{fig:teaser}. Instead, inspired by \cite{gal2023an}, we use a {\it pseudo\/} class name $s_c$ learned per class $c$. This is implemented by prompting the text encoder $T$ with a prompt $s_c$ per class $c$, and learning prompts $s_c$ as discussed below. 
The resulting set of {\it semantic sensitive AD prompts} $\cP = \{[v^n; \ s_c], \ [v^a; \ s_c]\}_c$ is mapped to a set of classifier parameters $\{(t_{n,c}, \ t_{a,c})\}_c$ by the text encoder $T$ of the ALIGN model, according to
\begin{equation}
    t_{n,c} = T([v^n; \ s_c]) \quad \quad t_{a,c} = T([v^a; \ s_c]). 
    \label{eq:ts}
\end{equation}

\noindent{\bf \ours score:} The overall anomaly score of a patch feature $p_i$ of class $c$ is defined as the linear combination
\begin{equation}
    \cS(p_{i},c) = \cS_{rec}(p_{i}) + \lambda \cS_{sem}(p_{i},\ c),
    \label{eq:as_fusion}
\end{equation}
where $\lambda$ is the hyperparameter to balance ${\cal S}_{rec}(p_{i})$ and ${\cal S}_{sem}(p_{i},\ c)$ such that both scores have comparable ranges.

\section{Training} 

\begin{figure}
    \centering
    \includegraphics[width=\linewidth]{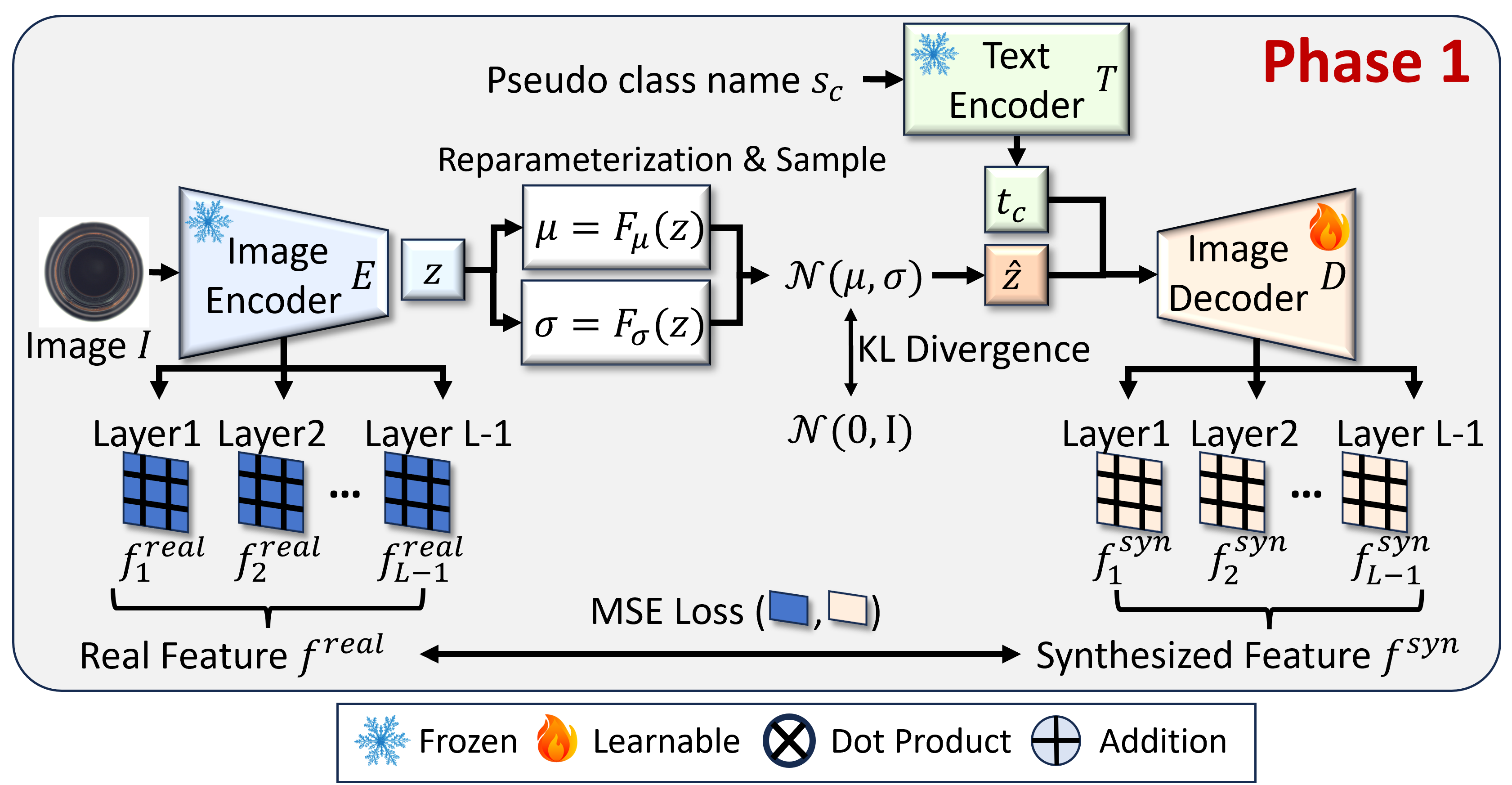}
    \caption{Phase 1 of \ours training learns a VAE-style decoder $D$ for feature augmentation conditioned on a learned pseudo class name $s_c$.
    }
    \vspace{-10pt}
    \label{fig:main_phase1}
\end{figure}

The training of \ours is divided into two phases. 
\subsection{Phase 1: Class sensitive data augmentation}
This training phase seeks two goals: to 1) overcome the data scarcity of long-tailed AD, by augmenting the training set with normal examples of minority classes and abnormal examples of all classes, 2) learn the class sensitive prompts $s_c$ required by the semantic AD score of~(\ref{eq:pta_as})-(\ref{eq:ts}).   

Fig.~\ref{fig:main_phase1} 
summarizes this training procedure. 
Given an image $I \in \RR^{W \times H \times 3}$ of class $c \in \cC$, the pretrained encoder $E$ extracts the feature tensor $f^{real}= [f_1^{real} ; \dots ; f_{L-1}^{real}]$ plus a latent code $z=f_{L}^{real}$, which is the feature vector from the last encoder layer ($L$).
An image decoder $D$, whose architecture is the mirror copy of $E$, is then trained to sample corresponding feature vectors, using a procedure inspired by the variational autoencoder (VAE)~\cite{vae, vae_intro}.
A latent feature $\hat{z}$ is sampled from a normal distribution  $\cN(\mu,\ \sigma)$ of parameters $\mu = F_\mu(z)$ and $\sigma = F_\sigma(z)$, where $F_\mu$ and $F_\sigma$ are learned linear transformations. The decoder $D$ then synthesizes a feature tensor from $\hat{z}$.

In the long-tailed setting, the performance of $D$ degrades for classes $c$ with few training images. To ameliorate this problem, $D$ is conditioned by the prior knowledge about the class, in the form of a text-derived prototype feature vector $t_c$ that represents class $c$ for feature synthesis. This is obtained by prompting the text encoder of ALIGN with the pseudo-class name $s_c$,
\ie, $t_c=T(s_c)$. The feature prototype is then concatenated with the image-dependent latent feature $\hat{z}$ to create the input to $D$, which finally synthesizes a feature tensor $\{f_l^{syn}\}_{l=1}^{L-1} = D(\hat{z} , t_c )$ of dimensions equal to those of $f_l^{real}$. 

Following standard practices for VAE training, $D$ and $s_c$ are learned by optimizing a loss function 
\fontsize{9.8pt}{19.6pt}
\begin{equation}
    \cL_{\PP_1} = \frac{1}{L-1}\sum_{l=1}^{L-1} || f_l^{syn} - f_l^{real}||^2 - KL(\cN(\hat{z} - \mu,\ \sigma) || \cN(0,\ I)),
\end{equation}
\normalsize
that combines the reconstruction mean square error (MSE), a regularization constraint based on the Kullback-Leibler divergence ($KL$) that encourages a normal distribution, and the reparametrization trick of~\cite{vae}. The text $T$ and image $E$ encoders are those of the pretrained ALIGN model and kept frozen throughout training. Note that this process encourages the simultaneous satisfaction of multiple goals: 1) learning a decoder that can be used to synthesize features from the tail classes, 2) align these features with the semantic representation $t_c$ produced by the text encoder of ALIGN, and 3) improve the quality of feature synthesis for tail classes, by leveraging this alignment. After training, the learned prompts $s_c$ are used in (\ref{eq:ts}).

\subsection{Data augmentation} 
When this training phase is completed, the decoder $D$ works as a data augmentation device, producing synthetic feature tensors $f^{syn}$ in the semantic neighborhood of a feature tensor $f^{real}$ extracted from a real image. This is used to augment the training data in an online fashion during the second phase of training. Two types of data augmentation are considered.

\noindent{\bf Long-tailed classes.} To counteract the imbalanced nature of long-tailed datasets, data augmentation is implemented by selecting the real $f^{real}$ or synthetic $f^{syn}$ feature vectors with probabilities $(p_c,\ 1-p_c)$ respectively. 
The selected feature vector $f$ is split into $W_1 \times H_1$ patch feature vectors $\{p_{i}^n\}_{i=1}^{W_1 \times H_1}$ where the $n$ superscript denotes that these are normal features.

\begin{figure}
    \centering
    \includegraphics[width=\linewidth]{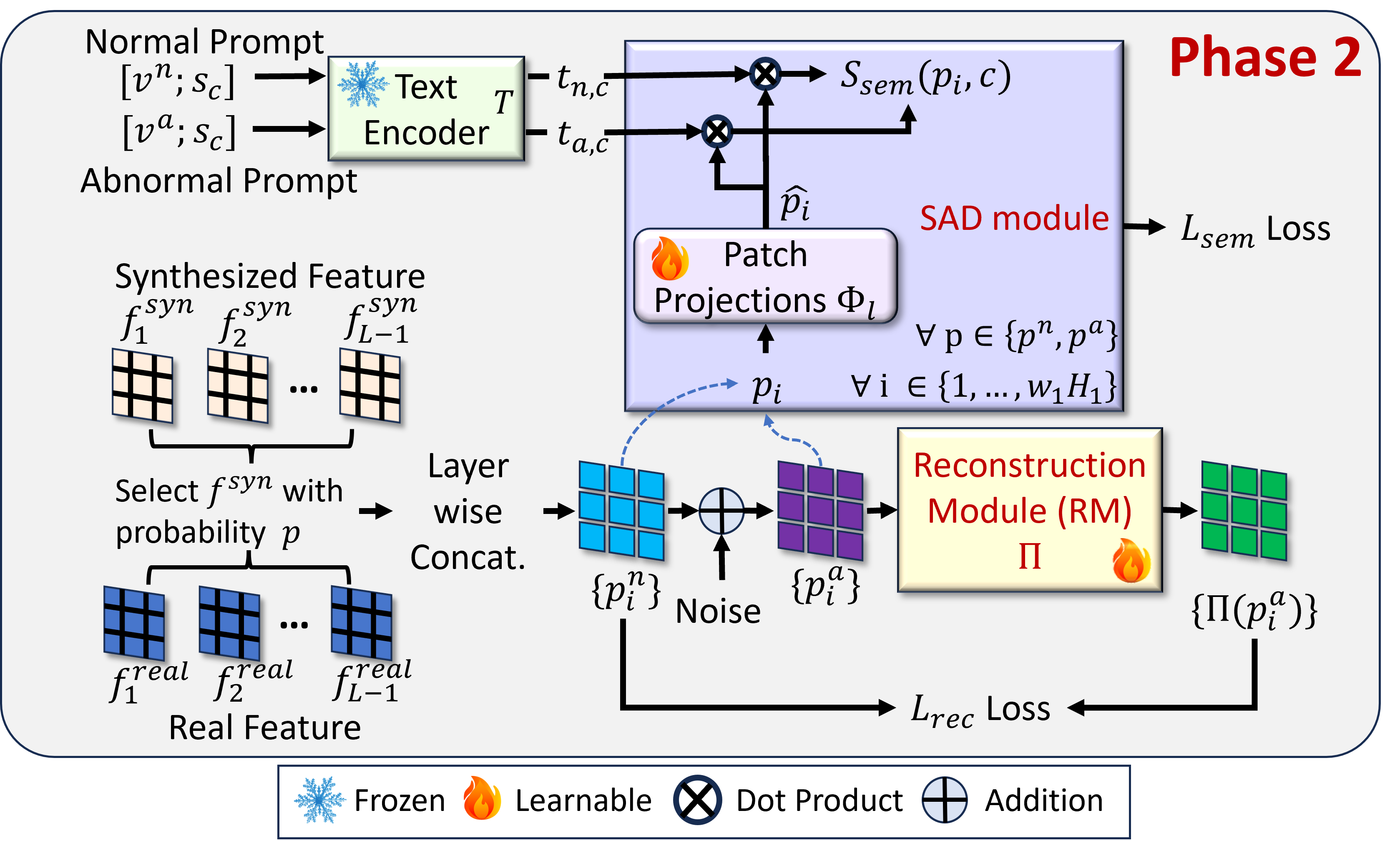}
    \caption{Phase 2 of \ours training learns the parameters of the reconstruction module (RM) and patch projections $\Phi_l$ that map visual features into the semantic space of the semantic AD (SAD) module.
    }
    \vspace{-10pt}
    \label{fig:main_phase2}
\end{figure}

\noindent{\bf Anomalies.} To counteract the lack of anomalies during training, random noise (sampled from normal distribution) is added to  normal patch features $p_{i}^n$ to produce pseudo-anomaly patch features $p^a_{i}$, as in \cite{uniad}. This process is repeated for all normal patches during training. No random noise is added during inference.

\subsection{Phase 2: Anomaly detection} 
Data augmentation is used in the second phase of training to learn the parameters of 1) the RM transformer $\Pi(.)$ used to reconstruct features and 2) the modules $\Phi_l(.)$ used in (\ref{eq:hatp}) to project patch features into the semantic space of ALIGN. 

\paragraph{Reconstruction Module (RM):}
As shown in the bottom right of Fig.~\ref{fig:main_phase2}, the RM $\Pi(.)$ is trained to project the pseudo anomaly patch features $p^a_i$ into the reconstructed patch features $\Pi(p^a_i)$ in the manifold of the normal patch features $p^n_i$. This is implemented with the RM transformer to  minimize the loss
\begin{equation}
    \cL_{rec} = \frac{1}{W_1 H_1}\sum_{i=1}^{W_1 H_1} ||\Pi(p^a_i) -  p_i^n||^2.
\end{equation}

\begin{table*}

    \centering
 \begin{minipage}{0.7\linewidth}
    
    \centering
    \resizebox{\linewidth}{!}{
    \begin{tabular}{cc>{\columncolor{gray!10}}c>{\columncolor{gray!10}}c>{\columncolor{gray!10}}c>{\columncolor{gray!10}}c>{\columncolor{gray!10}}c>{\columncolor{gray!10}}c>{\columncolor{red!10}}c>{\columncolor{red!10}}c}
    \toprule
     Config. &  Task & Cut \& Paste & MKD & DRAEM & RegAD & UniAD & AnomalyGPT & \ours w/o SAD & \ours \\
     \midrule
     \multirow{2}{*}{\e{100}} & Det. & 75.89 & 78.92 &  79.57 & 82.43 & 87.70 & 87.44 & \underline{88.74} & \textbf{88.86}\\
     & Seg. & N/A & 85.95 & 85.17 & \textbf{95.20} & 93.95 & 89.68 & 94.00 & \underline{94.46} \\
     \cmidrule(l){1-10}
     \multirow{2}{*}{\e{200}} & Det. & 75.07 & 79.93 & 78.82 & N/A & \underline{86.21} & 85.80 & \textbf{86.94} & 86.05 \\
     & Seg. & N/A & 86.01 & 82.95 & N/A & 93.26 & 90.15 & \underline{93.40} & \textbf{94.18}\\
     \midrule
     \multirow{2}{*}{\s{100}} & Det. & 76.57 & 79.61 & 69.82 & 81.54 & 83.37 & 85.95 & \underline{87.05} & \textbf{87.36} \\
     & Seg. & N/A & 85.90 & 79.65 & \textbf{95.10} & 91.47 & 89.28 & 93.13 & \underline{93.83} \\
     \cmidrule(l){1-10}
     \multirow{2}{*}{\s{200}} & Det. & 76.53 & 79.31 & 71.64 & N/A & 81.32 & 82.47 & \underline{85.33} & \textbf{85.60 }\\
     & Seg. & N/A & 86.03 & 76.79 & N/A & 89.29 & 89.45 & \underline{91.78} & \textbf{92.12} \\
     \bottomrule   
    \end{tabular}
    }
    \captionof{table}{Quantitative comparison on MVTec dataset. 
    }
    \vspace{-5pt}
    \label{tab:mvtec}
\end{minipage}
\hspace{3pt}
\begin{minipage}{0.24\linewidth}
\centering
\includegraphics[width=0.95\linewidth]{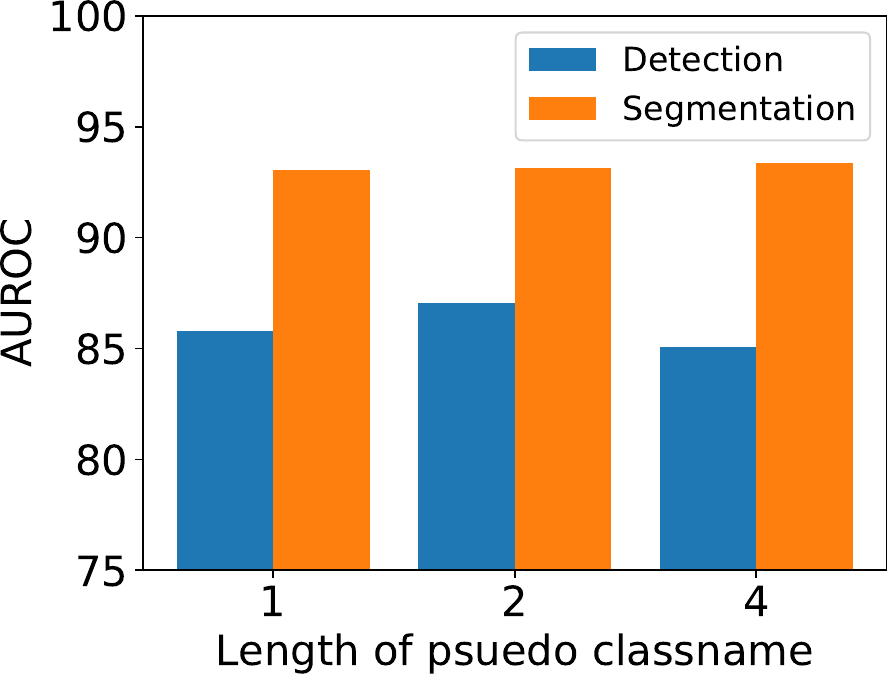}
\captionof{figure}{Ablation on the length of pseudo class name.}
\vspace{-15pt}
\label{fig:psuedo_classname_size}
\end{minipage}
\end{table*}

\paragraph{Semantic patch projections:} As shown in the top right of Fig.~\ref{fig:main_phase2}, the functions $\Phi_l$ compute projections of the patches $p_i$ into the semantic space of ALIGN, where the classifier parameters of (\ref{eq:ts}) are defined. The functions $\Phi_l$  are trained to encourage the alignment between the projected patch features and the text features by minimizing the binary cross entropy loss
\fontsize{9.8pt}{19.6pt}
\begin{equation}
    {\cal L}_{sem}(c) = \frac{-1}{W_1 H_1} \sum_{i=1}^{W_1 H_1} y_i \log(\cS_{sem}(p_{i},\ c)), \ y_i = \begin{cases} 1, \ \text{if} \ p_{i} = p^a_{i} \\ 0, \ \text{if} \ p_{i} = p^n_{i}, \end{cases}
\end{equation}
\normalsize
where $c$ is the image class and ${\cal S}_{sem}(.)$ is the semantic score of~(\ref{eq:pta_as}). Note that 
both phase 1 and phase 2 share the same text encoder $T$, which is fixed in both phases. The total loss function for phase 2 is $\cL_{\PP_2} = \cL_{rec} + \cL_{sem}(c)$.

\section{Experiments}

In this section, we report on various experiments designed to evaluate \ours. While we compute average performance across majority (High), minority (Low), and all (All) classes in all experiments, we omit the High and Low values in some cases, for brevity. A complete listing is provided in the supplement.

\noindent\textbf{Baselines:} For a given training configuration (\eg, \e{100}), the same training set is used for all the baselines~\cite{cutandpaste, draem,mkd, anomalygpt, uniad} and \ours. For baselines other than RegAD~\cite{regad}, their official code and training/testing setting are used. RegAD is trained on all classes, instead of its leave-one-class-out setting, for fair comparison. During testing, we use 2 training examples per class as the support set to estimate the normal distribution. Since RegAD requires a support set of 2 images per class, it is not applicable to some configurations (\eg, \e{200} and \s{200} in MVTec) where only 1 image is available for some minority classes. Cut \& Paste~\cite{cutandpaste} supports anomaly detection but not localization.

\noindent\textbf{Training details:} To train \ours, we use the pretrained visual language foundation model ALIGN~\cite{align}. Each input image is scaled to 224$\times$224 and the features $f^{real}$ are extracted from layers $\{3,\ 10,\ 17,\ 37\}$ of the ALIGN image encoder. 
By default, the length of the pseudo class name is set to 2 and initialized with the text ``object object." In phase 2, the probability $p_c$ of selecting real features is 0.5. 
The hyperparameter $\lambda$ in (\ref{eq:as_fusion}) is set to 500, 400, and 300 for MVtec, VisA, and DAGM, respectively.
Unless otherwise noted, we use $v^n=$ ``a" and $v^a=$ ``a broken" in (\ref{eq:ts}) and $\tau=1$ in (\ref{eq:pta_as}).
Phase 1 is trained for 100 epochs and phase 2 for 500 epochs, both using the AdamW optimizer with the learning rate 1e-4. Pytorch~\cite{pytorch} is used for implementation.

\begin{table*}[]
    \centering
 \begin{minipage}{0.49\linewidth}
 
 \adjustbox{max width=\linewidth}{
    \begin{tabular}{cc>{\columncolor{gray!10}}c>{\columncolor{gray!10}}c>{\columncolor{gray!10}}c>{\columncolor{red!10}}c>{\columncolor{red!10}}c}
    \toprule
     Config. &  Task & RegAD & UniAD & AnomalyGPT & \ours w/o SAD & \ours \\ \midrule 
     \multirow{2}{*}{\e{100}} & Det. & 71.36 & 77.31 & 70.34 & \underline{79.27} & \textbf{80.00} \\
     & Seg. & 94.40 & 95.03 & 80.32 & \underline{95.07} & \textbf{95.56}\\
     \cmidrule(l){1-7}
     \multirow{2}{*}{\e{200}} & Det. & 72.10 & 76.87 & 69.78 & \underline{78.55} & \textbf{80.21} \\
     & Seg. & 94.69 & \underline{94.80} & 79.48 & 94.51 & \textbf{95.36}\\
     \cmidrule(l){1-7}
     \multirow{2}{*}{\e{500}} & Det. & N/A & 73.67 & 68.18 & \underline{77.25} & \textbf{78.53} \\
     & Seg. & N/A & \underline{94.35} & 78.83 & 94.04 & \textbf{94.66}\\
     \midrule
     \multirow{2}{*}{\s{100}} & Det. & 71.80 & 78.83 & 71.98 & \underline{82.80} & \textbf{84.80} \\
     & Seg. & 94.99 & 96.04 & 82.30 & \underline{96.16} & \textbf{96.57}\\
     \cmidrule(l){1-7}
     \multirow{2}{*}{\s{200}} & Det. & 71.65 & 77.64 & 69.78 & \underline{83.79} & \textbf{84.03} \\
     & Seg. & 94.52 & 95.66 & 81.97 & \underline{95.89} & \textbf{96.27}\\
     \cmidrule(l){1-7}
     \multirow{2}{*}{\s{500}} & Det. & N/A & 71.84 & 62.88 & \underline{82.42} & \textbf{83.33} \\
     & Seg. & N/A & 95.03 & 81.48 & \underline{95.50} & \textbf{96.41}\\
     \bottomrule     
    \end{tabular}
    \vspace{-5pt}
}
\captionof{table}{Quantitative comparison on VisA dataset.}
\label{tab:visa}
 \end{minipage}%
 \hspace{3pt}
  \begin{minipage}{0.49\linewidth}
  
 \adjustbox{max width=\linewidth}{
    \begin{tabular}{cc>{\columncolor{gray!10}}c>{\columncolor{gray!10}}c>{\columncolor{gray!10}}c>{\columncolor{red!10}}c>{\columncolor{red!10}}c}
    \toprule
     Config. &  Task & RegAD & UniAD & AnomalyGPT & \ours w/o SAD & \ours \\
     \midrule
     \multirow{2}{*}{\e{100}} & Det. & 84.86 & 84.34 & 85.31 & \underline{93.35} & \textbf{94.40} \\
     & Seg. & 90.29 & 90.13 & 77.20 & \underline{96.93} & \textbf{97.30}\\
     \cmidrule(l){1-7}
     \multirow{2}{*}{\e{200}} & Det. & 84.86 & 83.56 & 83.29 & \underline{92.83} & \textbf{94.29} \\
     & Seg. & 90.29 & 89.73 & 77.16 & \underline{96.16} & \textbf{97.19}\\
     \cmidrule(l){1-7}
     \multirow{2}{*}{\e{500}} & Det. & 84.86 & 81.35 & 83.47 & \underline{92.08} & \textbf{93.54} \\
     & Seg. & 90.29 & 88.63 & 76.87 & \underline{95.99} & \textbf{97.01}\\
     \midrule
     \multirow{2}{*}{\s{100}} & Det. & 84.86 & 81.11 & 86.48 & \underline{91.94} & \textbf{93.97} \\
     & Seg. & 90.28 & 89.11 & 78.76 & \underline{96.38} & \textbf{97.07}\\
     \cmidrule(l){1-7}
     \multirow{2}{*}{\s{200}} & Det. & 84.86 & 80.33 & 84.73 & \underline{91.78} & \textbf{93.79} \\
     & Seg. & 90.29 & 89.07 & 78.29 & \underline{96.04} & \textbf{96.84}\\
     \cmidrule(l){1-7}
     \multirow{2}{*}{\s{500}} & Det. & 84.86 & 80.04 & 85.08 & \underline{91.82} & \textbf{92.78} \\
     & Seg. & 90.29 & 88.53 & 78.75 & \underline{95.64} & \textbf{96.65}\\
     \bottomrule
    \end{tabular}
    \vspace{-5pt}
}
\captionof{table}{Quantitative comparison on DAGM dataset.}
\label{tab:dagm}
 \end{minipage}
\vspace{-8pt} 
\end{table*}

\subsection{Comparisons to the state-of-the-art}
Tab.~\ref{tab:mvtec}-\ref{tab:dagm} summarize the AD and AS performance of all the methods on MVTec, VisA, and DAGM, respectively. The best and second best performances for each dataset configuration are highlighted in \textbf{bold} and \underline{underline}, respectively. On MVTec, \ours is compared to 6 baselines under 4 different configurations. As shown in Tab.~\ref{tab:mvtec}, early methods, such as Cut \& Paste~\cite{cutandpaste}, MKD~\cite{mkd}, and DRAEM~\cite{draem}, are not suitable to detect and localize defects across classes with a single model, leading to inferior performance. While more recent models~\cite{uniad, regad, anomalygpt} can detect defects across classes, they do not perform well across levels of  dataset imbalance. \ours is less affected by skewed distributions and outperforms most baselines in all tasks. The only exceptions are RegAD, which outperforms \ours in 2 of the 8 tasks, and UniAD in 1 of the 8 tasks. 
Tab.~\ref{tab:visa}-\ref{tab:dagm} show that the gains of \ours over the baselines are even larger for VisA and DAGM, where it achieves the best performance in all tasks. Overall, \ours outperforms the baselines on 29 out of the 32 (90.6\%) tasks defined by the 3 datasets.

\begin{figure*}
    \centering
    \setlength{\tabcolsep}{1.7pt}
    \begin{tabular}{|c|c|c|}
    \hline
      \includegraphics[width=0.32\linewidth]{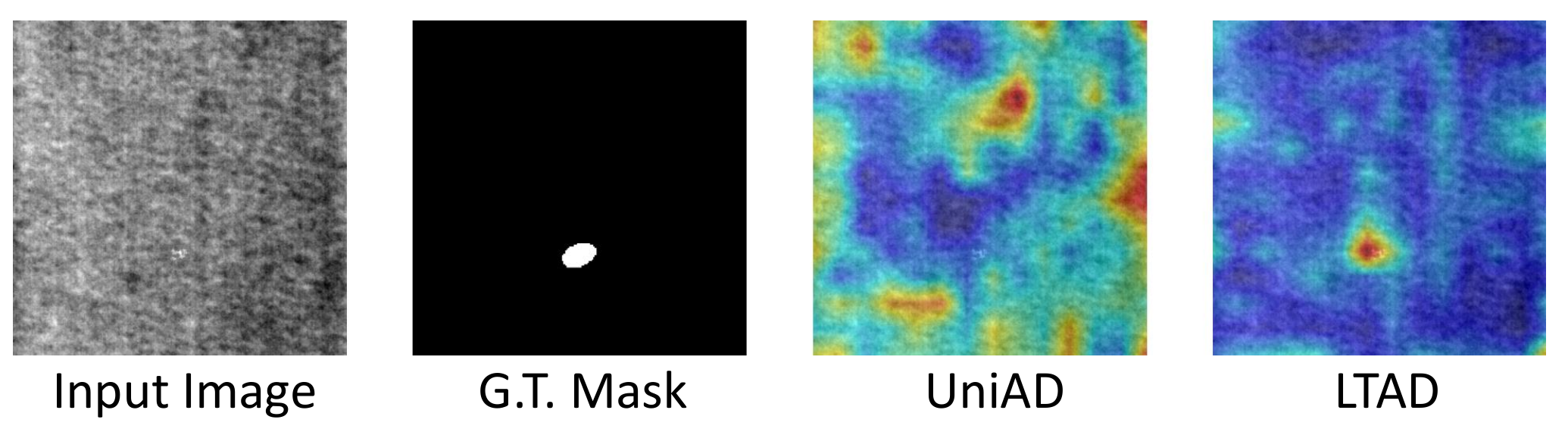}   & \includegraphics[width=0.32\linewidth]{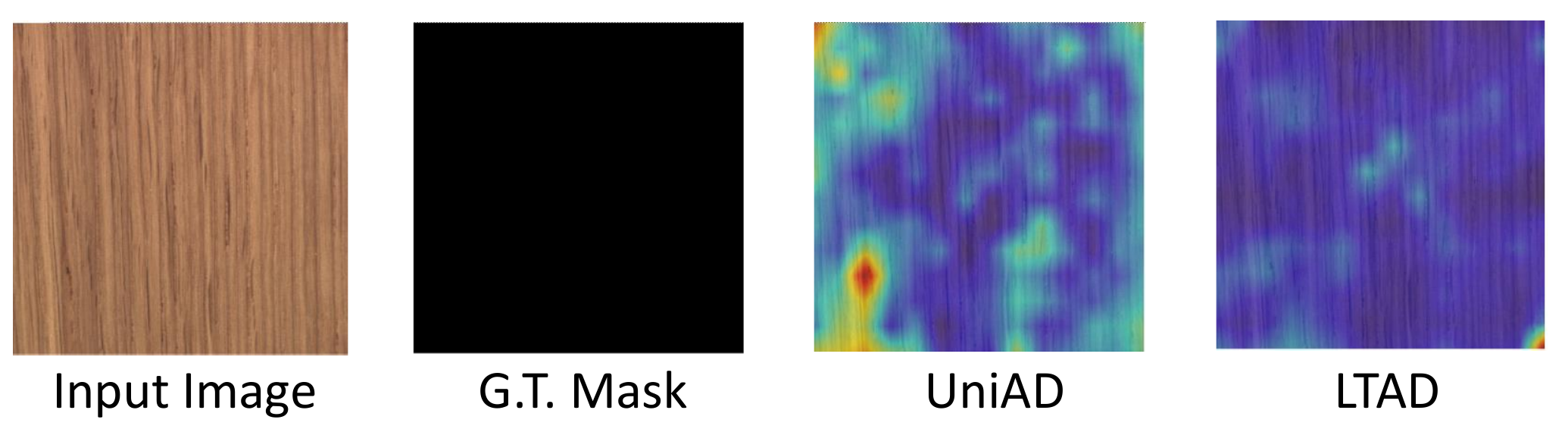} 
       & \includegraphics[width=0.32\linewidth]{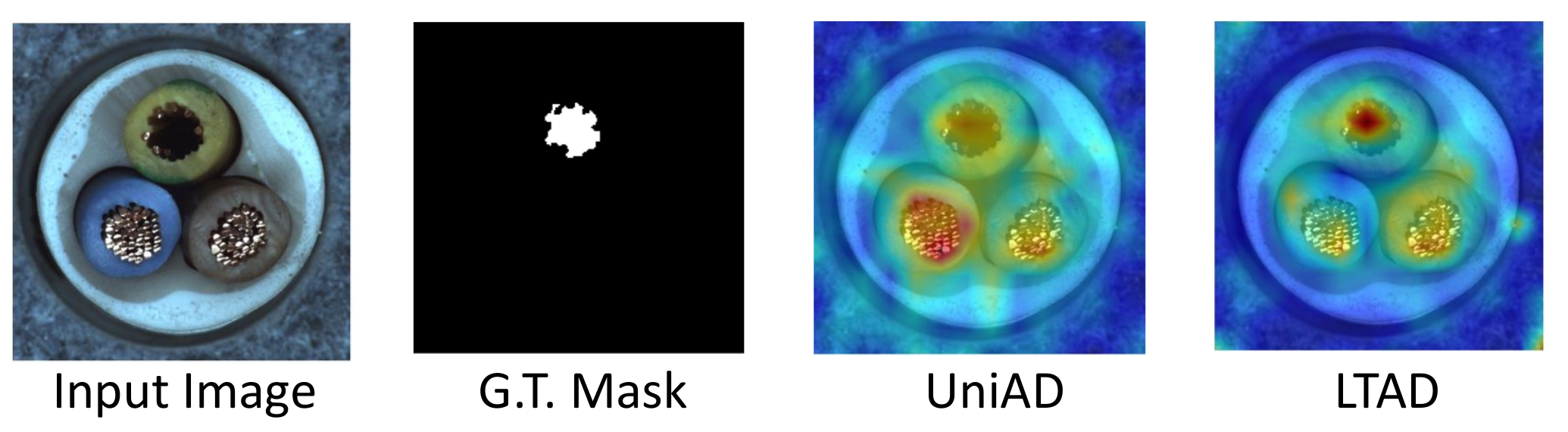} \\
           \hline
       \includegraphics[width=0.32\linewidth]{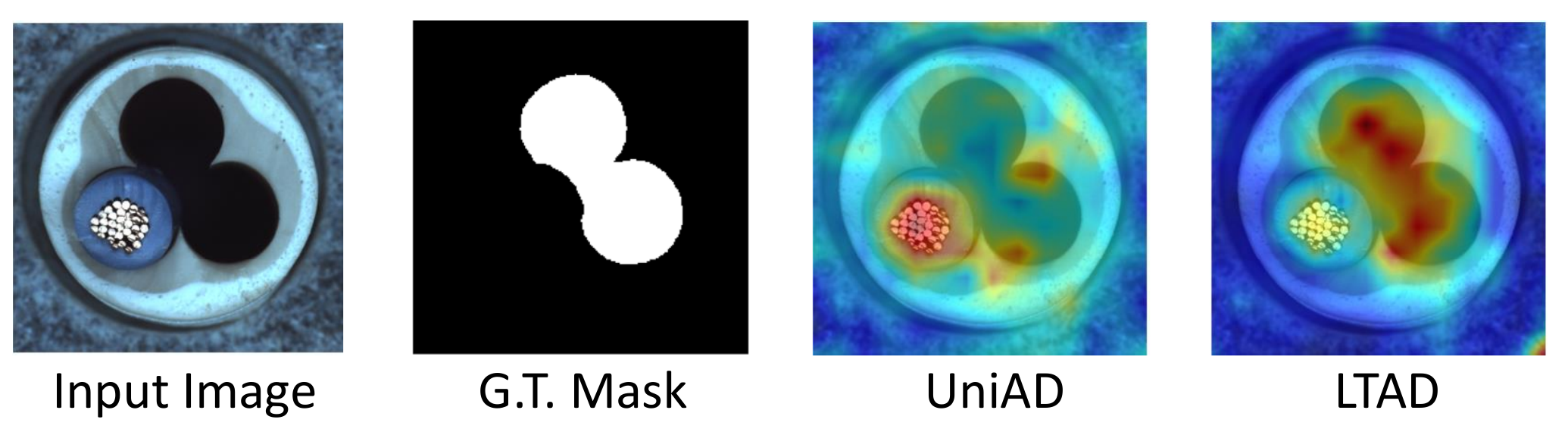}   & \includegraphics[width=0.32\linewidth]{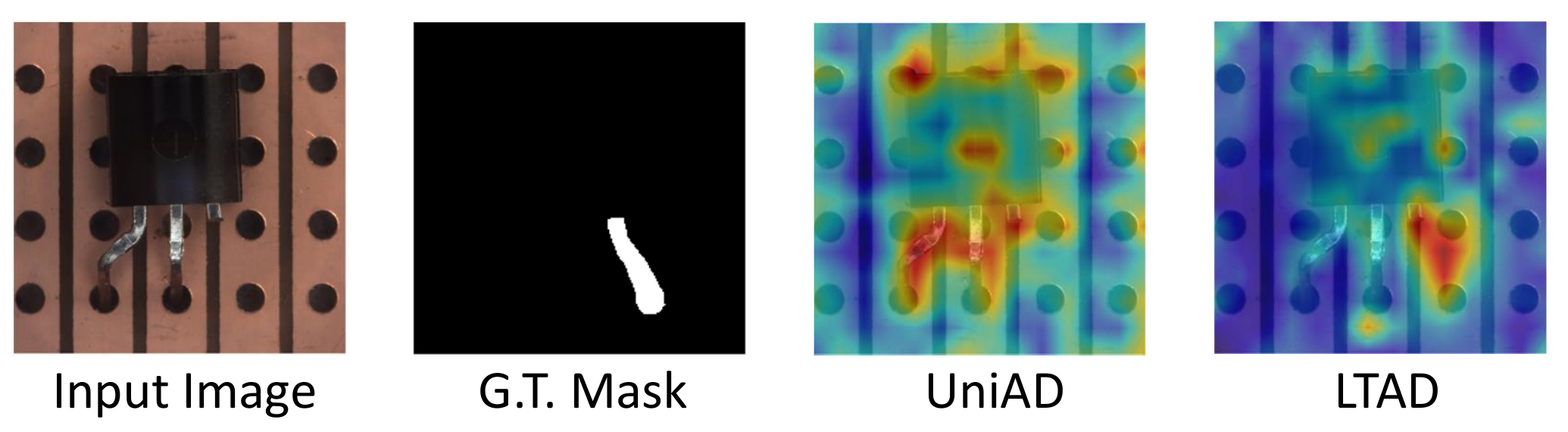} 
       & \includegraphics[width=0.33\linewidth]{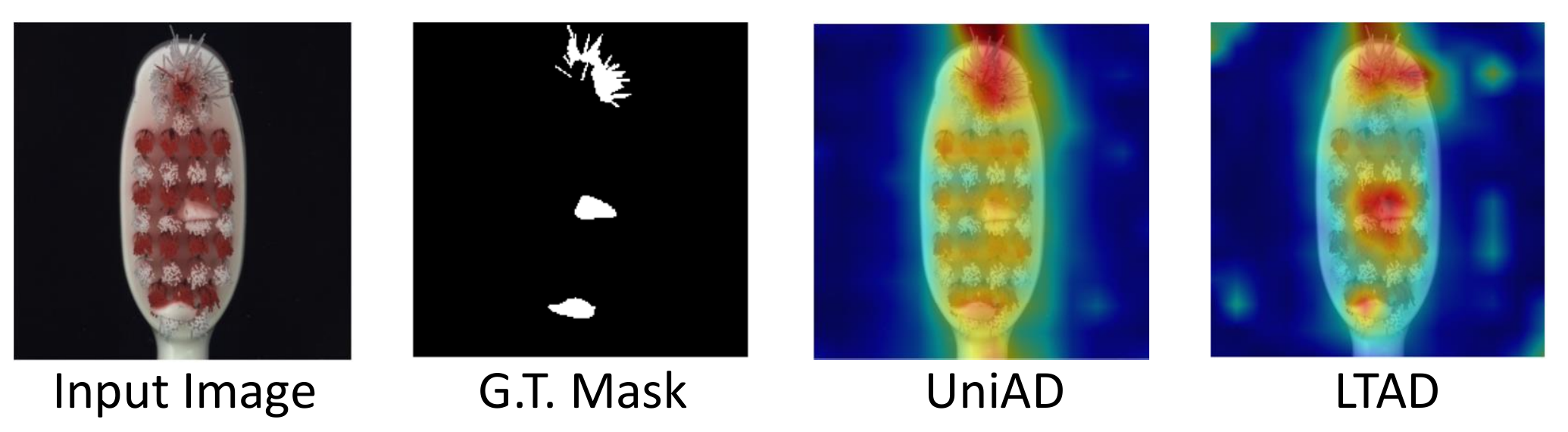}
       \\
        \hline 
    \end{tabular}
    
    \caption{Qualitative comparison of UniAD and \ours anomalies. The images without defects (\ie, normal image) have a black ground truth mask. 
    }
    \vspace{-10pt}
    \label{fig:vis}
\end{figure*}

Fig.~\ref{fig:vis} shows a comparative visualization of anomaly detections by UniAD and \ours. Note how the detections of \ours are much more localized and  selective of the anomaly. Additional visualizations are provided in the supplement.

Beyond \ours, the performance of \ours without the semantic AD module is shown in the penultimate columns of Tab.~\ref{tab:mvtec}-\ref{tab:dagm}. Even without this module, \ours beats the baselines on 28 out of 32 (87.5\%) tasks. However, by adding the semantic AD module, performance improves on 31 out of 32 (96.9\%) tasks, highlighting the efficacy of the semantic AD module.

\subsection{Ablation study}
We ablate different designs of \ours~in Tab.~\ref{tab:ablate_rm}-\ref{tab:classname_ablate}, where the experiments are denoted by the experiment ID (expID).

\noindent{\textbf{Ablation study of AD by reconstruction:}} 
Fig.~\ref{fig:psuedo_classname_size} and Tab.~\ref{tab:ablate_rm} compare different RM designs (\ie, \ours~without the semantic AD module) on the MVTec \s{100} dataset.  Fig.~\ref{fig:psuedo_classname_size} studies the effect of the length of the pseudo class name prompt on AD performance, showing that the performance peaks for length 2. 
Tab.~\ref{tab:ablate_rm} summarizes the detection and segmentation performance of several variants of \ours for majority (High), minority (Low) and all (All) classes. 
expID \ref{tab:ablate_rm}.2 first shows that replacing the EfficientNet~\cite{efficientnet} features of UniAD by those of ALIGN increases both AD and AS performance by 2.38\% and 1.44\%, respectively.  In expID \ref{tab:ablate_rm}.3, we replace the text conditioned VAE of \ours by an autoencoder trained on ALIGN features, to generate features for phase 1 of \ours training. This underperforms \ours (expID \ref{tab:ablate_rm}.8) by 2.57\% (0.6\%) for AD (AS), highlighting the efficacy of the VAE of \ours. In expID~\ref{tab:ablate_rm}.4, we replace the probability $p_c = 0.5$, 
used for synthetic feature selection in phase 2 of \ours, by a more sophisticated design, where $p_c$ is inversely proportional to the class cardinality (\ie, the classes with less samples have higher probability of selecting synthesized features).  When compared to \ours (expID \ref{tab:ablate_rm}.8), there is no benefit in adopting this distribution aware sampling (DAS) mechanism. 
The remaining experiments then ablate different possibilities for the class prompt $s_c$. 
Experiments expID~\ref{tab:ablate_rm}.5 and expID~\ref{tab:ablate_rm}.6 consider the case where the class names are available in the dataset. In expID~\ref{tab:ablate_rm}.5, the name of class $c$ is simply used as $s_c$. This underperforms all other experiments, where $s_c$ is a learned pseudo class name.
These experiments vary on the procedure used to initialize $s_c$ for learning. ExpID~\ref{tab:ablate_rm}.6, which uses the class name in the dataset as initialization, outperforms expID~\ref{tab:ablate_rm}.7, which randomly initializes $s_c$. Altogether, these results support the claim that, while the class specific prompts are important, not all the class names are informative and the prompts should be learned. Finally, all methods underperform \ours (expID~\ref{tab:ablate_rm}.8), which initializes all $s_c$ with the word ``object." This initialization also outperformed various other words that we have tried. The configuration of expID~\ref{tab:ablate_rm}.8 is used as the default for all other experiments in the paper.

\hypersetup{allcolors=black}
\begin{table}[]
    \centering
    \resizebox{\linewidth}{!}{
    \begin{tabular}{@{}c@{\hspace{.5em}}c@{\hspace{.3em}}c@{\hspace{.5em}}c@{\hspace{.5em}}c@{\hspace{.5em}}c@{}c@{\hspace{.5em}}c@{\hspace{.7em}}c@{\hspace{.7em}}c@{\hspace{.7em}}c@{\hspace{.7em}}c@{\hspace{.7em}}c@{}}
        \toprule
         \multirow{2.5}{*}{expID} &\multirow{2.5}{*}{method} &\multirow{2.5}{*}{\shortstack{use\\ALIGN}} &\multirow{2.5}{*}{\shortstack{AE or\\VAE}} &\multirow{2.5}{*}{\shortstack{use\\DAS}} &\multirow{2.5}{*}{\shortstack{learned\\$s_c$}} &\multirow{2.5}{*}{$s_c$ init.} &\multicolumn{3}{c}{Detection} & \multicolumn{3}{c}{Segmentation}  \\
         \cmidrule(l){8-10} \cmidrule(l){11-13}
         & & & & & & & All & High & Low & All & High & Low \\
        \midrule
        \ref{tab:ablate_rm}.1 &UniAD &\ccross &N/A &N/A &N/A &N/A & 82.63 & \textbf{99.60} & 67.79 & 91.47 & \textbf{96.15} & 87.38\\
        \ref{tab:ablate_rm}.2 &UniAD  &\ccheck &N/A &N/A &N/A &N/A & 85.01 & 99.39 & 72.43 & 92.91 & 95.99 & 90.20\\
        \ref{tab:ablate_rm}.3 &\ours &\ccheck &AE &\ccross &\ccross &N/A  & 84.48 & 99.13 & 71.67 & 92.53 & 94.77 & 90.57\\
        \ref{tab:ablate_rm}.4 &\ours &\ccheck &VAE &\ccheck &\ccheck &``object"  & 86.06 & 97.77 & 75.81 & 92.99 & 94.44 & \textbf{91.73}\\
        \ref{tab:ablate_rm}.5 &\ours &\ccheck &VAE &\ccross &\ccross & class name  & 85.39 & 99.11 & 73.39 & 92.57 &	94.56 & 90.82
        \\
        \ref{tab:ablate_rm}.6 &\ours &\ccheck &VAE &\ccross &\ccheck & class name  & 86.12 &	99.07 &	74.80 & 92.94 &	94.60 & 91.49 \\
        \ref{tab:ablate_rm}.7 &\ours &\ccheck &VAE &\ccross &\ccheck &random  & 85.95 & 99.01 & 74.53 & 92.92 & 95.07 & 91.03\\
        \ref{tab:ablate_rm}.8 &\ours &\ccheck &VAE &\ccross &\ccheck &``object" & \textbf{87.05} & 99.07 & \textbf{76.54} & \textbf{93.13} & 95.07 & 91.44\\
        \bottomrule
    \end{tabular}
    }
    \caption{Ablation Study without the SAD module on MVTec-\s{100}. Acronyms: DAS: distribution aware sampling; init.: initialization. 
    }
    \label{tab:ablate_rm}
    \vspace{-10pt}
\end{table}
\hypersetup{allcolors=red}
\hypersetup{citecolor=blue}

\noindent{\textbf{Ablation study of Semantic AD:}}
Tab.~\ref{tab:ablate_pta} ablates different SAD module designs (\ie, \ours~without the RM module) on the MVTec \s{100} dataset. ExpID~\ref{tab:ablate_pta}.1  replaces (\ref{eq:hatp}) by a direct projection of the concatenated vectors $p_i$ into the $d$-dimensional text space, using a linear transformation $\hat{\Phi}: \RR^{\sum_{l=1}^{L-1} C_l}\to\RR^{d}$.
When compared to \ours (ExpID~\ref{tab:ablate_pta}.4), this degrades AS performance significantly, most likely due to a greater difficulty in accounting for the different resolutions of features from different layers. The remaining experiments compare the implementation of (\ref{eq:hatp}) with different pooling operations, namely max vs. mean. Experiments expID~\ref{tab:ablate_pta}.2 and~\ref{tab:ablate_pta}.4 show that mean pooling is better than no layer aware projection, but inferior to the max pooling of \ours. Finally, expID~\ref{tab:ablate_pta}.3 uses max pooling and investigates the use of multiple word-prompts $v^n$ and $v^a$ in (\ref{eq:ts}). This is inspired by the gains reported for ensembling multiple prompts when visual language foundation models are used for open-set classification~\cite{clip}. Comparing experiments expID \ref{tab:ablate_pta}.3 and \ref{tab:ablate_pta}.4 shows that there is no similar advantage for semantic AD. 
Note that jointly using 2 anomaly scores on MVTec \s{100} (See Tab. \ref{tab:mvtec} \s{100}) outperforms using either the reconstruction by AD of expID \ref{tab:ablate_rm}.8 or the semantic AD of expID \ref{tab:ablate_pta}.4 solely, indicating both modules contribute to the gain.

Tab.~\ref{tab:classname_ablate} ablates the importance of the learned pseudo class name $s_c$ for the semantic AD module, showing that it does not solely rely on the normal/abnormal prompt for anomaly detection.
ExpID~\ref{tab:classname_ablate}.1 shuffles the pseudo class names across classes by assigning $s_{c=i}$, the pseudo class name of class $i$, to class $j$, where $i\neq j$. Compared to \ours (expID~\ref{tab:classname_ablate}.3), this hurts performance significantly. 
ExpID~\ref{tab:classname_ablate}.2  tests the alternative of eliminating the text encoder of ALIGN completely, simply learning a binary classifier of weight vectors $t_{n,c},\ t_{a,c}$ per image class $c$. This approach is even less effective, showing the importance of the prior knowledge encoded in the foundation model about both classes and normal/abnormal images. The small sizes of the AD datasets are not sufficient to overcome the use of this prior.

\hypersetup{allcolors=black}
\begin{table}[]
    \centering
    \resizebox{\linewidth}{!}{
    \begin{tabular}{@{}c@{\hspace{.3em}}c@{\hspace{.3em}}c@{\hspace{.3em}}c@{\hspace{.5em}}c@{\hspace{.7em}}c@{\hspace{.7em}}c@{\hspace{.7em}}c@{\hspace{.7em}}c@{\hspace{.7em}}c@{}}
        \toprule
        \multirow{2.5}{*}{expID} &\multirow{2.5}{*}{\shortstack{Layer Aware\\Projection}} & \multirow{2.5}{*}{\shortstack{Pooling\\Operation}} &\multirow{2.5}{*}{Prompt} & \multicolumn{3}{c}{Detection} & \multicolumn{3}{c}{Segmentation}  \\
        \cmidrule(l){5-7} \cmidrule(l){8-10}
        & & & & All & High & Low & All & High & Low \\
        \midrule
        \ref{tab:ablate_pta}.1 &\ccross & N/A & Single & 83.78 & 96.35 & 72.79 & 87.18 & 91.08 & 83.77 \\
        \ref{tab:ablate_pta}.2 &\ccheck & Mean & Single & 81.25 & 90.95 & 72.76 & 88.28 & 91.89 & 85.12 \\
        \ref{tab:ablate_pta}.3 &\ccheck & Max & Multiple & 77.19 & 94.06 & 62.43 & 91.14 & \textbf{95.37} & 87.44 \\
        \ref{tab:ablate_pta}.4 (\ours) &\ccheck & Max & Single & \textbf{84.12} & \textbf{97.02} & \textbf{72.84} & \textbf{91.36} & 95.13 & \textbf{88.07} \\
        \bottomrule
    \end{tabular}
    }
    \caption{Ablation Study without RM on MVTec-\s{100} dataset.}
    \vspace{-5pt}
    \label{tab:ablate_pta}
\end{table}
\hypersetup{allcolors=red}
\hypersetup{citecolor=blue}

\hypersetup{allcolors=black}
\begin{table}[]
    \centering
    \resizebox{\linewidth}{!}{
    \begin{tabular}{@{}c@{}c@{\hspace{1em}}c@{\hspace{1em}}cccccc@{}}
    \toprule
        \multirow{2.5}{*}{expID} 
        &\multirow{2.5}{*}{\shortstack{assign $s_{c=i}$\\to class $i$}} 
        &\multirow{2.5}{*}{\shortstack{use text\\encoder $T$}} 
        &\multicolumn{3}{c}{Detection} & \multicolumn{3}{c}{Segmentation}  \\
        \cmidrule(l){4-6} \cmidrule(l){7-9}
      & & & All & High & Low & All & High & Low \\
      \midrule
      \ref{tab:classname_ablate}.1 & \ccross & \ccheck & 72.76 & 81.06 & 65.49 & 63.74 & 62.09 & 65.18 \\
      \ref{tab:classname_ablate}.2 & \ccheck & \ccross & 59.79 & 63.34 & 56.69 & 69.83 & 70.95 & 68.85\\
      \ref{tab:classname_ablate}.3 (\ours) & \ccheck & \ccheck & \textbf{84.12} & \textbf{97.02} & \textbf{72.84} & \textbf{91.36} & \textbf{95.13} & \textbf{88.07} \\
      \bottomrule
    \end{tabular}
    }
    \caption{Importance of pseudo class name $s_c$ on MVTec-\s{100}. 
    }
    \vspace{-10pt}
    \label{tab:classname_ablate}
\end{table}
\hypersetup{allcolors=red}
\hypersetup{citecolor=blue}

\begin{table}[]
    \centering
    \resizebox{\linewidth}{!}{
    \begin{tabular}{c|cccc}
     \toprule
         & \multicolumn{4}{c}{$v^a$} \\ 
       $v^n$  & a broken & a damaged & an abnormal & a defective \\
       \midrule
       a  & \textbf{84.12} / 91.36 
 & 82.95 / 91.70
 & 82.20 / 91.33
 & 83.66 / \textbf{91.87}
\\
       a normal  &  83.71 / 91.39
 & 82.74 / 91.21
  & 83.47 / 91.23
 & 82.94 / 91.26
\\
       a good  &  75.68 / 90.75
 & 82.14 / 91.22
  & 81.03 / 91.15
& 82.09 / 91.23
\\
       a flawless &  65.63 / 87.61
& 79.09 / 91.00
& 76.13 / 90.24
& 83.89 / 91.42
\\
    \bottomrule
    \end{tabular}}
    \caption{Ablation on different normal/abnormal text prompts (\ie, $v^n$ and $v^a$) on MVTec \s{100}. The AD/AS performances are reported.}
    \vspace{-10pt}
    \label{tab:prompt_ablate}
\end{table}

\begin{figure}
    \centering
    \resizebox{1.03\linewidth}{!}{
    \begin{tabular}{cc}
       \includegraphics[width=0.49\linewidth]{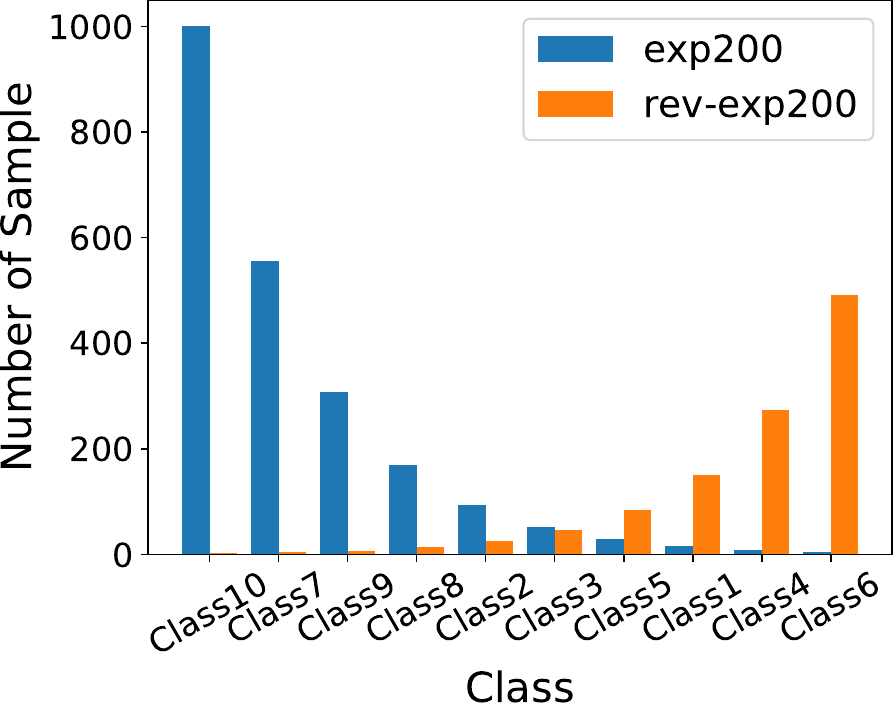}  &  
       \includegraphics[width=0.49\linewidth]{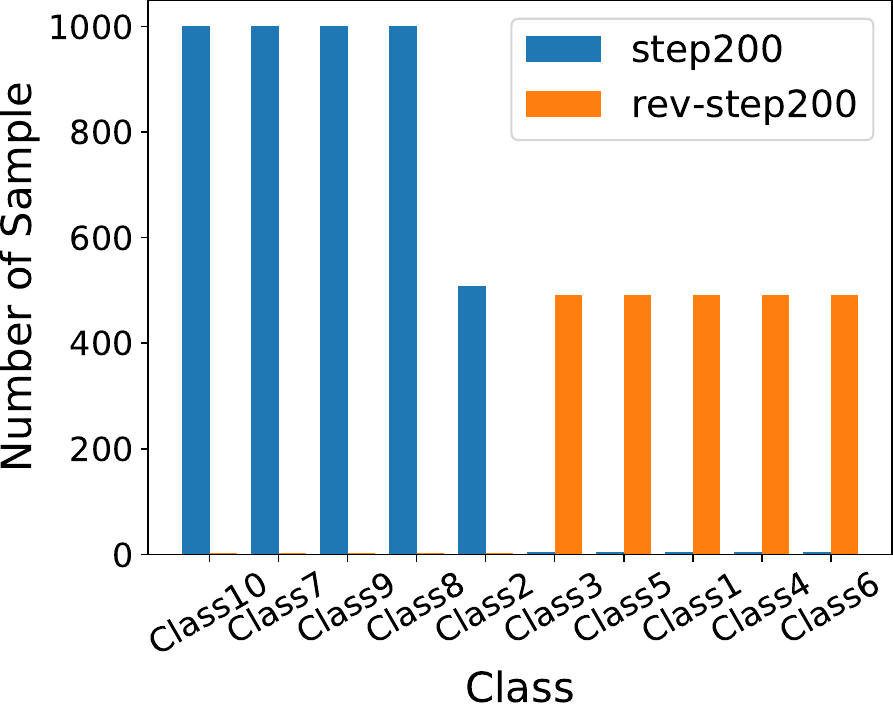}  \\
       \small{(a) \e{}; sample distributions} & \small{(b) \s{}; sample distributions} \\
        \includegraphics[width=0.49\linewidth]{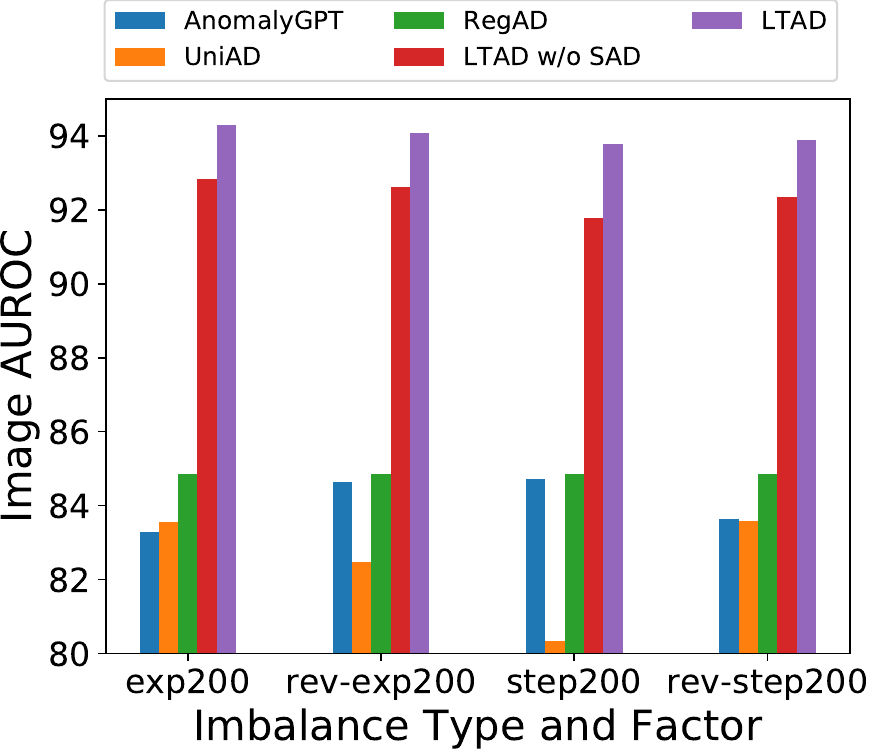}  &  
       \includegraphics[width=0.49\linewidth]{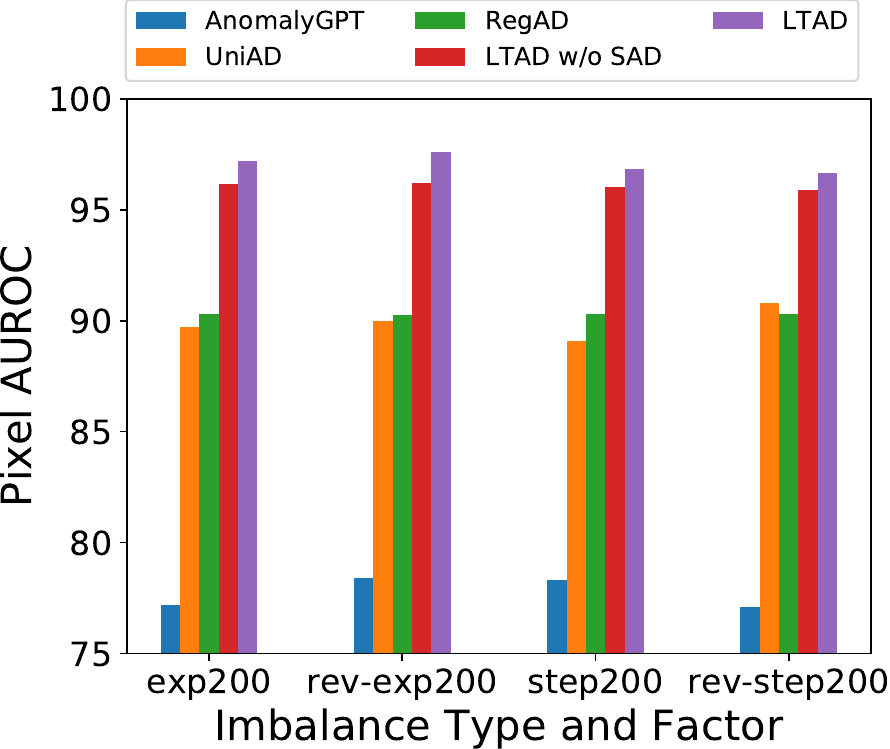}  \\
       \small{(c) detection performance} & \small{(d) segmentation performance} 
    \end{tabular}
    }
    \caption{Top: Sample distributions before and after class reversal, for exponential and step imbalance, respectively. Bottom: AS and AD AUROC vs. imbalance for different dataset configurations.  
    }
    \vspace{-10pt}
    \label{fig:reverse_sample_ablate}
\end{figure}

Tab.~\ref{tab:prompt_ablate} further ablates different combinations for normal  $v^n$ and abnormal $v^a$ text prompts. The combinations of (``a", ``a broken") and (``a", ``a defective") outperform all others. Note that a poor choice of these prompts can degrade performance, although the latter seems to be more sensitive to the choice of normal than abnormal prompt.
This shows the importance of leveraging the prior knowledge of  foundation model about the two conditions.

\noindent{\textbf{Distribution imbalance:}} 
A set of experiments were performed on DAGM~\cite{dagm} to evaluate the robustness of \ours to the class imbalance of the training dataset. To avoid the possibility of overfitting on a specific class order, we repeated these experiment with reverse class order (the least populated classes become the most populated ones). We considered both exponential and step dataset imbalance. Fig.~\ref{fig:reverse_sample_ablate} (a,b) show the class cardinalities before (blue) and after (orange) reversing  the class order. Fig.~\ref{fig:reverse_sample_ablate} (c-d) compare the AD and AS performance of the different approaches. For all dataset configurations, \ours outperforms the baselines by at least 9.1\% (5.9\%) for AD (AS). 
Part of these gains are due to SAD module, which improves performance by 1.62\% (1.01\%) for AD (AS) on average. These results show that \ours generalizes across imbalance factors, that SAD module consistently improves performance, and that its gains are insensitive to class order.

\vspace{-5pt}
\section{Conclusion} \vspace{-3pt}
In this work, we have introduced the task of long-tailed AD, by proposing datasets and performance metrics and a novel AD method, \ours, tailored for the long-tailed setting. \ours detect defects from multiple and long-tailed classes, without relying on dataset class names. It combines AD by reconstruction and semantic AD modules. AD by reconstruction is implemented with a transformer-based reconstruction module. Semantic AD is implemented with a binary classifier, which relies on learned pseudo class names and a pretrained foundation model. These modules are learned over two phases. Phase 1 learns the pseudo-class names and a VAE for feature synthesis that augments the training data to combat long-tails. Phase 2 then learns the parameters of the reconstruction and classification modules of \ours. Experiments show that \ours outperforms the SOTA AD methods on most long-tailed datasets considered and all the components of \ours contribute to its superior performance.

\noindent\textbf{Acknowledgement} 
CH and KCP were supported by Mitsubishi
Electric Research Laboratories. CH and NV were partially funded by NSF awards IIS-2303153 and gift from Qualcomm.

{
    \small
    \bibliographystyle{ieeenat_fullname}
    \bibliography{main}

\begin{thebibliography}{89}
\providecommand{\natexlab}[1]{#1}
\providecommand{\url}[1]{\texttt{#1}}
\expandafter\ifx\csname urlstyle\endcsname\relax
  \providecommand{\doi}[1]{doi: #1}\else
  \providecommand{\doi}{doi: \begingroup \urlstyle{rm}\Url}\fi

\bibitem[Akçay et~al.(2018)Akçay, Atapour-Abarghouei, and Breckon]{Akay2018GANomalySA}
Samet Akçay, Amir Atapour-Abarghouei, and T. Breckon.
\newblock {GANomaly}: Semi-supervised anomaly detection via adversarial training.
\newblock \emph{ArXiv}, abs/1805.06725, 2018.

\bibitem[Arora et~al.(2017)Arora, Verma, Mishra, and Rai]{Arora2017GeneralizedZL}
Gundeep Arora, Vinay~Kumar Verma, Ashish Mishra, and Piyush Rai.
\newblock Generalized zero-shot learning via synthesized examples.
\newblock \emph{2018 IEEE/CVF Conference on Computer Vision and Pattern Recognition}, pages 4281--4289, 2017.

\bibitem[Bae et~al.(2022)Bae, Lee, and Kim]{Bae2022PNII}
Jaehyeok Bae, Jaehyeon Lee, and Seyun Kim.
\newblock {PNI}: Industrial anomaly detection using position and neighborhood information.
\newblock 2022.

\bibitem[Bergmann et~al.(2018)Bergmann, L{\"o}we, Fauser, Sattlegger, and Steger]{Bergmann2018ImprovingUD}
Paul Bergmann, Sindy L{\"o}we, Michael Fauser, David Sattlegger, and Carsten Steger.
\newblock Improving unsupervised defect segmentation by applying structural similarity to autoencoders.
\newblock \emph{ArXiv}, abs/1807.02011, 2018.

\bibitem[Bergmann et~al.(2019)Bergmann, Fauser, Sattlegger, and Steger]{mvtec}
Paul Bergmann, Michael Fauser, David Sattlegger, and Carsten Steger.
\newblock {MVTec AD} — {A} comprehensive real-world dataset for unsupervised anomaly detection.
\newblock In \emph{2019 IEEE/CVF Conference on Computer Vision and Pattern Recognition (CVPR)}, pages 9584--9592, 2019.

\bibitem[Buda et~al.(2017)Buda, Maki, and Mazurowski]{Buda2017ASS}
Mateusz Buda, Atsuto Maki, and Maciej~A. Mazurowski.
\newblock A systematic study of the class imbalance problem in convolutional neural networks.
\newblock \emph{Neural networks: the official journal of the International Neural Network Society}, 106:\penalty0 249--259, 2017.

\bibitem[Cao et~al.(2019{\natexlab{a}})Cao, Wei, Gaidon, Ar{\'e}chiga, and Ma]{Cao2019LearningID}
Kaidi Cao, Colin Wei, Adrien Gaidon, Nikos Ar{\'e}chiga, and Tengyu Ma.
\newblock Learning imbalanced datasets with label-distribution-aware margin loss.
\newblock In \emph{Neural Information Processing Systems}, 2019{\natexlab{a}}.

\bibitem[Cao et~al.(2019{\natexlab{b}})Cao, Wei, Gaidon, Arechiga, and Ma]{cao2019learning}
Kaidi Cao, Colin Wei, Adrien Gaidon, Nikos Arechiga, and Tengyu Ma.
\newblock Learning imbalanced datasets with label-distribution-aware margin loss.
\newblock In \emph{Advances in Neural Information Processing Systems}, 2019{\natexlab{b}}.

\bibitem[Cao et~al.()Cao, Xu, Sun, Cheng, Du, Gao, and Shen]{sam_ad}
Yunkang Cao, Xiaohao Xu, Chen Sun, Yuqi Cheng, Zongwei Du, Liang Gao, and Weiming Shen.
\newblock Segment any anomaly without training via hybrid prompt regularization.
\newblock \penalty0 ({arXiv}:2305.10724).

\bibitem[Cao et~al.(2023)Cao, Xu, Sun, Huang, and Shen]{Cao2023TowardsGA}
Yunkang Cao, Xiaohao Xu, Chen Sun, Xiaonan Huang, and Weiming Shen.
\newblock Towards generic anomaly detection and understanding: Large-scale visual-linguistic model ({GPT-4V}) takes the lead.
\newblock \emph{ArXiv}, abs/2311.02782, 2023.

\bibitem[Chalapathy and Chawla(2019)]{Chalapathy2019DeepLF}
Raghavendra Chalapathy and Sanjay Chawla.
\newblock Deep learning for anomaly detection: A survey.
\newblock \emph{ArXiv}, abs/1901.03407, 2019.

\bibitem[Chandola et~al.(2009)Chandola, Banerjee, and Kumar]{Chandola09}
Varun Chandola, Arindam Banerjee, and Vipin Kumar.
\newblock Anomaly detection: A survey.
\newblock \emph{ACM Comput. Surv.}, 41\penalty0 (3), 2009.

\bibitem[Chawla et~al.(2002)Chawla, Bowyer, Hall, and Kegelmeyer]{Chawla2002SMOTESM}
N. Chawla, K. Bowyer, Lawrence~O. Hall, and W.~Philip Kegelmeyer.
\newblock {SMOTE}: Synthetic minority over-sampling technique.
\newblock \emph{ArXiv}, abs/1106.1813, 2002.

\bibitem[Chen et~al.(2023)Chen, Han, and Zhang]{april_gan}
Xuhai Chen, Yue Han, and Jiangning Zhang.
\newblock A zero-/few-shot anomaly classification and segmentation method for {CVPR} 2023 {VAND} workshop challenge tracks 1\&2: 1st place on zero-shot {AD} and 4th place on few-shot {AD}.
\newblock \emph{ArXiv}, abs/2305.17382, 2023.

\bibitem[Chu et~al.(2020)Chu, Bian, Liu, and Ling]{Chu2020FeatureSA}
Peng Chu, Xiao Bian, Shaopeng Liu, and Haibin Ling.
\newblock Feature space augmentation for long-tailed data.
\newblock In \emph{European Conference on Computer Vision}, 2020.

\bibitem[Cohen and Hoshen(2020)]{spade}
Niv Cohen and Yedid Hoshen.
\newblock Sub-image anomaly detection with deep pyramid correspondences.
\newblock \emph{ArXiv}, abs/2005.02357, 2020.

\bibitem[Collin and Vleeschouwer(2020)]{Collin2020ImprovedAD}
Anne Collin and Christophe~De Vleeschouwer.
\newblock Improved anomaly detection by training an autoencoder with skip connections on images corrupted with stain-shaped noise.
\newblock \emph{2020 25th International Conference on Pattern Recognition (ICPR)}, pages 7915--7922, 2020.

\bibitem[Cui et~al.(2019)Cui, Jia, Lin, Song, and Belongie]{Cui2019ClassBalancedLB}
Yin Cui, Menglin Jia, Tsung-Yi Lin, Yang Song, and Serge~J. Belongie.
\newblock Class-balanced loss based on effective number of samples.
\newblock \emph{2019 IEEE/CVF Conference on Computer Vision and Pattern Recognition (CVPR)}, pages 9260--9269, 2019.

\bibitem[Defard et~al.(2020)Defard, Setkov, Loesch, and Audigier]{padim}
Thomas Defard, Aleksandr Setkov, Ang{\'e}lique Loesch, and Romaric Audigier.
\newblock {PaDiM}: A patch distribution modeling framework for anomaly detection and localization.
\newblock In \emph{ICPR Workshops}, 2020.

\bibitem[Deng and Li(2022)]{RD}
Hanqiu Deng and Xingyu Li.
\newblock Anomaly detection via reverse distillation from one-class embedding.
\newblock \emph{2022 IEEE/CVF Conference on Computer Vision and Pattern Recognition (CVPR)}, pages 9727--9736, 2022.

\bibitem[Dong et~al.(2023)Dong, Zhou, Yan, and Zuo]{dong2023lpt}
Bowen Dong, Pan Zhou, Shuicheng Yan, and Wangmeng Zuo.
\newblock {LPT}: Long-tailed prompt tuning for image classification.
\newblock In \emph{The Eleventh International Conference on Learning Representations}, 2023.

\bibitem[Drummond(2003)]{Drummond2003C4}
Chris Drummond.
\newblock C4.5, class imbalance, and cost sensitivity: Why under-sampling beats oversampling.
\newblock 2003.

\bibitem[Fridman et~al.(2021)Fridman, Rusanovsky, and Oren]{pcb_defect}
Yehonatan Fridman, Matan Rusanovsky, and Gal Oren.
\newblock {ChangeChip}: A reference-based unsupervised change detection for {PCB} defect detection.
\newblock \emph{2021 IEEE Physical Assurance and Inspection of Electronics (PAINE)}, pages 1--8, 2021.

\bibitem[Gal et~al.(2023)Gal, Alaluf, Atzmon, Patashnik, Bermano, Chechik, and Cohen-or]{gal2023an}
Rinon Gal, Yuval Alaluf, Yuval Atzmon, Or Patashnik, Amit~Haim Bermano, Gal Chechik, and Daniel Cohen-or.
\newblock An image is worth one word: Personalizing text-to-image generation using textual inversion.
\newblock In \emph{The Eleventh International Conference on Learning Representations}, 2023.

\bibitem[Girdhar et~al.(2023)Girdhar, El-Nouby, Liu, Singh, Alwala, Joulin, and Misra]{imageblind}
Rohit Girdhar, Alaaeldin El-Nouby, Zhuang Liu, Mannat Singh, Kalyan~Vasudev Alwala, Armand Joulin, and Ishan Misra.
\newblock Imagebind one embedding space to bind them all.
\newblock \emph{2023 IEEE/CVF Conference on Computer Vision and Pattern Recognition (CVPR)}, pages 15180--15190, 2023.

\bibitem[Gu et~al.(2023)Gu, Zhu, Zhu, Chen, Tang, and Wang]{anomalygpt}
Zhaopeng Gu, Bingke Zhu, Guibo Zhu, Yingying Chen, Ming Tang, and Jinqiao Wang.
\newblock {AnomalyGPT}: Detecting industrial anomalies using large vision-language models.
\newblock \emph{arXiv preprint arXiv:2308.15366}, 2023.

\bibitem[Han et~al.(2005)Han, Wang, and Mao]{Han2005BorderlineSMOTEAN}
Hui Han, Wenyuan Wang, and Binghuan Mao.
\newblock {Borderline-SMOTE}: A new over-sampling method in imbalanced data sets learning.
\newblock In \emph{International Conference on Intelligent Computing}, 2005.

\bibitem[He and Garcia(2009)]{He2009LearningFI}
Haibo He and Edwardo~A. Garcia.
\newblock Learning from imbalanced data.
\newblock \emph{IEEE Transactions on Knowledge and Data Engineering}, 21:\penalty0 1263--1284, 2009.

\bibitem[He et~al.(2008)He, Bai, Garcia, and Li]{He2008ADASYNAS}
Haibo He, Yang Bai, Edwardo~A. Garcia, and Shutao Li.
\newblock {ADASYN}: Adaptive synthetic sampling approach for imbalanced learning.
\newblock \emph{2008 IEEE International Joint Conference on Neural Networks (IEEE World Congress on Computational Intelligence)}, pages 1322--1328, 2008.

\bibitem[Huang et~al.(2022)Huang, Guan, Jiang, Zhang, Spratlin, and Wang]{regad}
Chaoqin Huang, Haoyan Guan, Aofan Jiang, Ya Zhang, Michael Spratlin, and Yanfeng Wang.
\newblock Registration based few-shot anomaly detection.
\newblock In \emph{European Conference on Computer Vision (ECCV)}, 2022.

\bibitem[Hyun et~al.(2023)Hyun, Kim, Jeon, Kim, Bae, and Kang]{ReConPatch}
Jeeho Hyun, Sangyun Kim, Giyoung Jeon, Seungwook Kim, Kyunghoon Bae, and Byungjin Kang.
\newblock {ReConPatch}: Contrastive patch representation learning for industrial anomaly detection.
\newblock \emph{ArXiv}, abs/2305.16713, 2023.

\bibitem[Jeong et~al.(2023)Jeong, Zou, Kim, Zhang, Ravichandran, and Dabeer]{winclip}
Jongheon Jeong, Yang Zou, Taewan Kim, Dongqing Zhang, Avinash Ravichandran, and Onkar Dabeer.
\newblock {WinCLIP}: Zero-/few-shot anomaly classification and segmentation.
\newblock \emph{2023 IEEE/CVF Conference on Computer Vision and Pattern Recognition (CVPR)}, pages 19606--19616, 2023.

\bibitem[Jia et~al.(2021)Jia, Yang, Xia, Chen, Parekh, Pham, Le, Sung, Li, and Duerig]{align}
Chao Jia, Yinfei Yang, Ye Xia, Yi-Ting Chen, Zarana Parekh, Hieu Pham, Quoc~V. Le, Yun-Hsuan Sung, Zhen Li, and Tom Duerig.
\newblock Scaling up visual and vision-language representation learning with noisy text supervision.
\newblock In \emph{International Conference on Machine Learning}, 2021.

\bibitem[Kang et~al.(2019)Kang, Xie, Rohrbach, Yan, Gordo, Feng, and Kalantidis]{Kang2019DecouplingRA}
Bingyi Kang, Saining Xie, Marcus Rohrbach, Zhicheng Yan, Albert Gordo, Jiashi Feng, and Yannis Kalantidis.
\newblock Decoupling representation and classifier for long-tailed recognition.
\newblock \emph{ArXiv}, abs/1910.09217, 2019.

\bibitem[Kang et~al.(2021)Kang, Li, Xie, Yuan, and Feng]{kang2021exploring}
Bingyi Kang, Yu Li, Sa Xie, Zehuan Yuan, and Jiashi Feng.
\newblock Exploring balanced feature spaces for representation learning.
\newblock In \emph{International Conference on Learning Representations}, 2021.

\bibitem[Kim et~al.(2022)Kim, Park, Cho, and Lee]{Kim2022FAPMFA}
Donghyeon Kim, Chaewon Park, Suhwan Cho, and Sangyoun Lee.
\newblock {FAPM}: Fast adaptive patch memory for real-time industrial anomaly detection.
\newblock \emph{ICASSP 2023 - 2023 IEEE International Conference on Acoustics, Speech and Signal Processing (ICASSP)}, pages 1--5, 2022.

\bibitem[Kingma and Welling(2013)]{vae}
Diederik~P. Kingma and Max Welling.
\newblock Auto-encoding variational bayes.
\newblock \emph{CoRR}, abs/1312.6114, 2013.

\bibitem[Kingma and Welling(2019)]{vae_intro}
Diederik~P. Kingma and Max Welling.
\newblock An introduction to variational autoencoders.
\newblock \emph{ArXiv}, abs/1906.02691, 2019.

\bibitem[Kirillov et~al.(2023)Kirillov, Mintun, Ravi, Mao, Rolland, Gustafson, Xiao, Whitehead, Berg, Lo, Doll{\'a}r, and Girshick]{sam}
Alexander Kirillov, Eric Mintun, Nikhila Ravi, Hanzi Mao, Chloe Rolland, Laura Gustafson, Tete Xiao, Spencer Whitehead, Alexander~C. Berg, Wan-Yen Lo, Piotr Doll{\'a}r, and Ross~B. Girshick.
\newblock Segment anything.
\newblock \emph{2023 IEEE/CVF International Conference on Computer Vision (ICCV)}, 2023.

\bibitem[Krishnan et~al.(2018)Krishnan, Sharma, Sankar, and Sundaram]{Krishnan2018AnAA}
Adit Krishnan, Ashish Sharma, Aravind Sankar, and H. Sundaram.
\newblock An adversarial approach to improve long-tail performance in neural collaborative filtering.
\newblock \emph{Proceedings of the 27th ACM International Conference on Information and Knowledge Management}, 2018.

\bibitem[Lee et~al.(2022)Lee, Lee, and Song]{Lee2022CFACF}
Sungwook Lee, Seunghyun Lee, and Byung~Cheol Song.
\newblock {CFA}: Coupled-hypersphere-based feature adaptation for target-oriented anomaly localization.
\newblock \emph{IEEE Access}, 10:\penalty0 78446--78454, 2022.

\bibitem[Lee and Kang(2022)]{AnoViT}
Yunseung Lee and Pilsung Kang.
\newblock {AnoViT}: Unsupervised anomaly detection and localization with vision transformer-based encoder-decoder.
\newblock \emph{IEEE Access}, PP:\penalty0 1--1, 2022.

\bibitem[Li et~al.(2021{\natexlab{a}})Li, Sohn, Yoon, and Pfister]{cutandpaste}
Chun-Liang Li, Kihyuk Sohn, Jinsung Yoon, and Tomas Pfister.
\newblock {CutPaste}: Self-supervised learning for anomaly detection and localization.
\newblock \emph{2021 IEEE/CVF Conference on Computer Vision and Pattern Recognition (CVPR)}, pages 9659--9669, 2021{\natexlab{a}}.

\bibitem[Li et~al.(2023)Li, Hu, Li, Chen, Zheng, and Shen]{Li2023TargetBS}
Hanxi Li, Jianfei Hu, Bo Li, Hao Chen, Yongbin Zheng, and Chunhua Shen.
\newblock Target before shooting: Accurate anomaly detection and localization under one millisecond via cascade patch retrieval.
\newblock \emph{ArXiv}, abs/2308.06748, 2023.

\bibitem[Li et~al.(2021{\natexlab{b}})Li, Gong, Liu, Wang, Qiao, and Cheng]{Li2021MetaSAugMS}
Shuang Li, Kaixiong Gong, Chi~Harold Liu, Yulin Wang, Feng Qiao, and Xinjing Cheng.
\newblock {MetaSAug}: Meta semantic augmentation for long-tailed visual recognition.
\newblock \emph{2021 IEEE/CVF Conference on Computer Vision and Pattern Recognition (CVPR)}, pages 5208--5217, 2021{\natexlab{b}}.

\bibitem[Lin et~al.(2017)Lin, Goyal, Girshick, He, and Dollár]{focal_loss}
Tsung-Yi Lin, Priya Goyal, Ross Girshick, Kaiming He, and Piotr Dollár.
\newblock Focal loss for dense object detection.
\newblock In \emph{2017 IEEE International Conference on Computer Vision (ICCV)}, pages 2999--3007, 2017.

\bibitem[Liu et~al.(2023)Liu, Xie, Wang, Li, Wang, Zheng, and Jin]{Liu2023DeepII}
Jiaqi Liu, Guoyang Xie, Jingbao Wang, Shangwen Li, Chengjie Wang, Feng Zheng, and Yaochu Jin.
\newblock Deep industrial image anomaly detection: A survey.
\newblock \emph{ArXiv}, abs/2301.11514, 2023.

\bibitem[Liu et~al.(2019)Liu, Li, Zheng, Karanam, Wu, Bhanu, Radke, and Camps]{Liu2019TowardsVE}
Wenqian Liu, Runze Li, Meng Zheng, Srikrishna Karanam, Ziyan Wu, Bir Bhanu, Richard~J. Radke, and Octavia~I. Camps.
\newblock Towards visually explaining variational autoencoders.
\newblock \emph{2020 IEEE/CVF Conference on Computer Vision and Pattern Recognition (CVPR)}, pages 8639--8648, 2019.

\bibitem[Lu et~al.(2023)Lu, Wu, Tian, Wang, Chen, Liu, and Hu]{Lu2023HierarchicalVQ}
Ruiying Lu, YuJie Wu, Long Tian, Dongsheng Wang, Bo Chen, Xiyang Liu, and Ruimin Hu.
\newblock Hierarchical vector quantized transformer for multi-class unsupervised anomaly detection.
\newblock \emph{ArXiv}, abs/2310.14228, 2023.

\bibitem[Ma et~al.(2023)Ma, Zhang, Yang, Cao, Xie, Tian, and Li]{wafer_defect}
Jianhong Ma, Tao Zhang, Cong Yang, Yangjie Cao, Lipeng Xie, Hui Tian, and Xuexiang Li.
\newblock Review of wafer surface defect detection methods.
\newblock \emph{Electronics}, 12\penalty0 (8), 2023.

\bibitem[Ma et~al.(2021)Ma, Geng, Wang, Shao, Lu, Li, Gao, and Qiao]{Ma2021ASL}
Teli Ma, Shijie Geng, Mengmeng Wang, Jing Shao, Jiasen Lu, Hongsheng Li, Peng Gao, and Y. Qiao.
\newblock A simple long-tailed recognition baseline via vision-language model.
\newblock \emph{ArXiv}, abs/2111.14745, 2021.

\bibitem[Ma et~al.(2022)Ma, Geng, Wang, Xu, Li, Zhang, Gao, and Qiao]{Ma_2022_BMVC}
Teli Ma, Shijie Geng, Mengmeng Wang, Sheng Xu, Hongsheng Li, Baochang Zhang, Peng Gao, and Yu Qiao.
\newblock Unleashing the potential of vision-language models for long-tailed visual recognition.
\newblock In \emph{33rd British Machine Vision Conference 2022, {BMVC} 2022, London, UK, November 21-24, 2022}. {BMVA} Press, 2022.

\bibitem[Mishra et~al.(2021)Mishra, Verk, Fornasier, Piciarelli, and Foresti]{vtadl}
Pankaj Mishra, Riccardo Verk, Daniele Fornasier, Claudio Piciarelli, and Gian~Luca Foresti.
\newblock {VT-ADL}: A vision transformer network for image anomaly detection and localization.
\newblock \emph{2021 IEEE 30th International Symposium on Industrial Electronics (ISIE)}, pages 01--06, 2021.

\bibitem[Mou et~al.(2023)Mou, Gu, Cao, Bai, Huang, Shan, and Shi]{rgi}
Shancong Mou, Xiaoyi Gu, Meng Cao, Haoping Bai, Ping Huang, Jiulong Shan, and Jianjun Shi.
\newblock {RGI}: Robust {GAN}-inversion for mask-free image inpainting and unsupervised pixel-wise anomaly detection.
\newblock In \emph{ICLR}, 2023.

\bibitem[Paszke et~al.(2019)Paszke, Gross, Massa, Lerer, Bradbury, Chanan, Killeen, Lin, Gimelshein, Antiga, Desmaison, K{\"o}pf, Yang, DeVito, Raison, Tejani, Chilamkurthy, Steiner, Fang, Bai, and Chintala]{pytorch}
Adam Paszke, Sam Gross, Francisco Massa, Adam Lerer, James Bradbury, Gregory Chanan, Trevor Killeen, Zeming Lin, Natalia Gimelshein, Luca Antiga, Alban Desmaison, Andreas K{\"o}pf, Edward Yang, Zach DeVito, Martin Raison, Alykhan Tejani, Sasank Chilamkurthy, Benoit Steiner, Lu Fang, Junjie Bai, and Soumith Chintala.
\newblock {PyTorch}: An imperative style, high-performance deep learning library.
\newblock In \emph{Neural Information Processing Systems}, 2019.

\bibitem[Pirnay and Chai(2021)]{Pirnay2021InpaintingTF}
Jonathan Pirnay and Keng~Yip Chai.
\newblock Inpainting transformer for anomaly detection.
\newblock In \emph{International Conference on Image Analysis and Processing}, 2021.

\bibitem[Radford et~al.(2021)Radford, Kim, Hallacy, Ramesh, Goh, Agarwal, Sastry, Askell, Mishkin, Clark, Krueger, and Sutskever]{clip}
Alec Radford, Jong~Wook Kim, Chris Hallacy, Aditya Ramesh, Gabriel Goh, Sandhini Agarwal, Girish Sastry, Amanda Askell, Pamela Mishkin, Jack Clark, Gretchen Krueger, and Ilya Sutskever.
\newblock Learning transferable visual models from natural language supervision.
\newblock In \emph{International Conference on Machine Learning}, 2021.

\bibitem[Ren et~al.(2020)Ren, Yu, Sheng, Ma, Zhao, Yi, and Li]{Ren2020BalancedMF}
Jiawei Ren, Cunjun Yu, Shunan Sheng, Xiao Ma, Haiyu Zhao, Shuai Yi, and Hongsheng Li.
\newblock Balanced meta-softmax for long-tailed visual recognition.
\newblock \emph{ArXiv}, abs/2007.10740, 2020.

\bibitem[Rezende and Mohamed(2015)]{JimenezRezende2015VariationalIW}
Danilo~Jimenez Rezende and Shakir Mohamed.
\newblock Variational inference with normalizing flows.
\newblock \emph{ArXiv}, abs/1505.05770, 2015.

\bibitem[Roth et~al.(2021)Roth, Pemula, Zepeda, Scholkopf, Brox, and Gehler]{patchcore}
Karsten Roth, Latha Pemula, Joaquin Zepeda, Bernhard Scholkopf, Thomas Brox, and Peter Gehler.
\newblock Towards total recall in industrial anomaly detection.
\newblock \emph{2022 IEEE/CVF Conference on Computer Vision and Pattern Recognition (CVPR)}, pages 14298--14308, 2021.

\bibitem[Rudolph et~al.(2021)Rudolph, Wandt, and Rosenhahn]{differnet}
Marco Rudolph, Bastian Wandt, and Bodo Rosenhahn.
\newblock Same same but differnet: Semi-supervised defect detection with normalizing flows.
\newblock In \emph{Winter Conference on Applications of Computer Vision (WACV)}, 2021.

\bibitem[Sabokrou et~al.(2018)Sabokrou, Khalooei, Fathy, and Adeli]{sabokrou2018adversarially}
Mohammad Sabokrou, Mohammad Khalooei, Mahmood Fathy, and Ehsan Adeli.
\newblock Adversarially learned one-class classifier for novelty detection.
\newblock In \emph{Proceedings of the IEEE Conference on Computer Vision and Pattern Recognition}, pages 3379--3388, 2018.

\bibitem[Saleh et~al.(2022)Saleh, Konyar, Kaplan, and Ertunç]{tire_defect}
Radhwan A.~A. Saleh, Mehmet~Zeki Konyar, Kaplan Kaplan, and H.~Metin Ertunç.
\newblock Tire defect detection model using machine learning.
\newblock In \emph{2022 2nd International Conference on Emerging Smart Technologies and Applications (eSmarTA)}, pages 1--5, 2022.

\bibitem[Salehi et~al.(2020)Salehi, Sadjadi, Baselizadeh, Rohban, and Rabiee]{mkd}
Mohammadreza Salehi, Niousha Sadjadi, Soroosh Baselizadeh, Mohammad~Hossein Rohban, and Hamid~R. Rabiee.
\newblock Multiresolution knowledge distillation for anomaly detection.
\newblock \emph{2021 IEEE/CVF Conference on Computer Vision and Pattern Recognition (CVPR)}, pages 14897--14907, 2020.

\bibitem[Salehi et~al.(2022)Salehi, Mirzaei, Hendrycks, Li, Rohban, and Sabokrou]{salehi2022a}
Mohammadreza Salehi, Hossein Mirzaei, Dan Hendrycks, Yixuan Li, Mohammad~Hossein Rohban, and Mohammad Sabokrou.
\newblock A unified survey on anomaly, novelty, open-set, and out of-distribution detection: Solutions and future challenges.
\newblock \emph{Transactions on Machine Learning Research}, 2022.

\bibitem[Shu et~al.(2019)Shu, Xie, Yi, Zhao, Zhou, Xu, and Meng]{Shu2019PushTS}
Jun Shu, Qi Xie, Lixuan Yi, Qian Zhao, Sanping Zhou, Zongben Xu, and Deyu Meng.
\newblock Push the student to learn right: Progressive gradient correcting by meta-learner on corrupted labels.
\newblock \emph{ArXiv}, abs/1902.07379, 2019.

\bibitem[Su et~al.(2023)Su, Lan, Li, Xu, Wang, and Cai]{Su2023PandaGPTOM}
Yixuan Su, Tian Lan, Huayang Li, Jialu Xu, Yan Wang, and Deng Cai.
\newblock {PandaGPT}: One model to instruction-follow them all.
\newblock \emph{ArXiv}, abs/2305.16355, 2023.

\bibitem[Tan and Le(2019)]{efficientnet}
Mingxing Tan and Quoc Le.
\newblock {E}fficient{N}et: Rethinking model scaling for convolutional neural networks.
\newblock In \emph{Proceedings of the 36th International Conference on Machine Learning}, pages 6105--6114. PMLR, 2019.

\bibitem[Tien et~al.(2023)Tien, Nguyen, Tran, Huy, Duong, Nguyen, and Truong]{RD_plus}
Tran~Dinh Tien, Anh~Tuan Nguyen, Nguyen~Hoang Tran, Ta~Duc Huy, Soan~T.M. Duong, Chanh D.~Tr. Nguyen, and Steven Q.~H. Truong.
\newblock Revisiting reverse distillation for anomaly detection.
\newblock In \emph{Proceedings of the IEEE/CVF Conference on Computer Vision and Pattern Recognition (CVPR)}, pages 24511--24520, 2023.

\bibitem[Wieler et~al.(2007)Wieler, Hahn, and Hamprecht]{dagm}
Matthias Wieler, Tobias Hahn, and Fred.~A. Hamprecht.
\newblock Weakly supervised learning for industrial optical inspection, 2007.
\newblock \url{https://hci.iwr.uni-heidelberg.de/node/3616}.

\bibitem[Wu et~al.(2020{\natexlab{a}})Wu, Huang, Liu, Wang, and Lin]{Wu2020DistributionBalancedLF}
Tong Wu, Qingqiu Huang, Ziwei Liu, Yu Wang, and Dahua Lin.
\newblock Distribution-balanced loss for multi-label classification in long-tailed datasets.
\newblock \emph{ArXiv}, abs/2007.09654, 2020{\natexlab{a}}.

\bibitem[Wu et~al.(2020{\natexlab{b}})Wu, Morgado, Wang, Ho, and Vasconcelos]{Wu2020SolvingLR}
Tz-Ying Wu, Pedro Morgado, Pei Wang, Chih-Hui Ho, and Nuno Vasconcelos.
\newblock Solving long-tailed recognition with deep realistic taxonomic classifier.
\newblock \emph{ArXiv}, abs/2007.09898, 2020{\natexlab{b}}.

\bibitem[Yang et~al.(2022)Yang, Jiang, Song, and Guo]{Yang2022ASO}
Lu Yang, He Jiang, Qing Song, and Jun Guo.
\newblock A survey on long-tailed visual recognition.
\newblock \emph{International Journal of Computer Vision}, 130:\penalty0 1837 -- 1872, 2022.

\bibitem[Yang and Xu(2020)]{Yang2020RethinkingTV}
Yuzhe Yang and Zhi Xu.
\newblock Rethinking the value of labels for improving class-imbalanced learning.
\newblock \emph{ArXiv}, abs/2006.07529, 2020.

\bibitem[Yap et~al.(2013)Yap, Rani, Rahman, Fong, Khairudin, and Abdullah]{Yap2013AnAO}
Bee~Wah Yap, Khatijahhusna~Abd Rani, Hezlin Aryani~Abd Rahman, Simon~James Fong, Zuraida Khairudin, and Nik~Nik Abdullah.
\newblock An application of oversampling, undersampling, bagging and boosting in handling imbalanced datasets.
\newblock In \emph{International Conference on Advanced Data and Information Engineering}, 2013.

\bibitem[You et~al.(2022)You, Cui, Shen, Yang, Lu, Zheng, and Le]{uniad}
Zhiyuan You, Lei Cui, Yujun Shen, Kai Yang, Xin Lu, Yu Zheng, and Xinyi Le.
\newblock A unified model for multi-class anomaly detection.
\newblock In \emph{Advances in Neural Information Processing Systems}, pages 4571--4584. Curran Associates, Inc., 2022.

\bibitem[Yu1 et~al.(2021)Yu1, Zheng, Wang, Li, Wu, Zhao, and Wu]{FastFlow}
Jiawei Yu1, Ye Zheng, Xiang Wang, Wei Li, Yushuang Wu, Rui Zhao, and Liwei Wu.
\newblock {FastFlow}: Unsupervised anomaly detection and localization via {2D} normalizing flows.
\newblock \emph{ArXiv}, abs/2111.07677, 2021.

\bibitem[Zaheer et~al.(2020)Zaheer, ha~Lee, Astrid, and Lee]{Zaheer2020OldIG}
M. Zaheer, Jin ha Lee, M. Astrid, and Seung-Ik Lee.
\newblock Old is gold: Redefining the adversarially learned one-class classifier training paradigm.
\newblock \emph{2020 IEEE/CVF Conference on Computer Vision and Pattern Recognition (CVPR)}, pages 14171--14181, 2020.

\bibitem[Zavrtanik et~al.(2021)Zavrtanik, Kristan, and Skocaj]{draem}
Vitjan Zavrtanik, Matej Kristan, and Danijel Skocaj.
\newblock {DRAEM} - {A} discriminatively trained reconstruction embedding for surface anomaly detection.
\newblock In \emph{Proceedings of the IEEE/CVF International Conference on Computer Vision (ICCV)}, pages 8330--8339, 2021.

\bibitem[Zhang et~al.(2021{\natexlab{a}})Zhang, Cui, Hung, and Lu]{defectgan}
Gongjie Zhang, Kaiwen Cui, Tzu-Yi Hung, and Shijian Lu.
\newblock {Defect-GAN}: High-fidelity defect synthesis for automated defect inspection.
\newblock \emph{2021 IEEE Winter Conference on Applications of Computer Vision (WACV)}, pages 2523--2533, 2021{\natexlab{a}}.

\bibitem[Zhang et~al.(2023{\natexlab{a}})Zhang, Wang, Wu, and Jiang]{DiffusionAD}
Hui~Min Zhang, Z. Wang, Zuxuan Wu, and Yuwei Jiang.
\newblock {DiffusionAD}: Denoising diffusion for anomaly detection.
\newblock \emph{ArXiv}, abs/2303.08730, 2023{\natexlab{a}}.

\bibitem[Zhang et~al.(2023{\natexlab{b}})Zhang, Chen, Xue, Wang, Wang, and Liu]{Zhang2023ExploringGP}
Jiangning Zhang, Xuhai Chen, Zhucun Xue, Yabiao Wang, Chengjie Wang, and Yong Liu.
\newblock Exploring grounding potential of {VQA}-oriented {GPT-4V} for zero-shot anomaly detection.
\newblock \emph{ArXiv}, abs/2311.02612, 2023{\natexlab{b}}.

\bibitem[Zhang et~al.(2023{\natexlab{c}})Zhang, Ding, Ban, and Yang]{goodad}
Jian Zhang, Runwei Ding, Miaoju Ban, and Ge Yang.
\newblock {PKU-GoodsAD}: A supermarket goods dataset for unsupervised anomaly detection and segmentation.
\newblock \emph{ArXiv}, abs/2307.04956, 2023{\natexlab{c}}.

\bibitem[Zhang et~al.(2023{\natexlab{d}})Zhang, Li, Li, Huang, Shan, and Chen]{DeSTSeg}
Xuan Zhang, Shiyu Li, Xi Li, Ping Huang, Jiulong Shan, and Ting Chen.
\newblock {DeSTSeg}: Segmentation guided denoising student-teacher for anomaly detection.
\newblock In \emph{Proceedings of the IEEE/CVF Conference on Computer Vision and Pattern Recognition (CVPR)}, pages 3914--3923, 2023{\natexlab{d}}.

\bibitem[Zhang et~al.(2021{\natexlab{b}})Zhang, Kang, Hooi, Yan, and Feng]{Zhang2021DeepLL}
Yifan Zhang, Bingyi Kang, Bryan Hooi, Shuicheng Yan, and Jiashi Feng.
\newblock Deep long-tailed learning: A survey.
\newblock \emph{IEEE Transactions on Pattern Analysis and Machine Intelligence}, 45:\penalty0 10795--10816, 2021{\natexlab{b}}.

\bibitem[Zhang et~al.(2021{\natexlab{c}})Zhang, Wei, Zhou, and Wu]{zhang2021tricks}
Yongshun Zhang, Xiu{-}Shen Wei, Boyan Zhou, and Jianxin Wu.
\newblock Bag of tricks for long-tailed visual recognition with deep convolutional neural networks.
\newblock In \emph{AAAI}, pages 3447--3455, 2021{\natexlab{c}}.

\bibitem[Zhao(2023)]{omnial}
Ying Zhao.
\newblock {OmniAL}: A unified cnn framework for unsupervised anomaly localization.
\newblock In \emph{Proceedings of the IEEE/CVF Conference on Computer Vision and Pattern Recognition (CVPR)}, pages 3924--3933, 2023.

\bibitem[Zou et~al.(2018)Zou, Yu, Kumar, and Wang]{Zou2018DomainAF}
Yang Zou, Zhiding Yu, B.~V. K.~Vijaya Kumar, and Jinsong Wang.
\newblock Domain adaptation for semantic segmentation via class-balanced self-training.
\newblock \emph{ArXiv}, abs/1810.07911, 2018.

\bibitem[Zou et~al.(2022)Zou, Jeong, Pemula, Zhang, and Dabeer]{visa}
Yang Zou, Jongheon Jeong, Latha Pemula, Dongqing Zhang, and Onkar Dabeer.
\newblock Spot-the-difference self-supervised pre-training for anomaly detection and segmentation.
\newblock \emph{arXiv preprint arXiv:2207.14315}, 2022.

\end{thebibliography}
}

\clearpage
\appendix

\section{Additional training details}
\subsection{Baselines}
We re-train all baselines on the proposed long-tailed configurations, using the github links of {\color{blue} \href{https://github.com/Runinho/pytorch-cutpaste}{Cut \& Paste}}, {\color{blue}\href{https://github.com/rohban-lab/Knowledge_Distillation_AD}{MKD}}, {\color{blue}\href{https://github.com/VitjanZ/DRAEM}{DRAEM}}, {\color{blue}\href{https://github.com/MediaBrain-SJTU/RegAD}{RegAD}}, {\color{blue}\href{https://github.com/zhiyuanyou/UniAD}{UniAD}} and {\color{blue}\href{https://github.com/CASIA-IVA-Lab/AnomalyGPT}{AnomalyGPT}}. By default, the training pipeline in the github link is followed. Both UniAD and \ours use the architecture of EfficientNetB7 for fair comparison. For RegAD, please refer to Sec.6 for the training details. 
For AnomalyGPT, it trains the model on additional data. For example, when the downstream dataset is MVTec, additional VisA dataset is used for training. For fair comparison, we re-train AnomalyGPT on the proposed long-tailed configurations \textbf{without} using additional dataset. 

\subsection{LTAD and its variants}
All \ours and its variants are trained with batch size of 8. The foundational model of ALIGN is used in \ours, where the architecture of EfficientNetB7 is the image encoder of ALIGN. The details of \ours variants are elaborated below.

\noindent\textbf{ExpID 6.5:}  In this experiment, we assume that the dataset class names is available. Note that such assumption is only for validating whether the dataset class name is helpful for AD/AS performance. The class name in the dataset is assigned to $s_c$ and $s_c$ is not further optimized in this experiment. 

\noindent\textbf{ExpID 6.6:} Similar to ExpID 6.5, we assume the dataset class names is available and such assumption is only for validating whether the dataset class name is helpful for AD/AS performance. Unlike ExpID 6.5, after the class name in the dataset is assigned to $s_c$, $s_c$ is optimized using the loss of (6) in this experiment.

\noindent\textbf{ExpID 6.7:} In this experiment, we randomly initialize the psuedo class name $s_c$ and $s_c$ is optimized using the loss of (6).

\noindent\textbf{ExpID 8.1:} In this experiment, we shuffles the pseudo class names across classes. This is done by, for example, assigning $s_1$ to class 2, $s_2$ to class 5 (i.e. assign the pseudo class name of class $i$, to class $j$, where $i\neq j$). We show that shuffling the pseudo class names across classes hurts the performance significantly. 

\noindent\textbf{ExpID 8.2:} To evaluate the importance of using prior knowledge of the ALIGN text encoder, we remove the text encoder of ALIGN and learn a binary classifier of weight vectors $t_{n,c}$, $t_{a,c}$ per image class $c$. In this experiment, the weight vectors of the binary classifier are randomly initialized and optimized with the loss of (8).

\section{Numerical result}
While we compute average performance across majority (High), minority (Low), and all (All) classes in all experiments, we omit some of the High and Low values in main paper, for brevity. The complete result for anomaly detection and anomaly segmentation on MVTec, VisA and DAGM can be find in Tab.~\ref{tab:mvtec_exp100_ad}-~\ref{tab:mvtec_step200_as}, Tab.~\ref{tab:visa_exp100_ad}-~\ref{tab:visa_step500_as} and Tab.~\ref{tab:dagm_exp50_ad}-~\ref{tab:dagm_revstep200_as}, respectively.

\section{Additional visualizations}
\begin{figure*}
    \centering
    \setlength{\tabcolsep}{1.7pt}
    \begin{tabular}{|c|c|}
        \hline
       \includegraphics[width=0.49\linewidth]{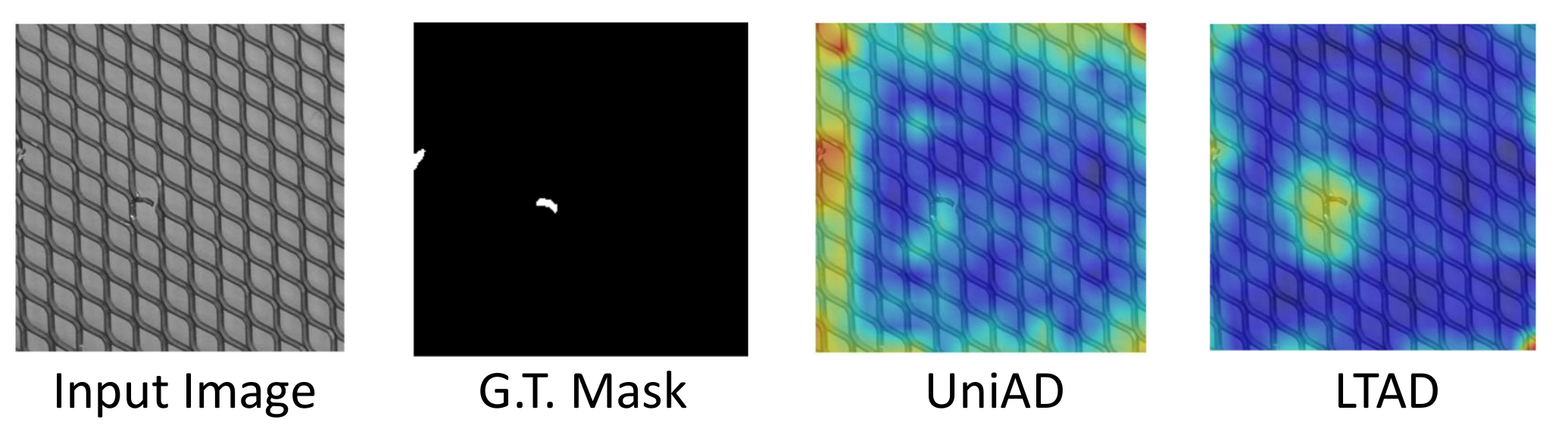}   & \includegraphics[width=0.49\linewidth]{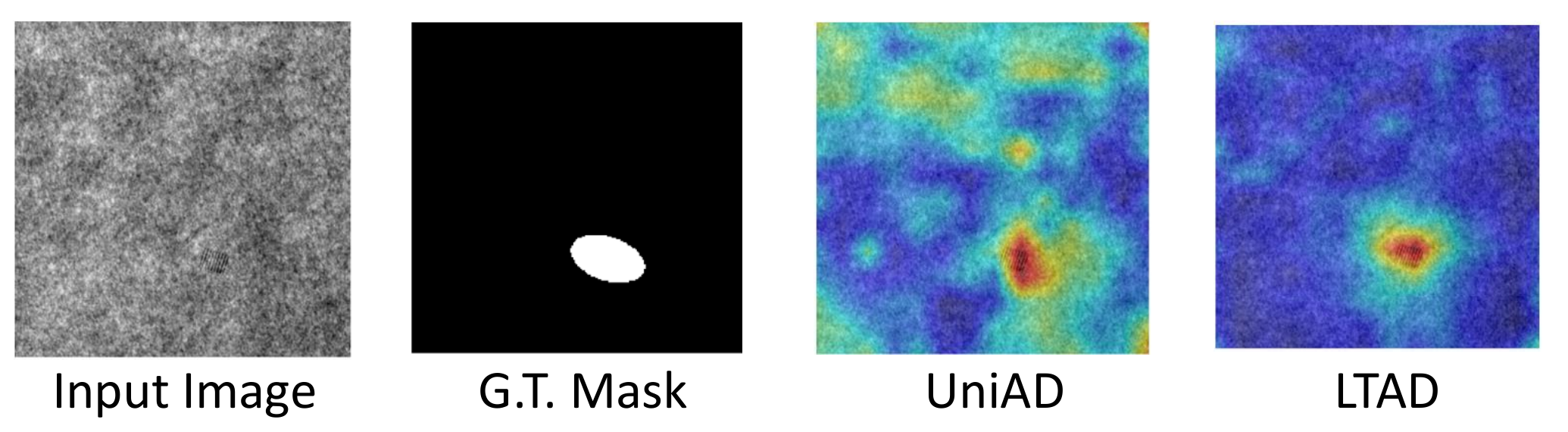} \\
       \includegraphics[width=0.49\linewidth]{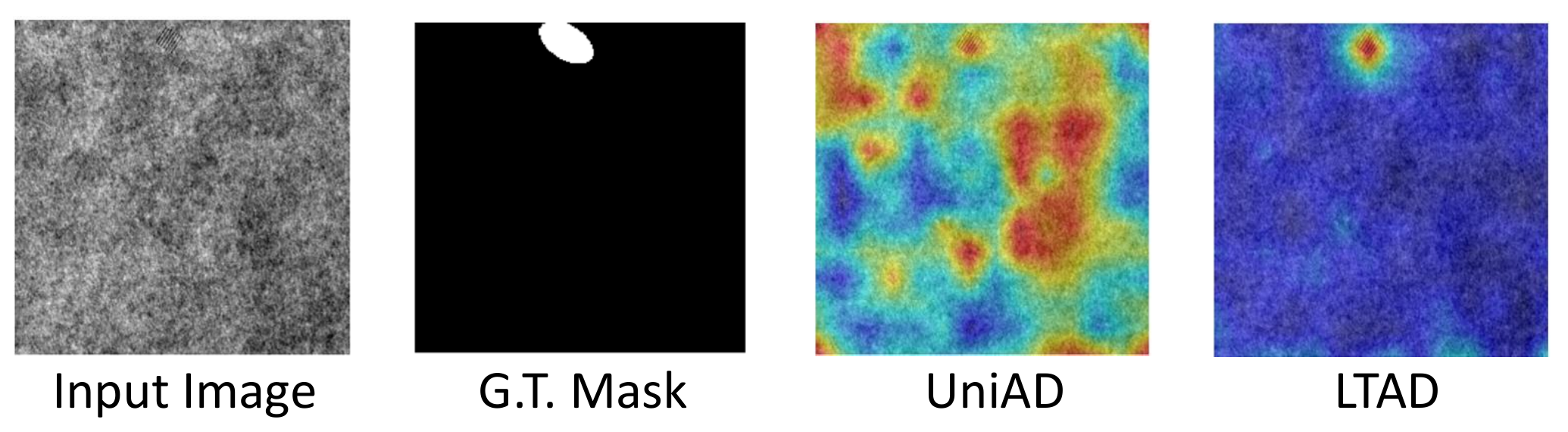} &
       \includegraphics[width=0.49\linewidth]{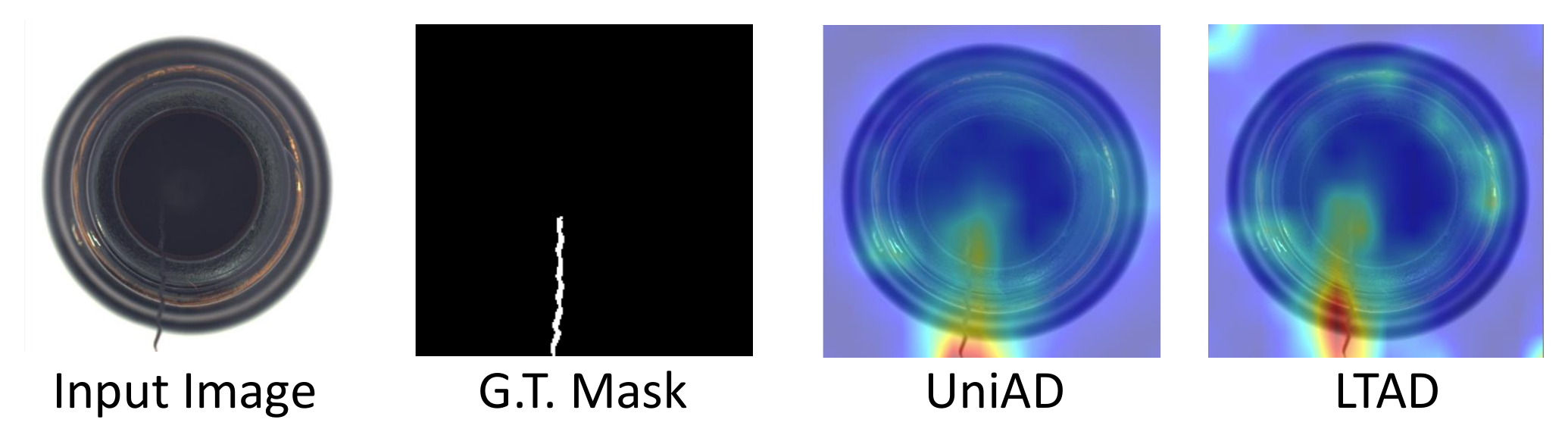}  \\
       \includegraphics[width=0.49\linewidth]{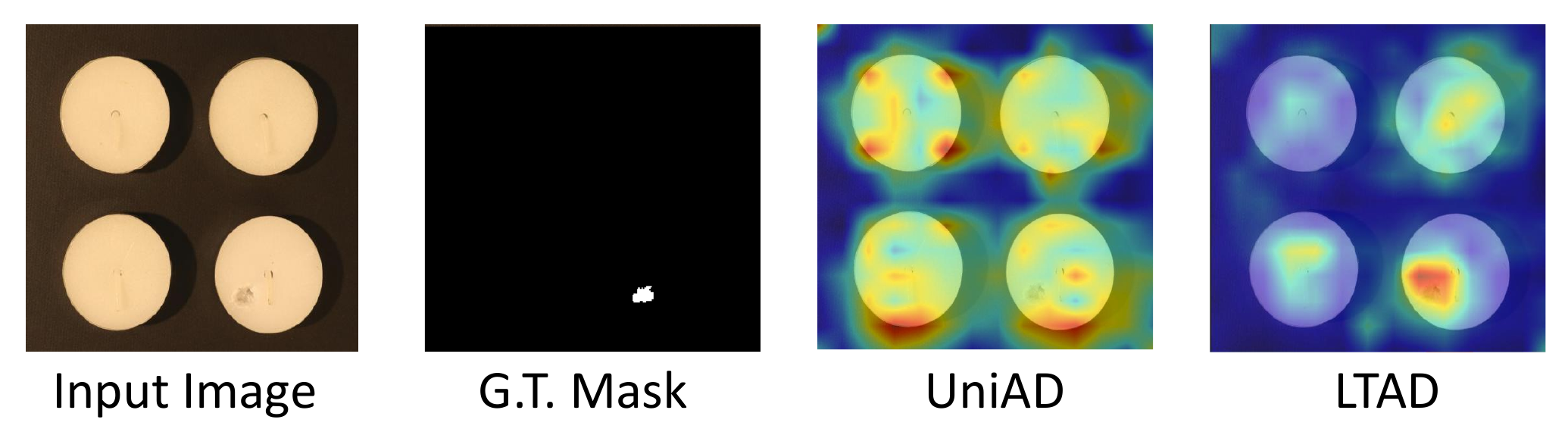} 
       & \includegraphics[width=0.49\linewidth]{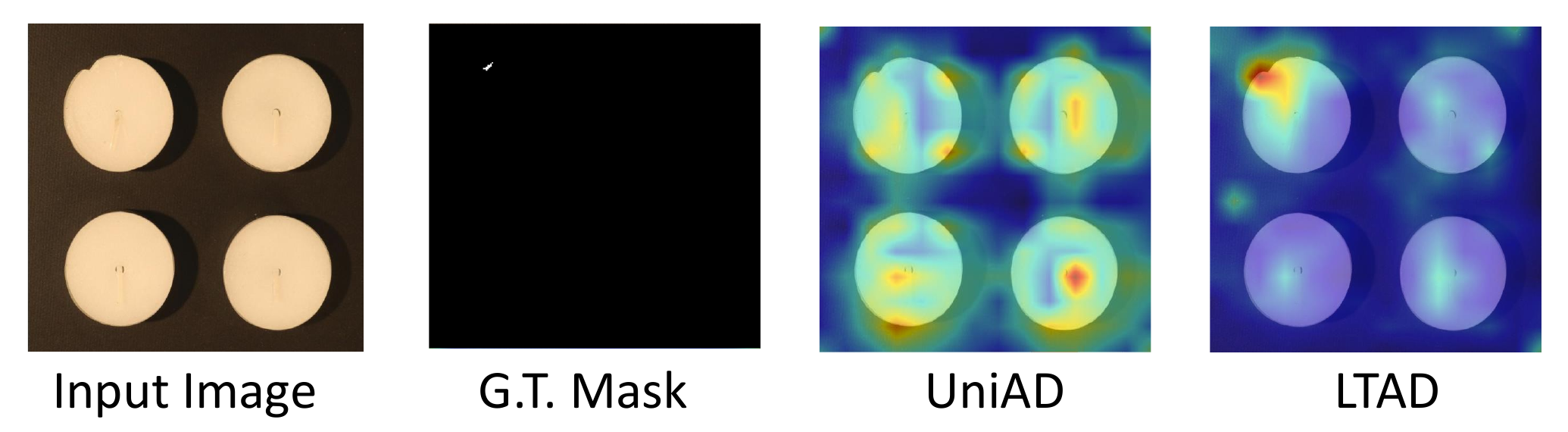} \\
       \includegraphics[width=0.49\linewidth]{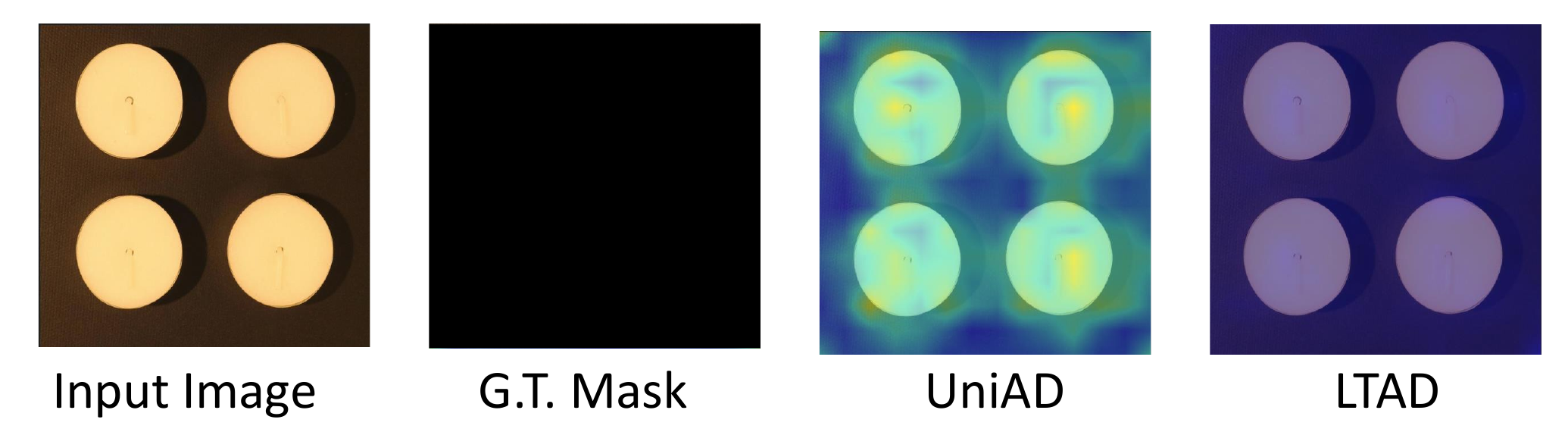}   & \includegraphics[width=0.49\linewidth]{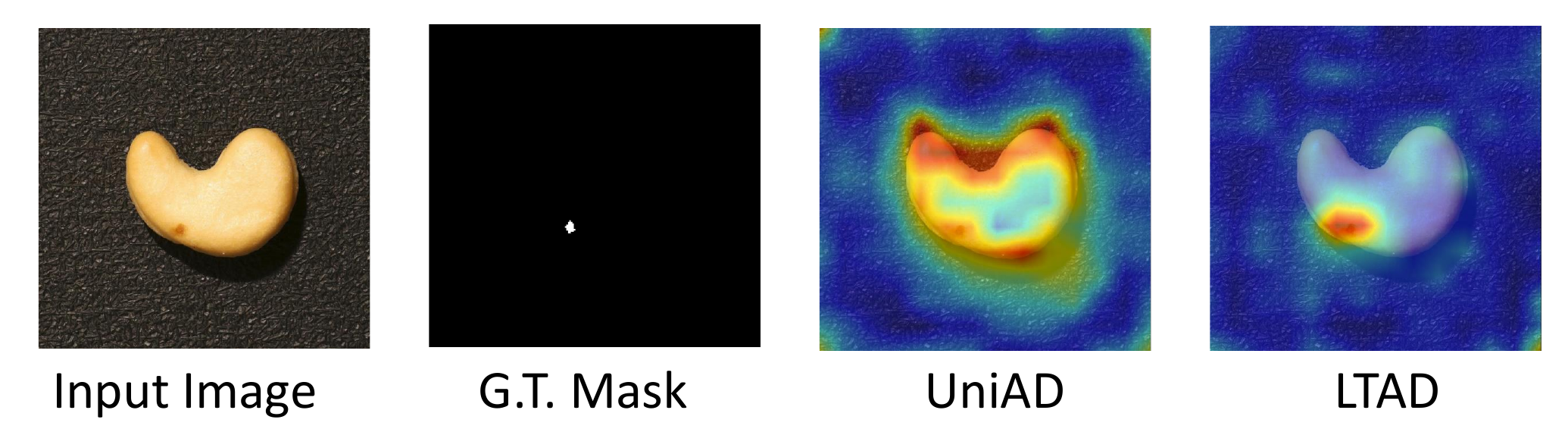} 
       \\ \includegraphics[width=0.49\linewidth]{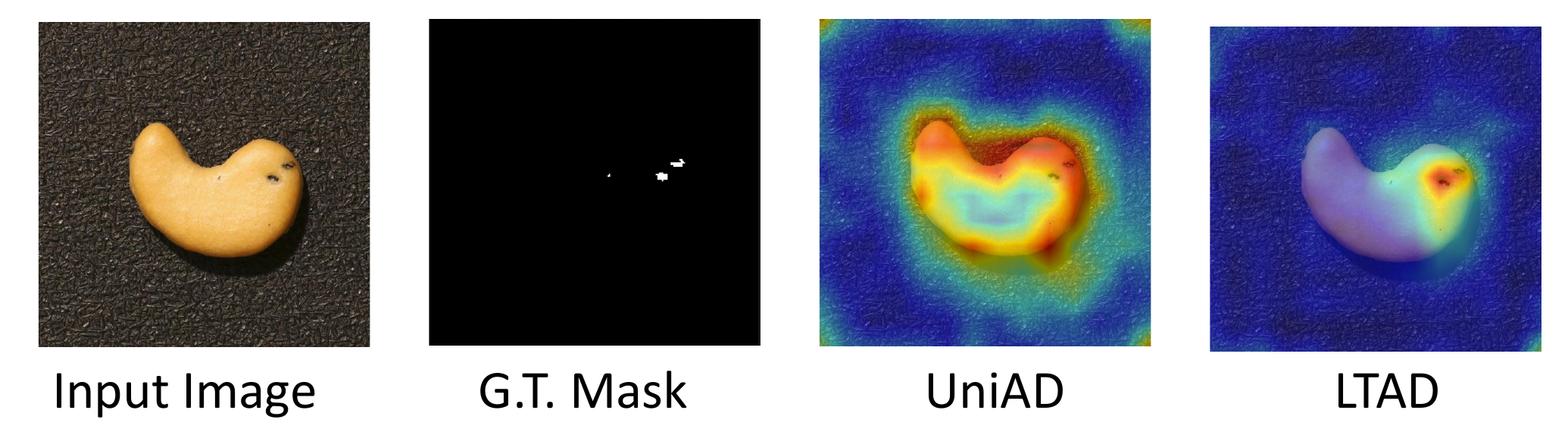} &
       \includegraphics[width=0.49\linewidth]{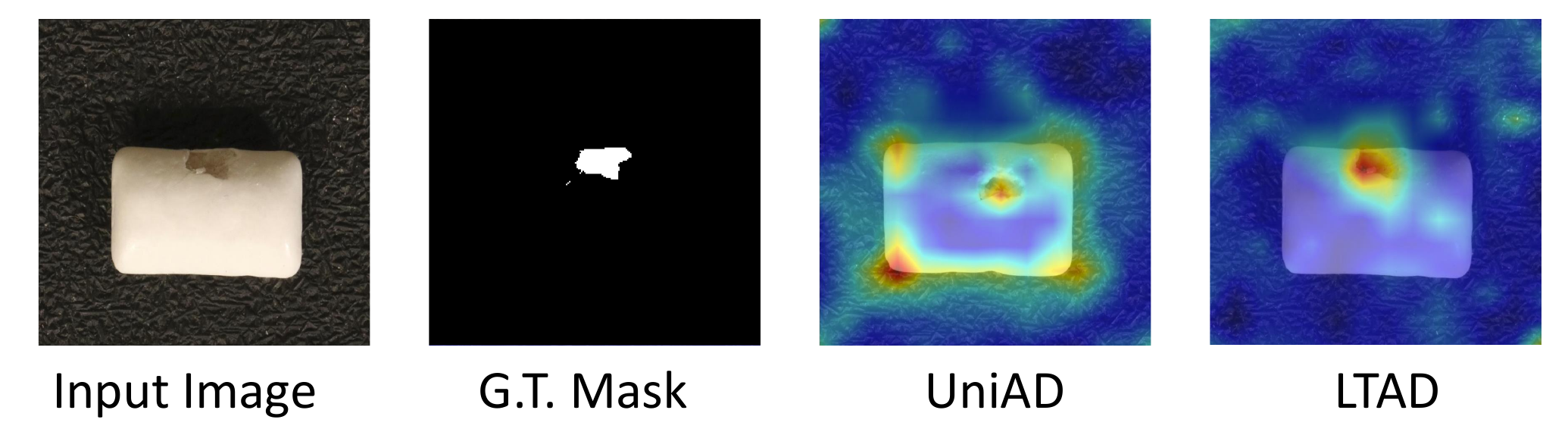}  \\
       \includegraphics[width=0.49\linewidth]{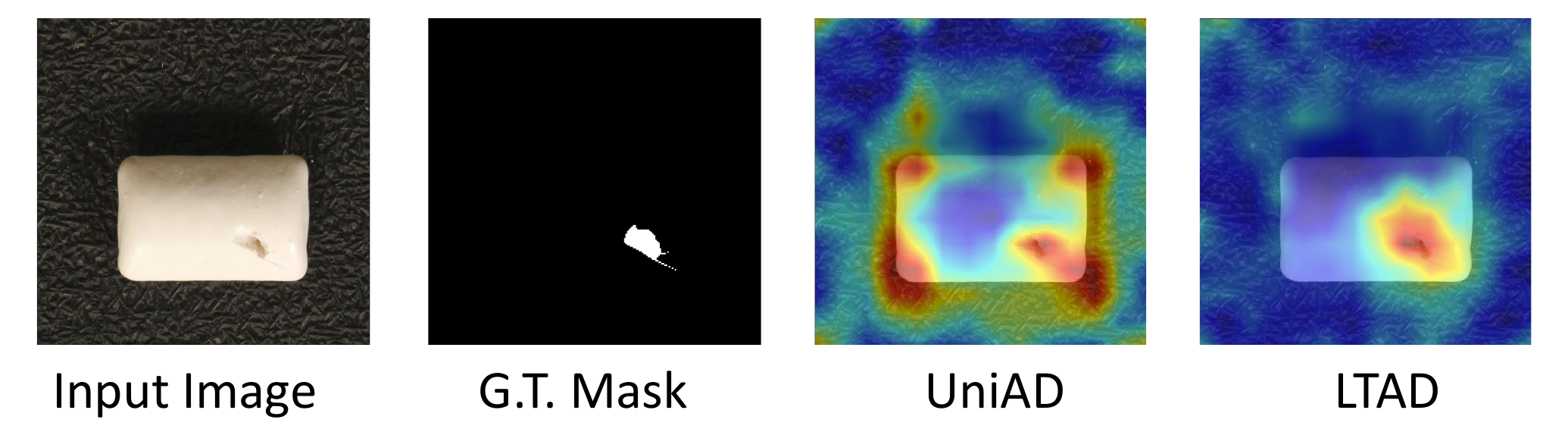} 
       & \includegraphics[width=0.49\linewidth]{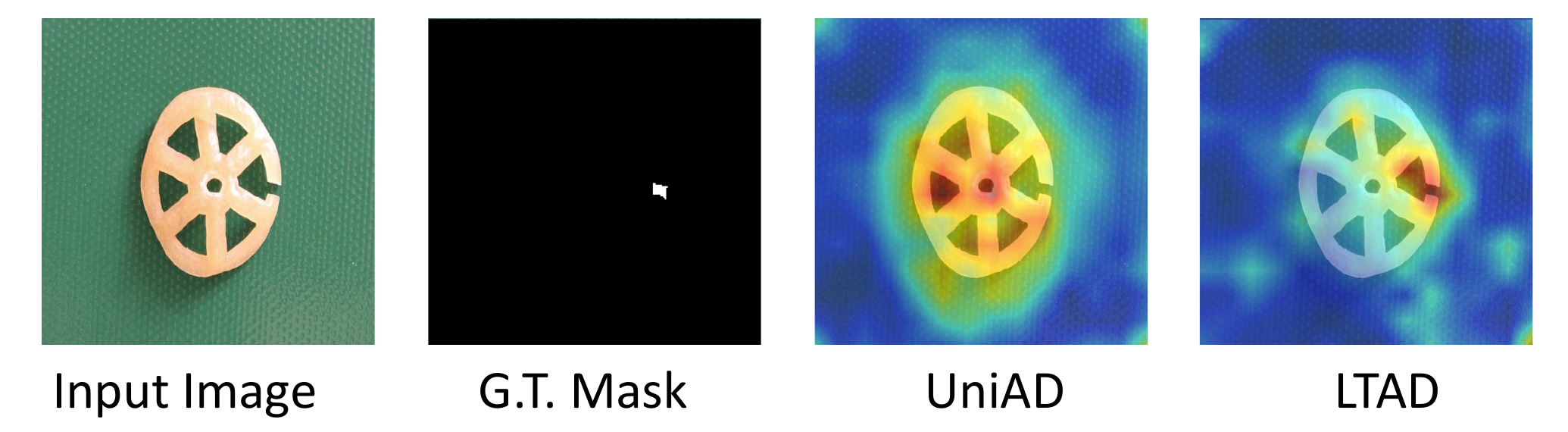} \\
       \includegraphics[width=0.49\linewidth]{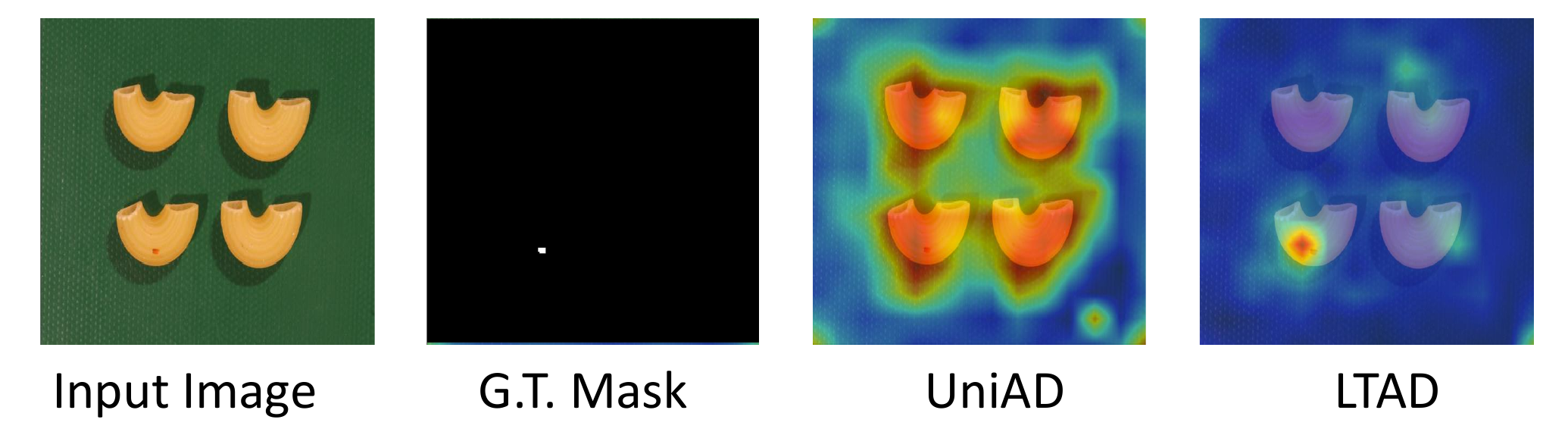}   & \includegraphics[width=0.49\linewidth]{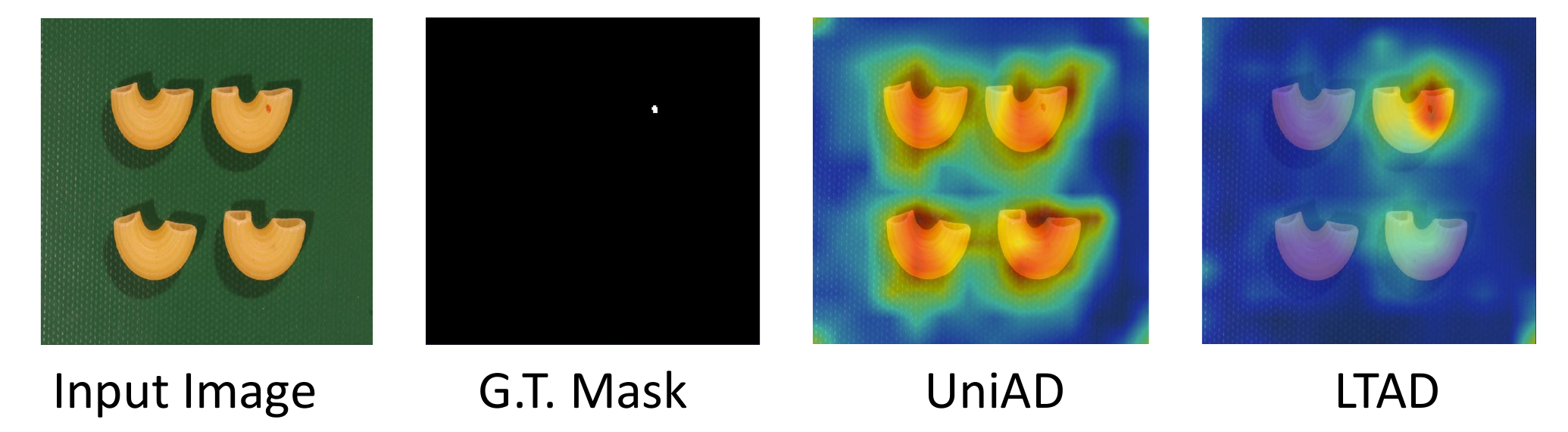} 
       \\ \includegraphics[width=0.49\linewidth]{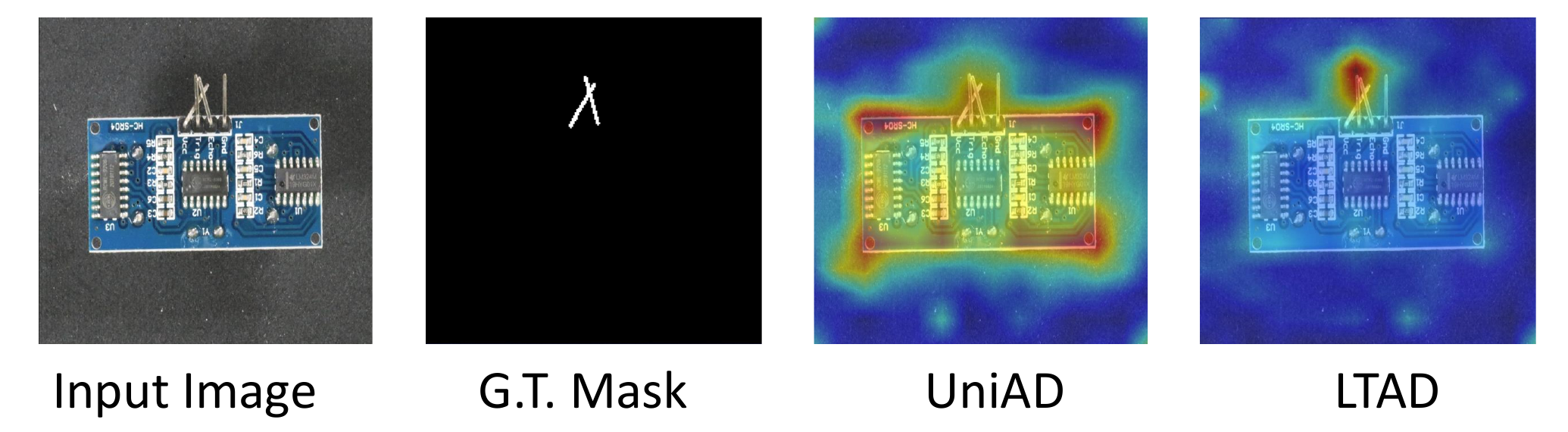} &
       \includegraphics[width=0.49\linewidth]{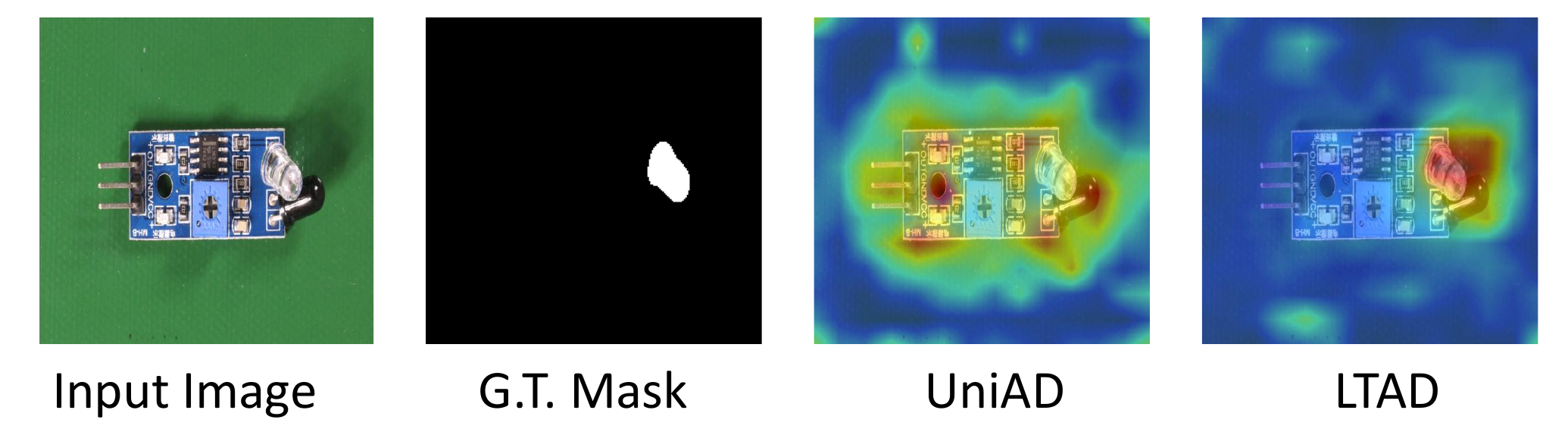}  \\ \includegraphics[width=0.49\linewidth]{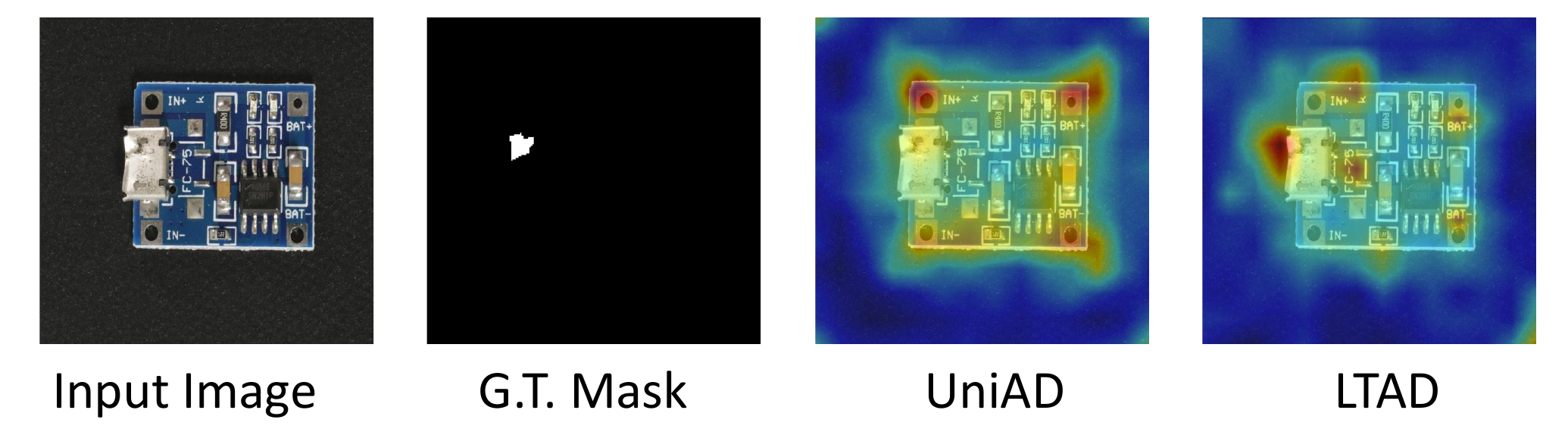} 
       & \includegraphics[width=0.49\linewidth]{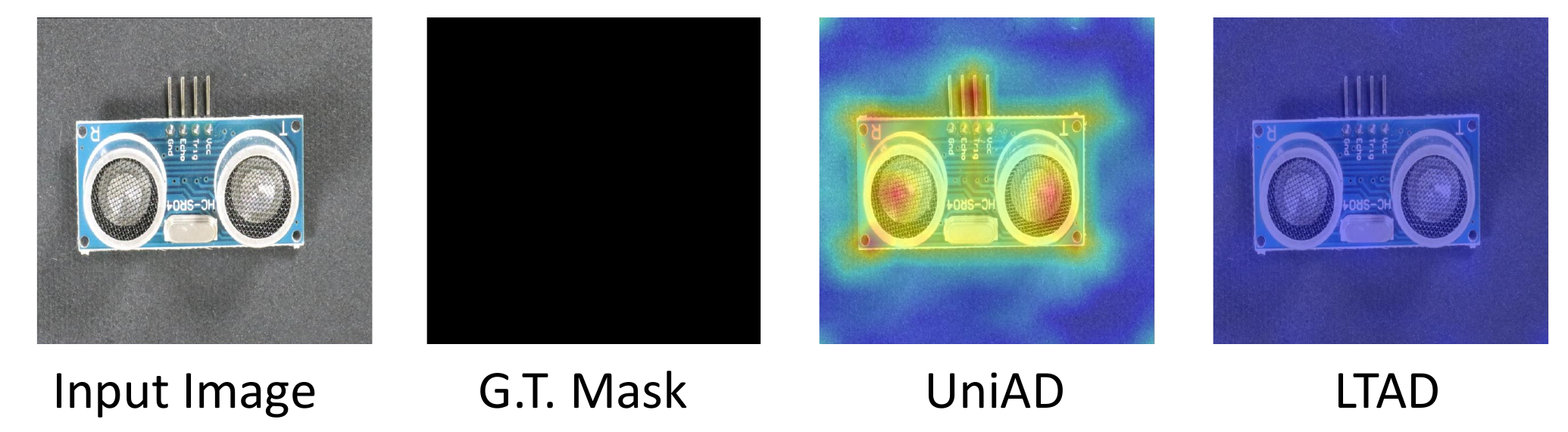} \\
      \hline
    \end{tabular}
    \caption{Qualitative comparison between UniAD and \ours. The images without defects (\ie, normal image) have a black ground truth mask.
    }
    \vspace{-10pt}
    \label{fig:appendix_vis}
\end{figure*}

In addition to the visualization shown in Sec. 6.1, we further provide more visualizations in Fig.~\ref{fig:appendix_vis}. \ours has more precise localization and generalizes to different object classes.

\newpage

\newpage

\begin{table*}
\adjustbox{max width=\textwidth}{
\begin{tabular}{|c||c|c|c||c|c|c|c|c|c|c|c|c|c|c|c|c|c|c|c|} \hline
Method  & All & HighShot & LowShot & HazelNut & Leather & Bottle & Wood & Carpet & Tile & MetalNut & ToothBrush & Zipper & Transistor & Grid & Pill & Capsule & Cable & Screw\\  \hline
Cut \& Paste & 75.89 & 85.79 & 67.23 & 94.64 & 88.85 & 97.93 & 94.29 & 75.72 & 82.90 & 66.22 & 85.55 & 75.23 & 83.62 & 39.09 & 63.39 & 62.46 & 78.82 & 49.72\\ \hline
MKD  & 78.92 & 83.11 & 75.26 & 97.00 & 78.94 & 98.97 & 91.05 & 74.48 & 78.94 & 62.37 & 72.22 & 90.49 & 79.33 & 64.24 & 78.4 & 69.01 & 78.54 & 69.81\\  \hline
DRAEM  & 79.57 & 95.22 & 65.87 & 98.32 & 98.91 & 99.68 & 99.91 & 92.17 & 97.94 & 79.66 & 83.61 & 93.01 & 62.54 & 43.02 & 68.14 & 41.76 & 53.31 & 81.57\\  \hline
RegAD  & 82.43 & 90.04 & 70.51 & 91.84 & 98.61 & 99.2 & 97.86 & 97.44 & 97.59 & 89.77 & 75.69 & 85.58 & 81.68 & 68.80 & 69.75 & 64.67 & 64.67 & 53.30\\  \hline
UniAD & 87.70 & {\bf 99.27} & 77.58 & 100 & 100 & 99.76 & 99.56 & 99.79 & 99.38 & 96.43 & 88.61 & 86.65 & 91.58 & 92.64 & 81.58 & 63.78 & 63.64 & 52.22\\  \hline
AnomalyGPT & 87.44 & 96.40 & 79.60 & 96.21 & 100.0 & 96.75 & 88.07 & 98.19 & 97.66 & 97.95 & 95.0 & 87.16 & 79.77 & 91.98 & 84.44 & 67.03 & 69.33 & 62.1\\  \hline
\hline
Ours w/o PTA & \underline{88.74} & \underline{99.13} & \underline{79.65} & 99.96 & 100 & 99.76 & 99.03 & 99.95 & 98.56 & 96.67 & 89.44 & 85.63 & 93.33 & 96.24 & 81.89 & 64.86 & 69.45 & 56.32\\  \hline
Ours & {\bf 88.86} & 99.09 & {\bf 79.90} & 99.82 & 100 & 99.92 & 99.30 & 99.88 & 99.24 & 95.50 & 87.5 & 89.23 & 93.79 & 87.80 & 84.04 & 64.82 & 77.49 & 54.56\\ 
 \hline
\end{tabular}
}
\caption{MVTec-LT (imbalance factor=100 ; exp decrease ) Image AUROC}
\label{tab:mvtec_exp100_ad}
\end{table*}

\begin{table*}
\adjustbox{max width=\textwidth}{
\begin{tabular}{|c||c|c|c||c|c|c|c|c|c|c|c|c|c|c|c|c|c|c|c|} \hline
Method  & All & HighShot & LowShot & HazelNut & Leather & Bottle & Wood & Carpet & Tile & MetalNut & ToothBrush & Zipper & Transistor & Grid & Pill & Capsule & Cable & Screw\\  \hline
MKD &  85.95 & 87.39 & 84.69 & 95.00 & 96.87 & 93.10 & 77.74 & 93.05 & 76.84 & 79.13 & 93.86 & 90.14 & 73.69 & 72.36 & 88.75 & 93.16 & 71.67 & 93.95\\ \hline
DRAEM & 85.17 & 93.68 & 77.73 & 97.80 & 97.56 & 95.18 & 97.33 & 94.33 & 97.36 & 76.21 & 97.34 & 97.43 & 66.95 & 76.55 & 89.00 & 44.65 & 59.90 & 90.08
\\ \hline
RegAD  & \textbf{95.2} & \textbf{96.92} & \textbf{93.69} & 98.07 & 99.15 & 98.12 & 95.46 & 98.67 & 94.18 & 94.81 & 96.88 & 97.12 & 92.44 & 79.05 & 97.29 & 96.42 & 96.42 & 93.96
\\ \hline
UniAD  & 93.95 & 95.46 & 92.64 & 97.83 & 98.79 & 95.64 & 90.66 & 98.52 & 91.65 & 95.18 & 97.96 & 92.35 & 96.38 & 93.80 & 88.66 & 94.68 & 88.93 & 88.36\\ \hline
AnomalyGPT & 89.68 & 93.64 & 86.21 & 92.06 & 99.58 & 93.42 & 89.47 & 98.97 & 94.00 & 88.02 & 97.06 & 93.22 & 67.18 & 94.33 & 75.00 & 87.84 & 85.00 & 89.17
\\ \hline
\hline
Ours w/o PTA & 94.00 & 95.27 & 92.89 & 97.67 & 98.83 & 95.56 & 89.51 & 98.49 & 91.22 & 95.64 & 98.06 & 92.28 & 97.07 & 93.69 & 88.71 & 94.87 & 90.57 & 87.90\\ \hline
Ours & \underline{94.46} & \underline{95.96} & \underline{93.15} & 98.12 & 99.29 & 96.84 & 90.45 & 98.52 & 92.52 & 95.96 & 97.91 & 93.93 & 96.89 & 93.76 & 90.09 & 95.02 & 91.67 & 85.92\\ \hline

\end{tabular}
}
\caption{MVTec-LT (imbalance factor=100 ; exp decrease ) Pixel AUROC}
\label{tab:mvtec_exp100_as}
\end{table*}

\begin{table*}
\adjustbox{max width=\textwidth}{
\begin{tabular}{|c||c|c|c||c|c|c|c|c|c|c|c|c|c|c|c|c|c|c|c|} \hline
Method  & All & HighShot & LowShot & HazelNut & Leather & Bottle & Wood & Carpet & Tile & MetalNut & ToothBrush & Zipper & Transistor & Grid & Pill & Capsule & Cable & Screw\\  \hline
Cut \& Paste  & 75.07 & 85.10 & 66.29 & 94.39 & 88.75 & 97.30 & 94.38 & 75.08 & 80.80 & 65.05 & 83.05 & 74.23 & 82.66 & 41.52 & 62.90 & 59.51 & 77.69 & 48.80\\ \hline
MKD & 79.93 & 84.47 & 75.96 & 96.93 & 78.19 & 98.97 & 91.23 & 74.36 & 86.98 & 64.66 & 72.78 & 89.97 & 79.08 & 67.84 & 76.7 & 70.56 & 78.84 & 71.9\\ 
\hline
DRAEM & 78.82 & 96.39 & 63.44 & 97.10 & 99.89 & 99.76 & 99.91 & 95.78 & 98.05 & 84.26 & 80.0 & 88.07 & 60.0 & 46.70 & 50.19 & 33.58 & 54.96 & 94.09\\ 
\hline
UniAD & 86.21 & {\bf 99.16} & 74.89 & 99.85 & 100 & 100 & 97.98 & 99.75 & 99.89 & 96.67 & 85.83 & 96.11 & 74.54 & 87.55 & 70.07 & 64.42 & 57.74 & 62.88\\ 
\hline
AnomalyGPT & 85.80 & 97.01 & \underline{76.00} & 98.57 & 100.0 & 96.59 & 89.04 & 98.92 & 97.84 & 98.09 & 88.61 & 86.52 & 77.06 & 90.14 & 88.54 & 62.76 & 57.01 & 57.38\\ 
\hline
\hline
Ours w/o PTA & {\bf 86.94} & 98.76 & {\bf 76.61} & 98.93 & 99.90 & 99.76 & 99.21 & 100 & 98.56 & 94.97 & 88.61 & 83.95 & 75.46 & 96.07 & 83.09 & 65.86 & 65.18 & 54.62\\ 
\hline
Ours & \underline{86.05} & \underline{98.99} & 74.73 & 99.29 & 99.93 & 99.76 & 99.56 & 99.96 & 99.39 & 95.06 & 86.94 & 85.56 & 79.67 & 88.47 & 81.91 & 64.42 & 58.62 & 52.22\\ 
 \hline
\end{tabular}
}
\caption{MVTec-LT (imbalance factor=200 ; exp decrease ) Image AUROC}
\label{tab:mvtec_exp200_ad}
\end{table*}

\begin{table*}
\adjustbox{max width=\textwidth}{
\begin{tabular}{|c||c|c|c||c|c|c|c|c|c|c|c|c|c|c|c|c|c|c|c|} \hline
Method  & All & HighShot & LowShot & HazelNut & Leather & Bottle & Wood & Carpet & Tile & MetalNut & ToothBrush & Zipper & Transistor & Grid & Pill & Capsule & Cable & Screw\\  \hline
MKD &  86.01 & 87.15 & 85.02 & 94.01 & 96.81 & 93.02 & 77.79 & 93.25 & 77.04 & 78.10 & 93.53 & 90.26 & 75.28 & 72.88 & 88.81 & 93.16 & 72.45 & 93.76\\ \hline
DRAEM & 82.95 & 93.98 & 73.30 & 96.96 & 99.15 & 94.34 & 95.62 & 95.25 & 95.34 & 81.24 & 96.43 & 96.15 & 59.05 & 77.78 & 83.77 & 38.02 & 44.28 & 90.94\\ \hline
UniAD & 93.26 & \underline{95.91} & 90.95 & 98.03 & 98.67 & 97.91 & 93.67 & 98.48 & 91.97 & 92.68 & 98.23 & 96.11 & 89.44 & 89.96 & 86.62 & 96.39 & 83.33 & 87.55\\ \hline
AnomalyGPT  & 90.15 & 94.01 & 86.76 & 92.47 & 99.54 & 94.34 & 89.7 & 98.96 & 94.62 & 88.48 & 96.87 & 92.39 & 67.64 & 94.95 & 76.63 & 89.01 & 85.46 & 91.16\\ \hline
\hline
Ours w/o PTA & \underline{93.4} & 95.07 & \underline{91.94} & 97.43 & 98.78 & 95.41 & 90.49 & 98.62 & 90.98 & 93.75 & 97.93 & 92.39 & 93.69 & 93.50 & 90.90 & 95.16 & 90.79 & 81.14\\ \hline
Ours & {\bf 94.18} & {\bf 96.25} & {\bf 92.37} & 97.87 & 99.04 & 96.79 & 92.16 & 98.65 & 94.23 & 95 & 97.83 & 92.86 & 93.45 & 93.85 & 93.11 & 95.37 & 85.92 & 86.55\\ \hline
\end{tabular}
}
\caption{MVTec-LT (imbalance factor=200 ; exp decrease ) Pixel AUROC}
\label{tab:mvtec_exp200_as}
\end{table*}

\begin{table*}
\adjustbox{max width=\textwidth}{
\begin{tabular}{|c||c|c|c||c|c|c|c|c|c|c|c|c|c|c|c|c|c|c|c|} \hline
Method  & All & HighShot & LowShot & HazelNut & Leather & Bottle & Wood & Carpet & Tile & MetalNut & ToothBrush & Zipper & Transistor & Grid & Pill & Capsule & Cable & Screw\\  \hline
Cut \& Paste & 76.57 & 89.01 & 65.68 & 93.49 & 89.53 & 97.93 & 93.85 & 78.57 & 89.97 & 79.76 & 86.38 & 69.56 & 80.16 & 35.92 & 62.71 & 63.46 & 78.03 & 49.29\\ \hline
MKD  & 79.61 & 84.34 & 75.46 & 96.61 & 78.43 & 98.81 & 91.32 & 75.96 & 85.14 & 64.13 & 76.39 & 90.36 & 79.33 & 62.91 & 78.18 & 69.57 & 78.86 & 68.15\\ \hline
DRAEM & 69.82 & 84.00 & 57.41 & 99.96 & 95.78 & 99.44 & 100 & 97.47 & 99.67 & 94.72 & 72.77 & 70.79 & 51.12 & 28.15 & 57.50 & 37.89 & 50.50 & 90.57\\ 
 \hline
RegAD & 81.54 & 96.34 & 68.59 & 90.53 & 100 & 99.76 & 98.50 & 95.94 & 95.90 & 93.76 & 70.27 & 74.69 & 83.70 & 66.41 & 68.97 & 64.33 & 63.05 & 57.31\\ 
\hline
UniAD  & 83.37 & \underline{99.52} & 69.24 & 99.92 & 100 & 100 & 98.07 & 99.67 & 99.81 & 99.21 & 80.55 & 91.46 & 57.75 & 80.45 & 66.50 & 61.38 & 58.30 & 57.55\\ \hline
AnomalyGPT & 85.95 & 98.69 & 74.79 & 98.93 & 100.0 & 97.94 & 95.96 & 99.6 & 99.39 & 99.07 & 88.06 & 86.19 & 70.69 & 92.65 & 86.82 & 61.47 & 51.2 & 61.28\\ \hline
\hline
Ours w/o PTA & \underline{87.05} & 99.07 & \underline{76.54} & 99.5 & 99.93 & 99.84 & 98.94 & 100 & 98.59 & 96.67 & 79.44 & 82.27 & 86.16 & 97.15 & 80.38 & 63.10 & 66.58 & 57.26\\ \hline
Ours & {\bf 87.36} & {\bf 99.33} & {\bf 76.89} & 99.79 & 99.93 & 99.76 & 99.12 & 99.96 & 99.17 & 97.61 & 78.89 & 86.50 & 87.17 & 90.48 & 74.96 & 66.61 & 76.31 & 54.21\\  \hline
\end{tabular}
}
\caption{MVTec-LT (imbalance factor=100 ; step decrease ) Image AUROC}
\label{tab:mvtec_step100_ad}
\end{table*}

\begin{table*}
\adjustbox{max width=\textwidth}{
\begin{tabular}{|c||c|c|c||c|c|c|c|c|c|c|c|c|c|c|c|c|c|c|c|} \hline
Method  & All & HighShot & LowShot & HazelNut & Leather & Bottle & Wood & Carpet & Tile & MetalNut & ToothBrush & Zipper & Transistor & Grid & Pill & Capsule & Cable & Screw\\  \hline
MKD & 85.90 & 86.92 & 85.01 & 94.61 & 96.24 & 92.63 & 77.85 & 93.20 & 76.96 & 77.01 & 93.87 & 90.92 & 73.73 & 72.96 & 89.09 & 93.41 & 72.46 & 93.60\\ \hline
DRAEM & 79.65 & 92.81 & 69.79 & 9454 & 9787 & 8975 & 9511 & 9808 & 9886 & 8053 & 9082 & 7491 & 6030 & 5084 & 7696 & 6539 & 5387 & 8526\\ \hline
RegAD & {\bf 95.10} & {\bf 97.25} & {\bf 93.21} & 98.24 & 99.22 & 97.92 & 95.87 & 98.37 & 94.37 & 96.78 & 95.69 & 95.71 & 93.16 & 79.73 & 96.53 & 95.91 & 94.54 & 94.48\\ \hline
UniAD & 91.47 & \underline{96.15} & 87.38 & 97.92 & 98.54 & 97.83 & 93.48 & 98.38 & 92.37 & 94.57 & 95.59 & 93.79 & 75.76 & 84.46 & 82.02 & 94.43 & 81.96 & 91.04\\ \hline
AnomalyGPT & 89.28 & 93.62 & 85.49 & 88.41 & 99.56 & 94.05 & 89.61 & 99.08 & 95.52 & 89.09 & 95.53 & 92.65 & 65.87 & 94.42 & 79.26 & 86.48 & 83.11 & 86.62\\ \hline
\hline
Ours w/o PTA & 93.13 & 95.07 & 91.44 & 97.77 & 98.74 & 95.39 & 90.55 & 98.63 & 89.04 & 95.34 & 97.63 & 89.63 & 93.78 & 92.22 & 90.01 & 94.85 & 90.98 & 82.40\\ \hline
Ours & \underline{93.83} & 95.90 & \underline{92.01} & 98.24 & 99.01 & 95.69 & 91.46 & 98.68 & 91.82 & 96.43 & 96.71 & 90.17 & 93.68 & 91.90 & 91.10 & 95.73 & 91.8 & 85.02\\ \hline
\end{tabular}
}
\caption{MVTec-LT (imbalance factor=100 ; step decrease ) Pixel AUROC}
\label{tab:mvtec_step100_as}
\end{table*}

\begin{table*}
\adjustbox{max width=\textwidth}{
\begin{tabular}{|c||c|c|c||c|c|c|c|c|c|c|c|c|c|c|c|c|c|c|c|} \hline
Method  & All & HighShot & LowShot & HazelNut & Leather & Bottle & Wood & Carpet & Tile & MetalNut & ToothBrush & Zipper & Transistor & Grid & Pill & Capsule & Cable & Screw\\  \hline
Cut \& Paste  & 76.53 & 89.07 & 65.56 & 94.17 & 90.28 & 97.85 & 93.94 & 78.49 & 89.89 & 78.88 & 86.66 & 69.38 & 79.37 & 35.25 & 62.52 & 62.74 & 78.18 & 50.39\\ \hline
MKD  & 79.31 & 84.48 & {\bf 74.79} & 97.04 & 80.03 & 99.05 & 91.58 & 75.52 & 85.21 & 62.9 & 72.78 & 90.7 & 77.67 & 64.75 & 76.38 & 69.45 & 78.94 & 67.7\\ \hline
DRAEM & 71.64 & 90.90 & 54.79 & 96.96 & 96.39 & 86.50 & 98.24 & 94.98 & 97.58 & 65.68 & 64.44 & 46.90 & 39.29 & 57.81 & 63.39 & 40.76 & 51.55 & 74.21\\ 
\hline
UniAD  & 81.32 & {\bf 99.61} & 65.31 & 100 & 100 & 100 & 98.15 & 99.39 & 99.92 & 99.85 & 81.11 & 82.32 & 54.58 & 86.38 & 53.19 & 64.22 & 50.52 & 50.21\\ \hline
AnomalyGPT & 82.47 & 98.07 & 68.82 & 98.82 & 100.0 & 97.3 & 92.81 & 99.8 & 98.45 & 99.32 & 73.89 & 88.46 & 64.0 & 92.77 & 84.79 & 48.64 & 50.67 & 47.39\\ \hline
\hline
Ours w/o PTA & \underline{85.33} & 98.88 & 73.48 & 98.57 & 99.93 & 99.84 & 98.94 & 100 & 98.48 & 96.43 & 75.83 & 79.67 & 82.41 & 95.82 & 78.58 & 59.47 & 65.04 & 51.05\\ 
\hline
Ours & {\bf 85.60} & \underline{99.15} & \underline{73.74}  & 99.43 & 99.93 & 99.68 & 99.39 & 99.80 & 99.21 & 96.58 & 75 & 85.9 & 81.71 & 90.48 & 76.68 & 63.98 & 65.54 & 50.65\\ 
\hline
\end{tabular}
}
\caption{MVTec-LT (imbalance factor=200 ; step decrease ) Image AUROC}
\label{tab:mvtec_step200_ad}
\end{table*}

\begin{table*}
\adjustbox{max width=\textwidth}{
\begin{tabular}{|c||c|c|c||c|c|c|c|c|c|c|c|c|c|c|c|c|c|c|c|} \hline
Method  & All & HighShot & LowShot & HazelNut & Leather & Bottle & Wood & Carpet & Tile & MetalNut & ToothBrush & Zipper & Transistor & Grid & Pill & Capsule & Cable & Screw\\  \hline
MKD  & 86.03 & 87.22 & 84.99 & 94.80 & 96.73 & 93.34 & 78.61 & 93.33 & 76.63 & 77.16 & 93.93 & 90.73 & 73.95 & 74.24 & 88.43 & 93.35 & 71.44 & 93.86\\ \hline
DRAEM & 76.79 & 84.23 & 70.29 & 95.21 & 88.39 & 76.81 & 82.64 & 94.62 & 89.27 & 62.70 & 89.65 & 70.96 & 56.29 & 57.39 & 85.65 & 54.66 & 56.30 & 91.44\\ \hline
UniAD & 89.29 & {\bf 96.10} & 83.34 & 97.93 & 98.57 & 97.83 & 93.46 & 98.36 & 92.25 & 94.32 & 92.69 & 90.73 & 66.60 & 76.28 & 78.25 & 92.27 & 79.87 & 90.05\\ \hline
AnomalyGPT & 89.45 & 93.75 & 85.69 & 88.29 & 99.56 & 95.12 & 89.94 & 99.0 & 95.49 & 88.82 & 94.98 & 91.61 & 66.4 & 93.55 & 78.7 & 90.56 & 82.42 & 87.33\\ \hline
\hline
Ours w/o PTA & \underline{91.78} & 94.62 & {\bf 89.30} & 97.21 & 98.74 & 95.32 & 89.99 & 98.61 & 87.37 & 95.10 & 96.77 & 89.74 & 90.31 & 92.07 & 88.95 & 89.33 & 84.70 & 82.54\\ \hline
Ours & {\bf 92.12} & \underline{95.45} & \underline{89.20} & 97.83 & 99.06 & 95.76 & 91.00 & 98.72 & 90.59 & 95.22 & 96.45 & 89.00 & 84.38 & 89.24 & 89.94 & 92.37 & 84.92 & 87.32\\
\hline
\end{tabular}
}
\caption{MVTec-LT (imbalance factor=200 ; step decrease ) Pixel AUROC}
\label{tab:mvtec_step200_as}
\end{table*}

\begin{table*}
\adjustbox{max width=\textwidth}{
\begin{tabular}{|c||c|c|c||c|c|c|c|c|c|c|c|c|c|c|c|c|c|} \hline
Method  & All & HighShot & LowShot & PCB3 & PCB2 & PCB1  & PCB4 & macaroni1 & macaroni2 & candle & cashew & fryum & capsules
 & chewinggum & pipe fryum \\  \hline
AnomalyGPT & 70.34 & 70.13 & 70.56 & 62.69 & 78.55 & 79.57 & 69.66 & 78.76 & 51.54 & 49.
96 & 77.15 & 80.77 & 63.88 & 71.76 & 79.84 \\ \hline
RegAD & 71.36 & 66.39 & 76.34 & 62.71 & 63.79 & 73.89 & 82.82 & 62.37 & 52.79 & 76.12 & 71.99 & 73.82 & 56.07 & 92.92 & 87.1
3 \\ \hline

UniAD  & 77.31 & \textbf{77.42} & 77.19 & 78.34 & 84.69 & 82.02 & 95.96 & 70.68 & 52.88 & 87.1 & 81.44 & 68.6 & 52.75 & 95.54 & 77.74 
\\ \hline
\hline
Ours w/o PTA & \underline{79.27} & 75.32 & \underline{83.21} & 82.38 & 82.68 & 82.38 & 97.12 & 54.79 & 52.57 & 83.
27 & 79.02 & 86.06 & 66.37 & 94.64 & 89.9 
\\ \hline
Ours & \textbf{80.00} & \underline{76.29} & \textbf{83.71} & 82.18 & 81.42 & 85.02 & 93.63 & 56.46 & 59.02 & 84.46 & 82.46 & 85.7 & 69.37 & 96.16 & 84.1
\\ \hline
\end{tabular}
}
\caption{VisA-LT (imbalance factor=100 ; exp decrease ) Image AUROC}
\label{tab:visa_exp100_ad}
\end{table*}

\begin{table*}
\adjustbox{max width=\textwidth}{
\begin{tabular}{|c||c|c|c||c|c|c|c|c|c|c|c|c|c|c|c|c|c|} \hline
Method  & All & HighShot & LowShot & PCB3 & PCB2 & PCB1  & PCB4 & macaroni1 & macaroni2 & candle & cashew & fryum & capsules
 & chewinggum & pipe fryum \\  \hline
AnomalyGPT & 80.32 & 77.43 & 83.20 & 79.41
 & 84.8 & 77.09 & 82.17 & 70.68 & 70.44
 & 77.41 & 86.7 & 90.21 & 62.72 & 92.1 & 90.0
7 \\ \hline
RegAD & 94.4  & 93.72 & 95.07 & 95.27
 & 92.58 & 97.26 & 91.82 & 94.88 & 90.53 & 95.03 & 96.82 & 94.97 & 87.20 & 97.37 & 99.07 \\ \hline
UniAD  & 95.03 & \underline{94.57} & 95.48 & 97.05 & 96.12 & 91.61 & 96.80 & 97.04 & 88.84 & 98.19 & 98.82 & 95.25 & 83.21 & 98.68  & 98.78 
\\ \hline
\hline
Ours w/o PTA & \underline{95.07} & 94.41 & \underline{95.72}
 & 97.62 & 95.92 & 90.09 & 95.73 & 96.41 & 90.71
 & 97.73 & 99.01 & 96.18 & 83.84 & 98.75  & 98.83
\\ \hline
Ours & \textbf{95.56} & \textbf{95.16} & \textbf{95.97} & 97.62 & 96.06 & 92.18 & 95.96 & 96.58 & 92.58 & 97.90 & 99.08 & 95.39 & 85.77 & 98.82 & 98.83

\\ \hline
\end{tabular}
}
\caption{VisA-LT (imbalance factor=100 ; exp decrease ) Pixel AUROC}
\label{tab:visa_exp100_as}
\end{table*}

\begin{table*}
\adjustbox{max width=\textwidth}{
\begin{tabular}{|c||c|c|c||c|c|c|c|c|c|c|c|c|c|c|c|c|c|} \hline
Method  & All & HighShot & LowShot & PCB3 & PCB2 & PCB1  & PCB4 & macaroni1 & macaroni2 & candle & cashew & fryum & capsules
 & chewinggum & pipe fryum \\  \hline
AnomalyGPT & 69.78 & 69.22 & 70.35 & 61.64 & 75.82 & 79.42 & 68.41  & 79.6  & 50.4 & 45.9 & 76.59  & 79.8 & 63.34 & 72.83  & 83.63 \\ \hline
RegAD & 72.10 & 67.19  & 77.02 & 61.92 & 64.41 & 73.16  & 79.95 & 67.55 & 56.16 & 76.53 & 77.21 & 79.73  & 55.81  & 88.92  & 83.94 \\ \hline
UniAD  & 76.87 & \underline{76.25} & 77.49 & 78.91 & 84.96
 & 81.85 & 96.34 & 62.01  & 53.44 & 86.91  & 78.
6  & 74.84 & 54.08 & 94.28 & 76.28
\\ \hline
\hline
Ours w/o PTA & \underline{78.55} & 75.38 & \underline{81.73}
 & 83.93  & 83.05 & 83.29  & 96.60  & 54.15
 & 51.23  & 82.47 & 80.66  & 82.1 & 60.75  & 95.16 & 89.26
\\ \hline
Ours & \textbf{80.21} & \textbf{76.81} & \textbf{83.60} & 84.98 & 83.77 & 86.86 & 95.37 & 57.66 & 52.2 & 85.73 & 81.82 & 83.3 & 63.23 & 96.82 & 90.72

\\ \hline
\end{tabular}
}
\caption{VisA-LT (imbalance factor=200 ; exp decrease ) Image AUROC}
\label{tab:visa_exp200_ad}
\end{table*}

\begin{table*}
\adjustbox{max width=\textwidth}{
\begin{tabular}{|c||c|c|c||c|c|c|c|c|c|c|c|c|c|c|c|c|c|} \hline
Method  & All & HighShot & LowShot & PCB3 & PCB2 & PCB1  & PCB4 & macaroni1 & macaroni2 & candle & cashew & fryum & capsules
 & chewinggum & pipe fryum \\  \hline
AnomalyGPT & 79.48 & 76.64 & 82.32 & 77.59 & 83.31 & 76.89 & 81.92 & 68.29 & 71.84 & 77.66 & 83.39 & 89.57 & 61.44 & 92.37 & 89.46
 \\ \hline
RegAD & 94.69 & 94.07 & \underline{95.30} & 95.78 & 94.22 & 97.87 & 90.41 & 96.18 & 90.01 & 95.64 & 96.80 & 95.89 & 87.54 & 96.73 & 99.23
\\ \hline
UniAD  & \underline{94.80} & \underline{94.34} & 95.26 & 97.07 & 96.06 & 91.64 & 96.53 & 96.18 & 88.56 & 98.06 & 98.61 & 95.24 & 82.72 & 98.32 & 98.66
\\ \hline
\hline
Ours w/o PTA & 94.51 & 93.87 & 95.15 & 97.52
 & 95.81 & 89.76 & 95.47 & 95.64 & 88.99
 & 97.53 & 97.53 & 94.93 & 81.82 & 98.84 & 98.80
\\ \hline
Ours & \textbf{95.36} & \textbf{94.98} & \textbf{95.69} & 97.66 & 95.03 & 91.56 & 96.42 & 96.20 & 89.58 & 98.31 & 99.02 & 95.16 & 83.74 & 98.87 & 99.02
\\ \hline
\end{tabular}
}
\caption{VisA-LT (imbalance factor=200 ; exp decrease ) Pixel AUROC}
\label{tab:visa_exp200_as}
\end{table*}

\begin{table*}
\adjustbox{max width=\textwidth}{
\begin{tabular}{|c||c|c|c||c|c|c|c|c|c|c|c|c|c|c|c|c|c|} \hline
Method  & All & HighShot & LowShot & PCB3 & PCB2 & PCB1  & PCB4 & macaroni1 & macaroni2 & candle & cashew & fryum & capsules
 & chewinggum & pipe fryum \\  \hline
AnomalyGPT & 68.18 & 68.35 & 68.00 & 60.83 & 76.03 & 78.94 & 66.37 & 77.5 & 50.42 & 51 & 74.71 & 72.72 & 61.73 & 73.85 & 74 \\ \hline
UniAD  & 73.67 & \textbf{76.61} & 70.73 & 79.17 & 85.11 & 82.97 & 96.38 & 62.94 & 53.13 & 85.7 & 65.38 & 76.92 & 53.11 & 89.22 & 54.06
\\ \hline
\hline
Ours w/o PTA & \underline{77.25} & 76.27 & \underline{78.24} & 86.13 & 83.76 & 82.94 & 96.91 & 51.44 & 56.44 & 80.14 & 77.6 & 74.46 & 60.1 & 92.24 & 84.88
\\ \hline
Ours & \textbf{78.53} & \underline{76.56} & \textbf{80.50} & 84.22 & 83.85 & 87.05 & 94.55 & 50.11 & 59.56 & 81.07 & 77.84 & 77.36 & 66.48 & 94.24 & 85.98

\\ \hline
\end{tabular}
}
\caption{VisA-LT (imbalance factor=500 ; exp decrease ) Image AUROC}
\label{tab:visa_exp500_ad}
\end{table*}

\begin{table*}
\adjustbox{max width=\textwidth}{
\begin{tabular}{|c||c|c|c||c|c|c|c|c|c|c|c|c|c|c|c|c|c|} \hline
Method  & All & HighShot & LowShot & PCB3 & PCB2 & PCB1  & PCB4 & macaroni1 & macaroni2 & candle & cashew & fryum & capsules
 & chewinggum & pipe fryum \\  \hline
AnomalyGPT & 78.83 & 76.04 & 81.62 & 78.63 & 84.05 & 75.66 & 83.38 & 65.46 & 69.04 & 77.46 & 81.01 & 89.62 & 58.23 & 93.03 & 90.36
 \\ \hline
UniAD  & \underline{94.35} & \underline{94.35} & \underline{94.35} & 97.01 & 96.21 & 91.49 & 96.56 & 95.92 & 88.95 & 97.96 & 98.35 & 94.94 & 81.48 & 97.76 & 95.62
\\ \hline
\hline
Ours w/o PTA & 94.04 & 93.74 & 94.33 & 97.67 & 95.90 & 89.99 & 95.48 & 94.49 & 88.92 & 97.25 & 98.76 & 94.94 & 78.55 & 98.66 & 97.83
\\ \hline
Ours & \textbf{94.66} & \textbf{94.42} & \textbf{94.90} & 97.75 & 95.60 & 92.76 & 96.23 & 94.80 & 89.39 & 97.54 & 98.86 & 95.16 & 80.98 & 98.69 & 98.18
\\ \hline
\end{tabular}
}
\caption{VisA-LT (imbalance factor=500 ; exp decrease ) Pixel AUROC}
\label{tab:visa_exp500_as}
\end{table*}

\begin{table*}
\adjustbox{max width=\textwidth}{
\begin{tabular}{|c||c|c|c||c|c|c|c|c|c|c|c|c|c|c|c|c|c|} \hline
Method  & All & HighShot & LowShot & PCB3 & PCB2 & PCB1  & PCB4 & macaroni1 & macaroni2 & candle & cashew & fryum & capsules
 & chewinggum & pipe fryum \\  \hline
AnomalyGPT & 71.89 & 71.41 & 72.37 & 65.9 & 75.9 & 74.12 & 75.06 & 82.88 & 54.61 & 50.73 & 80.22 & 94.65 & 57.1 & 76.24 & 75.26\\ \hline
RegAD & 71.8 & 66.71 & 76.89 & 61.04 & 65.64 & 72.92 & 81.27 & 66.91 & 52.48 & 74.35 & 73.46 & 71.42 & 57.14 & 91.51 & 93.49 \\ \hline
UniAD  &  78.83 & 83.06 & 74.61 & 79.91 & 85.51 & 88.36 & 97.41 & 82.31 & 64.89 & 77.53 & 60.9 & 77.34 & 56.8 & 93.44 & 81.66 
\\ \hline
\hline
Ours w/o PTA & \underline{82.8} & \underline{85.34} & \underline{80.26} & 84.59 & 83.04 & 88.33 & 96.65 & 85.72 & 73.69 & 80.34 & 76.72 & 77.00 & 64.3 & 94.8 & 88.42 
\\ \hline
Ours & \textbf{84.80} & \textbf{85.98} & \textbf{83.63} & 85.21 & 82.08 & 87.91 & 96.97 & 86.17 & 77.54 & 81.94 & 83.94 & 78.78 & 69.55 & 97.10 & 90.44
\\ \hline
\end{tabular}
}
\caption{VisA-LT (imbalance factor=100 ; step decrease ) Image AUROC}
\label{tab:visa_step100_ad}
\end{table*}

\begin{table*}
\adjustbox{max width=\textwidth}{
\begin{tabular}{|c||c|c|c||c|c|c|c|c|c|c|c|c|c|c|c|c|c|} \hline
Method  & All & HighShot & LowShot & PCB3 & PCB2 & PCB1  & PCB4 & macaroni1 & macaroni2 & candle & cashew & fryum & capsules
 & chewinggum & pipe fryum \\  \hline
AnomalyGPT & 82.3 & 79.41 & 85.19 & 80.35 & 86.1 & 78.21 & 80.42 & 76.54 & 74.83 & 81.56 & 91.79 & 91.02 & 60.06 & 93.07 & 93.63\\ \hline
RegAD &  94.99 & 94.81 & \underline{95.16} & 95.09 & 94.76 & 97.76 & 93.17 & 95.63 & 92.49 & 95.72 & 97.49 & 95.23 & 84.83 & 98.61 & 99.10\\ \hline
UniAD  & 96.04 & 97.37 & 94.71 & 97.25 & 96.91 & 99.18 & 97.42 & 98.21 & 95.30 & 95.89 & 96.89 & 94.24 & 83.91 & 98.51 & 98.84 
\\ \hline
\hline
Ours w/o PTA & \underline{96.16} & \textbf{97.61} & 94.71 & 97.51 & 97.07 & 98.69 & 97.70 & 97.98 & 96.48 & 95.74 & 98.61 & 95.35 & 80.88 & 98.77 & 98.88
\\ \hline
Ours & \textbf{96.57} & \underline{97.43} & \textbf{95.72} & 97.67 & 96.10 & 98.80 & 97.02 & 98.10 & 96.87 & 96.57 & 98.97 & 95.78 & 84.87 & 98.93 & 99.20
\\ \hline
\end{tabular}
}
\caption{VisA-LT (imbalance factor=100 ; step decrease ) Pixel AUROC}
\label{tab:visa_step100_as}
\end{table*}

\begin{table*}
\adjustbox{max width=\textwidth}{
\begin{tabular}{|c||c|c|c||c|c|c|c|c|c|c|c|c|c|c|c|c|c|} \hline
Method  & All & HighShot & LowShot & PCB3 & PCB2 & PCB1  & PCB4 & macaroni1 & macaroni2 & candle & cashew & fryum & capsules
 & chewinggum & pipe fryum \\  \hline
AnomalyGPT & 69.78 & 71.19 & 68.38 & 66.79 & 75.2 & 74.3 & 73.57 & 81.94 & 55.34 & 44.74 & 82.28 & 74.26 & 56.81 & 77.68 & 74.49
\\ \hline
RegAD & 71.65 & 67.02 & 76.28 & 57.60 & 65.20 & 72.95 & 81.89 & 67.74 & 56.76 & 72.69 & 72.51 & 73.50 & 56.74 & 89.39 & 92.86
 \\ \hline
UniAD  &  77.64 & 84.51 & 70.78 & 81.51 & 86.73 & 89.78 & 97.83 & 83.51 & 67.72 & 64.57 & 55.58 & 73.7 & 56.2 & 91.3 & 83.36
\\ \hline
\hline
Ours w/o PTA & \underline{83.79} & \textbf{89.39} & \underline{78.18} & 87.48 & 86.41 & 91.52 & 97.86 & 89.7 & 83.42 & 78.92 & 74.46 & 75.3 & 57.27 & 93.46 & 89.72
\\ \hline
Ours & \textbf{84.03} & \underline{87.34} & \textbf{80.72} & 85.00 & 82.47 & 90
.49 & 96.12 & 88.64 & 81.33 & 81.65 & 84.5 & 71.84 & 61.10 & 95.4 & 89.82
\\ \hline
\end{tabular}
}
\caption{VisA-LT (imbalance factor=200 ; step decrease ) Image AUROC}
\label{tab:visa_step200_ad}
\end{table*}

\begin{table*}
\adjustbox{max width=\textwidth}{
\begin{tabular}{|c||c|c|c||c|c|c|c|c|c|c|c|c|c|c|c|c|c|} \hline
Method  & All & HighShot & LowShot & PCB3 & PCB2 & PCB1  & PCB4 & macaroni1 & macaroni2 & candle & cashew & fryum & capsules
 & chewinggum & pipe fryum \\  \hline
AnomalyGPT & 81.97 & 78.69 & 85.25 & 80.22 & 85.88 & 77.08 & 81.05 & 74.01 & 73.9 & 80.96 & 91.6 & 90.77 & 61.16 & 93.04 & 93.99
\\ \hline
RegAD &  94.52 & 93.73 & \underline{95.32} & 94.80 & 93.28 & 97.06 & 90.95 & 94.87 & 91.44 & 95.24 & 98.00 & 95.30 & 87.24 & 97.00 & 99.14
\\ \hline
UniAD  & 95.66 & 97.45 & 93.87 & 97.41 & 96.87 &
99.15 & 97.53 & 98.38 & 95.4 & 95.37 & 95.29 & 93.63 & 82.25 & 98.1 & 98.61
\\ \hline
\hline
Ours w/o PTA & \underline{95.89} & \underline{97.59} & 94.18 & 97.88 & 96
.46 & 98.95 & 96.38 & 98.52 & 97.39 & 95.31 & 98.53 & 94.70 & 78.98 & 98.72 & 98.89
\\ \hline
Ours & \textbf{96.27} & \textbf{97.63} & \textbf{94.92} & 97.78 & 96.36 & 98
.89 & 96.90 & 98.45 & 97.39 & 95.87 & 98.65 & 95.23 & 81.87 & 98.82 & 99.06
\\ \hline
\end{tabular}
}
\caption{VisA-LT (imbalance factor=200 ; step decrease ) Pixel AUROC}
\label{tab:visa_step200_as}
\end{table*}

\begin{table*}
\adjustbox{max width=\textwidth}{
\begin{tabular}{|c||c|c|c||c|c|c|c|c|c|c|c|c|c|c|c|c|c|} \hline
Method  & All & HighShot & LowShot & PCB3 & PCB2 & PCB1  & PCB4 & macaroni1 & macaroni2 & candle & cashew & fryum & capsules
 & chewinggum & pipe fryum \\  \hline
AnomalyGPT & 62.88 & 70.15 & 55.61 & 63.36 & 74.22 & 74.4 & 75.84 & 79.34 & 53.74 & 41.4 & 71.25 & 66.18 & 49.11 & 41.74 & 63.95
 \\ \hline
UniAD  &  71.84 & 81.84 & 61.85 & 78.79 & 84.72 & 88.8 & 97.44 & 79.46 & 61.83 & 61.82 & 68.54 & 55.96 & 51.3 & 68.28 & 65.22
\\ \hline
\hline
Ours w/o PTA & \underline{82.42} & \underline{89.55} & \underline{75.28} & 88.75 & 86.42 & 90.95 & 97.70 & 89.8 & 83.7 & 79.44 & 61.16 & 70.74 & 65.53 & 90.34 & 84.46
\\ \hline
Ours & 83.33 & \textbf{88.40} & \textbf{78.26 }& 85.68 & 85.19 & 89.95 & 96.68 & 87.91 & 85.00 & 78.59 & 76.56 & 69.86 & 68.30 & 94.34 & 81.88
\\ \hline
\end{tabular}
}
\caption{VisA-LT (imbalance factor=500 ; step decrease ) Image AUROC}
\label{tab:visa_step500_ad}
\end{table*}

\begin{table*}
\adjustbox{max width=\textwidth}{
\begin{tabular}{|c||c|c|c||c|c|c|c|c|c|c|c|c|c|c|c|c|c|} \hline
Method  & All & HighShot & LowShot & PCB3 & PCB2 & PCB1  & PCB4 & macaroni1 & macaroni2 & candle & cashew & fryum & capsules
 & chewinggum & pipe fryum \\  \hline
AnomalyGPT & 81.48 & 78.88 & 84.09 & 79.88 & 84.91 & 78.51 & 79.59 & 75.69 & 74.69 & 78.52 & 92.32 & 90.94 & 59.75 & 89.52 & 93.49
\\ \hline
UniAD  & 95.06 & 97.21 & 92.91 & 97.15 & 96.79 & 99.16 & 97.35 & 97.98 & 94.87 & 93.25 & 91.47 & 90.63 & 90.63 & 96.35 & 95.16
\\ \hline
\hline
Ours w/o PTA & \underline{95.50} & \underline{97.61} & \underline{93.40} & 97.93 & 96.50 & 98.91 & 96.44 & 98.57 & 97.29 & 95.41 & 96.77 & 94.33 & 77.72 & 98.32 & 97.87
\\ \hline
Ours & \textbf{96.41} & \textbf{97.69} & 9\textbf{5.12} & 97.91 & 96.48 & 98.95 & 96.63 & 98.69 & 97.50 & 95.81 & 97.90 & 95.10 & 84.46 & 98.52 & 98.93
\\ \hline
\end{tabular}
}
\caption{VisA-LT (imbalance factor=500 ; step decrease ) Pixel AUROC}
\label{tab:visa_step500_as}
\end{table*}

\begin{table*}
\adjustbox{max width=\textwidth}{
    \begin{tabular}{|c||c|c|c||c|c|c|c|c|c|c|c|c|c|} \hline
Method  & All & HighShot & LowShot & Class10 & Class7 & Class9 & Class8 & Class2 & Class3 & Class5 & Class1 & Class4 & Class6\\ 
\hline
AnomalyGPT  & 84.86 & 84.68 & 85.04 & 82.55 & 100.0 & 87.28 & 70.07 & 83.52 & 93.61 & 96.88 & 83.91 & 53.0 & 97.79\\ \hline
RegAD  & 84.86 & 86.88 & 82.84 & 94.45 & 91.34 & 99.23 & 50.94 & 98.44 & 97.95 & 72.07 & 48.76 & 99.75 & 95.66\\  \hline
UniAD  & 84.51 & 87.34 & 81.69 & 99.25 & 95.29 & 91.61 & 52.97 & 97.61 & 79.73 & 77.55 & 51.19 & 100 & 99.99\\ \hline
\hline
Ours w/o PTA  & \underline{92.97} & \underline{93.17} & \underline{92.77} & 99.50 & 99.58 & 99.35 & 67.44 & 99.96 & 96.14 & 96.17 & 72.40 & 100 & 99.14\\ \hline
Ours  & {\bf 94.82} & {\bf 93.39} & {\bf 96.25} & 97.71 & 99.53 & 99.91 & 69.81 & 100 & 99.69 & 97.68 & 84.08 & 100 & 99.82\\ \hline
\end{tabular}
}
\caption{DAGM-LT (imbalance factor=50 ; exp decrease ) Image AUROC}
\label{tab:dagm_exp50_ad}
\end{table*}

\begin{table*}
\adjustbox{max width=\textwidth}{
    \begin{tabular}{|c||c|c|c||c|c|c|c|c|c|c|c|c|c|} \hline
Method  & All & HighShot & LowShot & Class10 & Class7 & Class9 & Class8 & Class2 & Class3 & Class5 & Class1 & Class4 & Class6\\ 
\hline
AnomalyGPT  & 77.37 & 78.89 & 75.84 & 73.04 & 86.3 & 88.76 & 66.11 & 80.26 & 78.17 & 78.85 & 79.04 & 54.37 & 88.79\\ \hline
RegAD  & 90.29 & 91.91 & 88.66 & 98.74 & 88.41 & 99.95 & 73.28 & 99.17 & 93.69 & 85.23 & 72.23 & 99.73 & 92.44\\ \hline
UniAD  & 90.70 & 89.75 & 91.64 & 99.76 & 93.54 & 99.29 & 56.45 & 99.74 & 90.87 & 89.53 & 79.49 & 98.91 & 99.44\\ \hline
\hline
Ours w/o PTA  & \underline{97.01} & \underline{97.30} & \underline{96.72} & 99.36 & 98.29 & 99.77 & 89.38 & 99.74 & 98.66 & 98.66 & 91.96 & 99.52 & 94.94\\ \hline
Ours & {\bf 97.4} & {\bf 97.11} & {\bf 97.68} & 99.09 & 97.49 & 99.93 & 89.18 & 99.87 & 98.94 & 98.79 & 93.90 & 99.49 & 97.27\\ \hline
\end{tabular}
}
\caption{DAGM-LT (imbalance factor=50 ; exp decrease ) Pixel AUROC}
\label{tab:dagm_exp50_as}
\end{table*}

\begin{table*}
\adjustbox{max width=\textwidth}{
    \begin{tabular}{|c||c|c|c||c|c|c|c|c|c|c|c|c|c|} \hline
Method  & All & HighShot & LowShot & Class10 & Class7 & Class9 & Class8 & Class2 & Class3 & Class5 & Class1 & Class4 & Class6\\ 
\hline
AnomalyGPT &  85.31 & 86.05 & 84.58 & 86.56 & 100.0 & 89.76 & 71.9 & 82.03 & 94.29 & 95.42 & 84.22 & 52.27 & 96.69\\  \hline
RegAD & 84.86 & 86.88 & 82.84 & 94.45 & 91.34 & 99.23 & 50.94 & 98.44 & 97.95 & 72.07 & 48.76 & 99.75 & 95.66\\ 
  \hline
UniAD  & 84.34 & 87.66 & 81.02 & 99.15 & 96.01 & 92.45 & 53.31 & 97.40 & 79.79 & 74.26 & 51.15 & 99.91 & 99.99\\ 
 \hline
 \hline
Ours w/o PTA & \underline{93.35} & \underline{92.94} & \underline{93.76} & 97.76 & 98.58 & 99.80 & 68.55 & 99.99 & 99.68 & 97.15 & 72.25 & 100 & 99.72\\  \hline
Ours & {\bf 94.40} & {\bf 93.24} & {\bf 95.56} & 97.78 & 99.60 & 99.89 & 68.92 & 100 & 99.66 & 97.35 & 81.01 & 100 & 99.78\\ \hline
\end{tabular}
}
\caption{DAGM-LT (imbalance factor=100 ; exp decrease ) Image AUROC}
\label{tab:dagm_exp100_ad}
\end{table*}

\begin{table*}
\adjustbox{max width=\textwidth}{
    \begin{tabular}{|c||c|c|c||c|c|c|c|c|c|c|c|c|c|} \hline
Method  & All & HighShot & LowShot & Class10 & Class7 & Class9 & Class8 & Class2 & Class3 & Class5 & Class1 & Class4 & Class6\\ 
\hline
AnomalyGPT  & 77.20 & 79.19 & 75.21 & 74.34 & 86.14 & 89.99 & 66.26 & 79.22 & 77.37 & 77.79 & 80.49 & 54.93 & 85.45\\ \hline
RegAD & 90.28 & 91.91 & 88.66 & 98.74 & 88.40 & 99.95 & 73.28 & 99.17 & 93.69 & 85.23 & 72.23 & 99.73 & 92.44\\ \hline
UniAD & 90.13 & 89.66 & 90.60 & 99.75 & 93.54 & 99.33 & 55.96 & 99.76 & 90.13 & 87.54 & 78.19 & 97.74 & 99.42\\ \hline
\hline
Ours w/o PTA & \underline{96.93} & \underline{96.82} & \underline{97.04} & 99.04 & 96.87 & 99.89 & 88.47 & 99.83 & 98.76 & 98.52 & 92.59 & 99.53 & 95.78\\ \hline
Ours & {\bf 97.30} & {\bf 97.01} & {\bf 97.58} & 99.09 & 97.42 & 99.93 & 88.74 & 99.87 & 98.95 & 98.62 & 93.76 & 99.59 & 96.98\\ \hline
\end{tabular}
}
\caption{DAGM-LT (imbalance factor=100 ; exp decrease ) Pixel AUROC}
\label{tab:dagm_exp100_as}
\end{table*}

\begin{table*}
\adjustbox{max width=\textwidth}{
    \begin{tabular}{|c||c|c|c||c|c|c|c|c|c|c|c|c|c|} \hline
Method  & All & HighShot & LowShot & Class10 & Class7 & Class9 & Class8 & Class2 & Class3 & Class5 & Class1 & Class4 & Class6\\ 
\hline
AnomalyGPT & 83.29 & 83.45 & 83.13 & 84.31 & 100.0 & 87.16 & 66.8 & 79.0 & 89.9 & 95.65 & 80.03 & 52.49 & 97.56\\ \hline
RegAD & 84.86 & 86.88 & 82.84 & 94.45 & 91.34 & 99.23 & 50.94 & 98.44 & 97.95 & 72.07 & 48.76 & 99.75 & 95.66\\  \hline
UniAD & 83.56 & 87.09 & 80.02 & 99.40 & 95.42 & 89.82 & 53.32 & 97.53 & 79.13 & 72.48 & 50.52 & 97.98 & 99.99\\ \hline
\hline
Ours w/o PTA & \underline{92.83} & {\bf 93.33} & \underline{92.34} & 99.49 & 99.24 & 99.21 & 68.69 & 100 & 98.78 & 96.45 & 67.25 & 100 & 99.21\\ \hline
Ours & {\bf 94.29} & \underline{93.08} & {\bf 95.51} & 97.78 & 99.62 & 99.80 & 68.21 & 100 & 99.51 & 97.34 & 80.90 & 100 & 99.82\\ \hline
\end{tabular}
}
\caption{DAGM-LT (imbalance factor=200 ; exp decrease ) Image AUROC}
\label{tab:dagm_exp200_ad}
\end{table*}

\begin{table*}
\adjustbox{max width=\textwidth}{
    \begin{tabular}{|c||c|c|c||c|c|c|c|c|c|c|c|c|c|} \hline
Method  & All & HighShot & LowShot & Class10 & Class7 & Class9 & Class8 & Class2 & Class3 & Class5 & Class1 & Class4 & Class6\\ 
\hline
AnomalyGPT  & 77.16 & 79.58 & 74.73 & 74.85 & 85.83 & 89.38 & 65.89 & 81.93 & 73.3 & 77.54 & 81.12 & 56.02 & 85.69\\ \hline
RegAD  & 90.29 & 91.91 & 88.66 & 98.74 & 88.40 & 99.95 & 73.28 & 99.17 & 93.69 & 85.23 & 72.23 & 99.73 & 92.44\\ \hline
UniAD  & 89.73 & 89.57 & 89.88 & 99.77 & 93.28 & 99.22 & 55.84 & 99.76 & 90.30 & 86.78 & 76.53 & 96.44 & 99.39\\ \hline
\hline
Ours w/o PTA & \underline{96.16} & {\bf 97.15} & \underline{95.18} & 99.34 & 97.83 & 99.87 & 88.86 & 99.83 & 98.08 & 97.91 & 89.89 & 99.25 & 90.75\\ \hline
Ours & {\bf 97.19} & \underline{96.98} & {\bf 97.39} & 99.02 & 97.38 & 99.92 & 88.72 & 99.86 & 98.88 & 98.50 & 93.26 & 99.52 & 96.80\\ \hline
\end{tabular}
}
\caption{DAGM-LT (imbalance factor=200 ; exp decrease ) Pixel AUROC}
\label{tab:dagm_exp200_as}
\end{table*}

\begin{table*}
\adjustbox{max width=\textwidth}{
    \begin{tabular}{|c||c|c|c||c|c|c|c|c|c|c|c|c|c|} \hline
Method  & All & HighShot & LowShot & Class10 & Class7 & Class9 & Class8 & Class2 & Class3 & Class5 & Class1 & Class4 & Class6\\ 
\hline
AnomalyGPT  & 83.47 & 84.38 & 82.55 & 86.38 & 99.99 & 91.88 & 64.78 & 78.88 & 90.31 & 95.06 & 79.88 & 52.72 & 94.78\\ \hline
RegAD & 84.86 & 86.88 & 82.84 & 94.45 & 91.34 & 99.23 & 50.94 & 98.44 & 97.95 & 72.07 & 48.76 & 99.75 & 95.66\\ 
\hline
UniAD  & 81.35 & 88.75 & 73.96 & 99.89 & 98.39 & 96.60 & 51.31 & 97.56 & 85.24 & 75.57 & 53.49 & 55.64 & 99.84\\ 
 \hline
 \hline
Ours w/o PTA & \underline{92.08} & {\bf 93.55} & \underline{90.6} & 99.56 & 99.37 & 99.13 & 69.71 & 100 & 98.41 & 94.91 & 60.47 & 100 & 99.21\\ 
\hline
Ours & {\bf 93.54} & \underline{93.23} & {\bf 93.85} & 97.76 & 99.77 & 99.79 & 68.81 & 100 & 99.36 & 97.04 & 73.00 & 100 & 99.83\\ \hline
\end{tabular}
}
\caption{DAGM-LT (imbalance factor=500 ; exp decrease ) Image AUROC}
\label{tab:dagm_exp500_ad}
\end{table*}

\begin{table*}
\adjustbox{max width=\textwidth}{
    \begin{tabular}{|c||c|c|c||c|c|c|c|c|c|c|c|c|c|} \hline
Method  & All & HighShot & LowShot & Class10 & Class7 & Class9 & Class8 & Class2 & Class3 & Class5 & Class1 & Class4 & Class6\\ 
\hline
AnomalyGPT & 76.87 & 78.93 & 74.81 & 76.11 & 85.42 & 90.89 & 62.51 & 79.73 & 72.28 & 77.29 & 82.21 & 57.35 & 84.93\\ \hline
RegAD  & 90.29 & 91.91 & 88.66 & 98.74 & 88.40 & 99.95 & 73.28 & 99.17 & 93.69 & 85.23 & 72.23 & 99.73 & 92.44\\ \hline
UniAD  & 88.63 & 89.44 & 87.82 & 99.77 & 93.10 & 99.22 & 55.37 & 99.73 & 89.05 & 84.62 & 75.15 & 90.93 & 99.33\\ \hline
\hline
Ours w/o PTA & \underline{95.99} & {\bf 97} & \underline{94.99} & 99.40 & 97.75 & 99.86 & 88.16 & 99.83 & 98.09 & 97.62 & 88.45 & 99.05 & 91.72\\ \hline
Ours & {\bf 97.01} & \underline{96.94} & {\bf 97.07} & 99.02 & 97.33 & 99.92 & 88.57 & 99.87 & 98.74 & 98.43 & 92.23 & 99.31 & 96.63\\ \hline
\end{tabular}
}
\caption{DAGM-LT (imbalance factor=500 ; exp decrease ) Pixel AUROC}
\label{tab:dagm_exp500_as}
\end{table*}

\begin{table*}
\adjustbox{max width=\textwidth}{
    \begin{tabular}{|c||c|c|c||c|c|c|c|c|c|c|c|c|c|} \hline
Method  & All & HighShot & LowShot & Class6 & Class4 & Class1 & Class5 & Class3 & Class2 & Class8 & Class9 & Class7 & Class10\\ 
\hline
AnomalyGPT &  84.65 & 86.58 & 82.72 & 99.11 & 67.43 & 74.65 & 97.48 & 94.21 & 88.79 & 63.56 & 92.07 & 99.96 & 69.2\\  \hline
RegAD &  84.85 & 82.83 & 86.88 & 95.66 & 99.75 & 48.76 & 72.07 & 97.95 & 98.44 & 50.94 & 99.23 & 91.34 & 94.45\\ 
 \hline
UniAD  & 82.48 & 85.78 & 79.17 & 99.93 & 100 & 63.41 & 79.04 & 86.52 & 97.00 & 51.42 & 98.31 & 71.48 & 77.67\\ 
 \hline
 \hline
Ours w/o PTA & \underline{92.62} & \underline{95.19} & \underline{90.05} & 99.85 & 100 & 80.02 & 96.88 & 99.22 & 99.99 & 68.40 & 99.09 & 90.46 & 92.31\\ 
 \hline
Ours & {\bf 94.09} & {\bf 96.85} & {\bf 91.34} & 99.76 & 100 & 88.00 & 97.19 & 99.29 & 100 & 67.80 & 99.75 & 95.03 & 94.12\\  \hline
\end{tabular}
}
\caption{DAGM-LT (imbalance factor=200 ; reverse exp decrease ) Image AUROC}
\label{tab:dagm_revexp200_ad}
\end{table*}

\begin{table*}
\adjustbox{max width=\textwidth}{
    \begin{tabular}{|c||c|c|c||c|c|c|c|c|c|c|c|c|c|} \hline
Method  & All & HighShot & LowShot & Class6 & Class4 & Class1 & Class5 & Class3 & Class2 & Class8 & Class9 & Class7 & Class10\\ 
\hline
AnomalyGPT  & 78.41 & 76.49 & 8043 & 94.92 & 57.21 & 69.48 & 81.27 & 79.1 & 89.15 & 65.79 & 95.83 & 83.85 & 67.53\\ \hline
RegAD &  90.28 & 88.66 & 91.91 & 92.44 & 99.73 & 72.23 & 85.23 & 93.69 & 99.17 & 73.28 & 99.95 & 88.40 & 98.74\\ \hline
UniAD &  89.98 & 92.69 & 87.27 & 99.49 & 99.86 & 84.59 & 90.23 & 89.30 & 99.65 & 53.89 & 99.55 & 85.42 & 97.87\\ \hline
\hline
Ours w/o PTA & \underline{96.22} & \underline{97.24} & \underline{95.20} & 96.65 & 99.52 & 93.38 & 98.30 & 98.38 & 99.80 & 85.42 & 99.78 & 93.15 & 97.87\\ \hline
Ours & {\bf 96.68} & {\bf 97.63} & {\bf 95.72} & 97.16 & 99.43 & 94.76 & 98.30 & 98.52 & 99.87 & 85.23 & 99.92 & 95.11 & 98.49\\  \hline
\end{tabular}
}
\caption{DAGM-LT (imbalance factor=200 ; reverse exp decrease ) Pixel AUROC}
\label{tab:dagm_revexp200_as}
\end{table*}

\begin{table*}
\adjustbox{max width=\textwidth}{
    \begin{tabular}{|c||c|c|c||c|c|c|c|c|c|c|c|c|c|} \hline
Method  & All & HighShot & LowShot & Class10 & Class7 & Class9 & Class8 & Class2 & Class3 & Class5 & Class1 & Class4 & Class6\\ 
\hline
AnomalyGPT & 87.27 & 88.29 & 86.24 & 89.51 & 100.0 & 88.81 & 70.88 & 92.25 & 95.97 & 96.69 & 83.77 & 57.59 & 97.2\\ \hline
RegAD & 84.85 & 86.88 & 82.83 & 94.45 & 91.34 & 99.23 & 50.94 & 98.44 & 97.95 & 72.07 & 48.76 & 99.75 & 95.66\\ \hline
UniAD & 82.37 & 87.11 & 77.63 & 99.06 & 95.02 & 92.84 & 51.38 & 97.26 & 67.98 & 70.70 & 50.01 & 99.48 & 99.99\\  \hline
\hline
Ours w/o PTA & \underline{92.6} & {\bf 93.29} & \underline{91.91} & 99.26 & 99.62 & 99.83 & 67.74 & 99.98 & 96.73 & 95.94 & 67.66 & 100 & 99.24\\ 
 \hline
Ours & {\bf 94.03} & \underline{93.2} & {\bf 94.85} & 97.67 & 99.57 & 99.85 & 68.92 & 99.99 & 99.38 & 97.17 & 77.95 & 100 & 99.77\\  \hline
\end{tabular}
}
\caption{DAGM-LT (imbalance factor=50 ; step decrease ) Image AUROC}
\label{tab:dagm_step50_ad}
\end{table*}

\begin{table*}
\adjustbox{max width=\textwidth}{
    \begin{tabular}{|c||c|c|c||c|c|c|c|c|c|c|c|c|c|} \hline
Method  & All & HighShot & LowShot & Class6 & Class4 & Class1 & Class5 & Class3 & Class2 & Class8 & Class9 & Class7 & Class10\\ 
\hline
AnomalyGPT & 78.28 & 79.78 & 76.77 & 75.75 & 86.67 & 89.54 & 66.73 & 80.23 & 79.02 & 79.42 & 81.74 & 55.34 & 88.31\\ \hline
RegAD &  90.28 & 91.90 & 88.66 & 98.74 & 88.40 & 99.95 & 73.28 & 99.17 & 93.69 & 85.23 & 72.23 & 99.73 & 92.44\\ \hline
UniAD &  89.52 & 89.87 & 89.16 & 99.71 & 93.11 & 99.36 & 57.49 & 99.71 & 86.33 & 84.15 & 78.58 & 97.41 & 99.37\\ \hline
\hline
Ours w/o PTA & \underline{96.93} & \underline{97.49} & \underline{96.37} & 99.35 & 98.45 & 99.87 & 90.00 & 99.79 & 98.32 & 98.37 & 90.58 & 99.51 & 95.09\\ \hline
Ours & {\bf 97.20} & {\bf 97.00} & {\bf 97.39} & 99.06 & 97.30 & 99.92 & 88.85 & 99.86 & 98.92 & 98.59 & 93.21 & 99.48 & 96.76\\ \hline
\end{tabular}
}
\caption{DAGM-LT (imbalance factor=50 ; step decrease ) Pixel AUROC}
\label{tab:dagm_step50_as}
\end{table*}

\begin{table*}
\adjustbox{max width=\textwidth}{
    \begin{tabular}{|c||c|c|c||c|c|c|c|c|c|c|c|c|c|} \hline
Method  & All & HighShot & LowShot & Class10 & Class7 & Class9 & Class8 & Class2 & Class3 & Class5 & Class1 & Class4 & Class6\\ 
\hline
AnomalyGPT & 86.48 & 88.35 & 84.60 & 90.24 & 100.0 & 87.79 & 70.42 & 93.31 & 95.21 & 96.41 & 81.82 & 51.01 & 98.56\\  \hline
RegAD &  84.86 & 86.88 & 82.84 & 94.45 & 91.34 & 99.23 & 50.94 & 98.44 & 97.95 & 72.07 & 48.76 & 99.75 & 95.66\\  \hline
UniAD & 81.11 & 87.19 & 75.03 & 99.12 & 95.23 & 91.99 & 52.38 & 97.23 & 67.80 & 69.05 & 50.48 & 87.83 & 99.99\\  \hline
\hline
Ours w/o PTA & \underline{91.94} & {\bf 93.26} & \underline{90.62} & 99.37 & 99.59 & 99.77 & 67.58 & 99.98 & 97.55 & 94.95 & 61.21 & 99.98 & 99.41\\ \hline
Ours & {\bf 93.97} & \underline{93.13} & {\bf 94.81} & 97.85 & 99.31 & 99.87 & 68.62 & 99.98 & 99.28 & 96.95 & 78.01 & 100 & 99.82\\  \hline
\end{tabular}
}
\caption{DAGM-LT (imbalance factor=100 ; step decrease ) Image AUROC}
\label{tab:dagm_step100_ad}
\end{table*}

\begin{table*}
\adjustbox{max width=\textwidth}{
    \begin{tabular}{|c||c|c|c||c|c|c|c|c|c|c|c|c|c|} \hline
Method  & All & HighShot & LowShot & Class6 & Class4 & Class1 & Class5 & Class3 & Class2 & Class8 & Class9 & Class7 & Class10\\ 
\hline
AnomalyGPT & 78.76 & 80.23 & 77.29 & 77.01 & 86.46 & 89.08 & 67.51 & 81.07 & 79.4 & 80.0 & 83.19 & 55.72 & 88.14\\ \hline
RegAD & 90.28 & 91.91 & 88.66 & 98.74 & 88.40 & 99.95 & 73.28 & 99.17 & 93.69 & 85.23 & 72.23 & 99.73 & 92.45\\ \hline
UniAD  & 89.11 & 90.08 & 88.15 & 99.71 & 93.50 & 99.25 & 58.29 & 99.68 & 84.99 & 83.20 & 78.56 & 94.66 & 99.34\\ \hline
\hline
Ours w/o PTA & \underline{96.38} & {\bf 97.47} & \underline{95.29} & 99.36 & 98.36 & 99.84 & 90.00 & 99.78 & 97.66 & 97.49 & 88.61 & 99.22 & 93.49\\ \hline
Ours  & {\bf 97.07} & \underline{96.94} & {\bf 97.20} & 99.11 & 97.16 & 99.92 & 88.63 & 99.86 & 98.51 & 98.52 & 92.69 & 99.44 & 96.83\\ \hline
\end{tabular}
}
\caption{DAGM-LT (imbalance factor=100 ; step decrease ) Pixel AUROC}
\label{tab:dagm_step100_as}
\end{table*}

\begin{table*}
\adjustbox{max width=\textwidth}{
    \begin{tabular}{|c||c|c|c||c|c|c|c|c|c|c|c|c|c|} \hline
Method  & All & HighShot & LowShot & Class10 & Class7 & Class9 & Class8 & Class2 & Class3 & Class5 & Class1 & Class4 & Class6\\ 
\hline
AnomalyGPT & 84.73 & 87.34 & 82.12 & 89.58 & 100.0 & 86.36 & 70.28 & 90.49 & 95.43 & 96.29 & 79.86 & 41.35 & 97.66\\ \hline
RegAD & 84.86 & 86.88 & 82.84 & 94.45 & 91.34 & 99.23 & 50.94 & 98.44 & 97.95 & 72.07 & 48.76 & 99.75 & 95.66\\ \hline
UniAD & 80.33 & 88.85 & 71.81 & 99.80 & 98.48 & 97.68 & 50.38 & 97.92 & 81.13 & 72.95 & 51.43 & 53.55 & 99.97\\  \hline
\hline
Ours w/o PTA & \underline{91.78} & \underline{93.27} & \underline{90.28} & 99.17 & 98.60 & 99.92 & 68.66 & 100 & 97.92 & 93.29 & 60.82 & 100 & 99.39\\  \hline
Ours & {\bf 93.79} & {\bf 93.69} & {\bf 93.90} & 98.74 & 99.52 & 99.91 & 70.26 & 100 & 99.02 & 95.46 & 75.41 & 100 & 99.59\\  \hline
\end{tabular}
}
\caption{DAGM-LT (imbalance factor=200 ; step decrease ) Image AUROC}
\label{tab:dagm_step200_ad}
\end{table*}

\begin{table*}
\adjustbox{max width=\textwidth}{
    \begin{tabular}{|c||c|c|c||c|c|c|c|c|c|c|c|c|c|} \hline
Method  & All & HighShot & LowShot & Class6 & Class4 & Class1 & Class5 & Class3 & Class2 & Class8 & Class9 & Class7 & Class10\\ 
\hline
AnomalyGPT  & 78.29 & 79.40 & 77.19 & 75.97 & 86.68 & 89.22 & 65.88 & 79.26 & 80.28 & 79.87 & 82.95 & 55.26 & 87.57\\ \hline
RegAD  & 90.29 & 91.91 & 88.66 & 98.74 & 88.4 & 99.95 & 73.28 & 99.17 & 93.69 & 85.23 & 72.23 & 99.73 & 92.44\\ \hline
UniAD & 89.07 & 90.61 & 87.40 & 99.72 & 94.02 & 99.43 & 60.11 & 99.77 & 85.71 & 83.04 & 78.28 & 90.59 & 99.40\\ \hline
\hline
Ours w/o PTA & \underline{96.04} & \underline{97.19} & \underline{94.89} & 99.28 & 97.57 & 99.91 & 89.31 & 99.86 & 97.51 & 97.13 & 88.26 & 99.08 & 92.45\\ \hline
Ours & {\bf 96.84} & {\bf 97.33} & {\bf 96.35} & 99.24 & 97.63 & 99.93 & 89.98 & 99.89 & 98.12 & 97.40 & 91.74 & 99.17 & 95.33\\ \hline
\end{tabular}
}
\caption{DAGM-LT (imbalance factor=200 ; step decrease ) Pixel AUROC}
\label{tab:dagm_step200_as}
\end{table*}

\begin{table*}
\adjustbox{max width=\textwidth}{
    \begin{tabular}{|c||c|c|c||c|c|c|c|c|c|c|c|c|c|} \hline
Method  & All & HighShot & LowShot & Class10 & Class7 & Class9 & Class8 & Class2 & Class3 & Class5 & Class1 & Class4 & Class6\\ 
\hline
AnomalyGPT & 85.08 & 88.73 & 81.43 & 91.35 & 100.0 & 89.07 & 71.74 & 91.48 & 93.3 & 96.99 & 77.94 & 41.77 & 97.14\\  \hline
RegAD & 84.86 & 86.88 & 82.84 & 94.45 & 91.34 & 99.23 & 50.94 & 98.44 & 97.95 & 72.07 & 48.76 & 99.75 & 95.66\\  \hline
UniAD  & 80.04 & 88.88 & 71.18 & 99.77 & 98.62 & 97.19 & 50.88 & 97.98 & 76.32 & 74.09 & 51.57 & 53.98 & 99.98\\  \hline
\hline
Ours w/o PTA & \underline{91.82} & \underline{93.83} & \underline{89.8} & 99.54 & 99.59 & 99.89 & 70.15 & 100 & 95.91 & 95.44 & 58.35 & 99.97 & 99.33\\ \hline
Ours & {\bf 92.78} & {\bf 93.96} & {\bf 91.59} & 99.36 & 99.84 & 99.88 & 70.77 & 99.97 & 96.85 & 95.99 & 65.57 & 100 & 99.52\\  \hline
\end{tabular}
}
\caption{DAGM-LT (imbalance factor=500 ; step decrease ) Image AUROC}
\label{tab:dagm_step500_ad}
\end{table*}

\begin{table*}
\adjustbox{max width=\textwidth}{
    \begin{tabular}{|c||c|c|c||c|c|c|c|c|c|c|c|c|c|} \hline
Method  & All & HighShot & LowShot & Class6 & Class4 & Class1 & Class5 & Class3 & Class2 & Class8 & Class9 & Class7 & Class10\\ 
\hline
AnomalyGPT & 78.75 & 79.93 & 77.57 & 77.9 & 86.27 & 89.89 & 65.79 & 79.78 & 78.81 & 80.68 & 84.82 & 56.79 & 86.73\\ \hline
RegAD  & 90.29 & 91.91 & 88.66 & 98.74 & 88.4 & 99.95 & 73.28 & 99.17 & 93.69 & 85.23 & 72.23 & 99.73 & 92.44\\ \hline
UniAD & 88.53 & 90.43 & 86.62 & 99.73 & 94.39 & 99.42 & 58.89 & 99.75 & 85.70 & 83.61 & 79.19 & 85.25 & 99.39\\ \hline
\hline
Ours w/o PTA & \underline{95.64} & \underline{97.51} & \underline{93.76} & 99.49 & 98.31 & 99.90 & 90.02 & 99.84 & 96.27 & 97.43 & 83.76 & 98.85 & 92.48\\ \hline
Ours & {\bf 96.65} & {\bf 97.67} & {\bf 95.64} & 99.45 & 98.44 & 99.92 & 90.66 & 99.87 & 97.07 & 97.88 & 88.97 & 98.89 & 95.39\\ \hline
\end{tabular}
}
\caption{DAGM-LT (imbalance factor=500 ; step decrease ) Pixel AUROC}
\label{tab:dagm_step500_as}
\end{table*}

\begin{table*}
\adjustbox{max width=\textwidth}{
    \begin{tabular}{|c||c|c|c||c|c|c|c|c|c|c|c|c|c|} \hline
Method  & All & HighShot & LowShot & Class10 & Class7 & Class9 & Class8 & Class2 & Class3 & Class5 & Class1 & Class4 & Class6\\ 
\hline
AnomalyGPT  & 83.64 & 88.06 & 79.21 & 98.71 & 66.42 & 80.31 & 98.41 & 96.47 & 70.21 & 64.49 & 91.11 & 100.0 & 70.22\\ \hline
RegAD  & 84.86 & 82.84 & 86.88 & 95.66 & 99.75 & 48.76 & 72.07 & 97.95 & 98.44 & 50.94 & 99.23 & 91.34 & 94.45\\ \hline
UniAD & 83.57 & 88.89 & 78.25 & 99.98 & 99.99 & 72.96 & 81.28 & 90.28 & 95.34 & 50.64 & 98.05 & 65.58 & 81.67\\  \hline
\hline
Ours w/o PTA & \underline{92.35} & \underline{96.03} & \underline{88.67} & 99.85 & 100 & 83.72 & 96.91 & 99.68 & 99.77 & 63.16 & 97.96 & 88.87 & 93.59\\  \hline
Ours  & {\bf 93.89} & {\bf 97.4} & {\bf 90.39} & 99.79 & 100 & 90.40 & 97.14 & 99.67 & 99.86 & 64.86 & 99.00 & 93.73 & 94.48\\  \hline
\end{tabular}
}
\caption{DAGM-LT (imbalance factor=200 ; reverse step decrease ) Image AUROC}
\label{tab:dagm_revstep200_ad}
\end{table*}

\begin{table*}
\adjustbox{max width=\textwidth}{
    \begin{tabular}{|c||c|c|c||c|c|c|c|c|c|c|c|c|c|} \hline
Method  & All & HighShot & LowShot & Class6 & Class4 & Class1 & Class5 & Class3 & Class2 & Class8 & Class9 & Class7 & Class10\\ 
\hline
AnomalyGPT & 77.11 & 75.54 & 78.68 & 91.83 & 54.33 & 67.86 & 81.32 & 82.37 & 79.32 & 69.19 & 97.15 & 85.78 & 61.97\\ \hline
RegAD &  90.28 & 88.67 & 91.91 & 92.44 & 99.73 & 72.23 & 85.24 & 93.69 & 99.17 & 73.28 & 99.95 & 88.41 & 98.74\\ \hline
UniAD &  90.79 & 94.62 & 86.96 & 99.47 & 99.82 & 87.93 & 93.47 & 92.41 & 99.28 & 54.11 & 99.48 & 83.94 & 98.00\\ \hline
\hline
Ours w/o PTA & \underline{95.89} & \underline{97.52} & \underline{94.26} & 96.42 & 99.49 & 94.31 & 98.52 & 98.87 & 99.70 & 82.78 & 99.66 & 91.47 & 97.72\\ \hline
Ours  & {\bf 96.66} & {\bf 97.93} & {\bf 95.40} & 97.22 & 99.49 & 95.31 & 98.57 & 99.04 & 99.79 & 84.36 & 99.89 & 94.52 & 98.45\\ \hline
\end{tabular}
}
\caption{DAGM-LT (imbalance factor=200 ; reverse step decrease ) Pixel AUROC}
\label{tab:dagm_revstep200_as}
\end{table*}

\end{document}